\documentclass{article} 
\usepackage{arxiv,times}
\usepackage{booktabs}
\usepackage{makecell}
\usepackage{graphicx}
\usepackage{rotating}
\usepackage{array}

\usepackage{amsmath,amsfonts,bm}

\def\eqref#1{equation~\ref{#1}}

\def\1{\bm{1}}

\DeclareMathAlphabet{\mathsfit}{\encodingdefault}{\sfdefault}{m}{sl}
\SetMathAlphabet{\mathsfit}{bold}{\encodingdefault}{\sfdefault}{bx}{n}

\usepackage{hyperref}
\usepackage{url}
\usepackage{wrapfig}
\usepackage{tabularx}
\usepackage{caption}
\usepackage{subcaption}
\usepackage{xcolor}
\usepackage{adjustbox}
\usepackage{float}
\usepackage{amsmath}
\usepackage{amssymb}
\usepackage{multirow}
\usepackage{longtable}
\usepackage{xltabular}
\usepackage{etoc}
\usepackage{needspace}

\DeclareMathOperator{\Unif}{Unif}
\newcommand{\lead}[1]{\textbf{#1}\ }

\newcommand{\LogUnif}{\operatorname{LogUniform}}

\title{PDE-JEPA: Predictive Representation Learning of Latent Dynamics Modeling for Parametric PDEs}

\author{Zhentao Tan, Jianrong Zhang, Ruijie Quan \& Yi Yang* \\
Zhejiang University. *Corresponding author. \\
\texttt{tanzhentao@zju.edu.cn, zhangjrjlu@gmail.com, quanruijie@zju.edu.cn} \\
\texttt{yangyics@zju.edu.cn}
}

\begin{document}

\maketitle


\begin{abstract}

Physical trajectories contain more than snapshots of a system: they also reveal how its states evolve under governing conditions. However, representation learning for parametric partial differential equations (PDEs) has largely relied on reconstruction-based objectives that emphasize recovering observed physical fields. In this paper, we investigate predictive representation pretraining as an alternative to reconstruction-based learning. We find that predictive representations preserve rich physical information, yet this advantage alone does not ensure accurate field evolution. Based on these observations, we introduce PDE-JEPA for parametric PDE dynamics. Specifically, we first train an encoder using a masked-latent prediction to capture the underlying regularities of PDE dynamics. To explicitly adapt the pretrained representation toward a more dynamics-aligned state space, we then introduce a geometry projector that aligns latent trajectory geometry with the evolution geometry of physical fields. Finally, building on this geometry-aligned latent space, we further develop a physics-structured latent predictor that decomposes the dynamics into parameter-independent evolution and parameter-dependent response components. 
Extensive experiments on nine widely used PDE benchmarks demonstrate that our framework outperforms existing state-of-the-art methods by an average of 33.4\% in-distribution, while achieving an average improvement of 51.4\% when extrapolating to unseen governing parameters. The project page is available \href{https://tanpig-x.github.io/PDE-JEPA/}{here}.

\end{abstract}

\section{Introduction}


Real-world systems exhibit complex dynamics across diverse physical processes~\citep{cross1993pattern}, which are mainly governed by partial differential equations (PDEs)~\citep{evans2022partial}. Solving these equations typically relies on classical numerical methods~\citep{leveque2007finite}, which can achieve high accuracy but often incur substantial computational cost. 
This computational bottleneck has motivated learning-based PDE solvers that approximate physical evolution directly from data~\citep{li2020fourier,karniadakis2021physics}. However, practical applications~\citep{wang2024recent,tan2026harnessing} often involve variations in coefficients, forcing terms, and boundary conditions across physical environments. Parametric PDEs describe the resulting families of related dynamics, requiring models to capture temporal evolution while generalizing to unseen governing conditions.


Existing approaches to parametric PDEs have largely focused on improving models through parameter conditioning~\citep{takamoto2023learning,zhou2024unisolver,bischof2026hypino}, adaptation across environments~\citep{kassai2024boosting,yang2026physics}, in-context learning~\citep{kassai2026enma,patel2026continuum}, or fine-tuning pretrained foundation models~\citep{hao2024dpot,herde2024poseidon,mccabe2025walrus}. In latent dynamics models, the latent state representations are commonly learned through reconstruction-based objectives~\citep{serrano2024aroma,wang2024latent,hagnberger2026calm}, sometimes augmented with dynamics-aware losses~\citep{wu2022learning,li2025latent}. However, reconstruction fidelity alone does not establish whether physically relevant information is readily accessible~\citep{qu2026representation}. Moreover, it does not guarantee that the learned latent space is well suited for accurate temporal evolution~\citep{brettin2025learning}, particularly under shifts in governing conditions. This raises a central question: \emph{what makes a latent representation suitable for accurate prediction and extrapolation of parametric PDE dynamics?}

\begin{figure}[t]
    \centering
    \includegraphics[width=\linewidth]{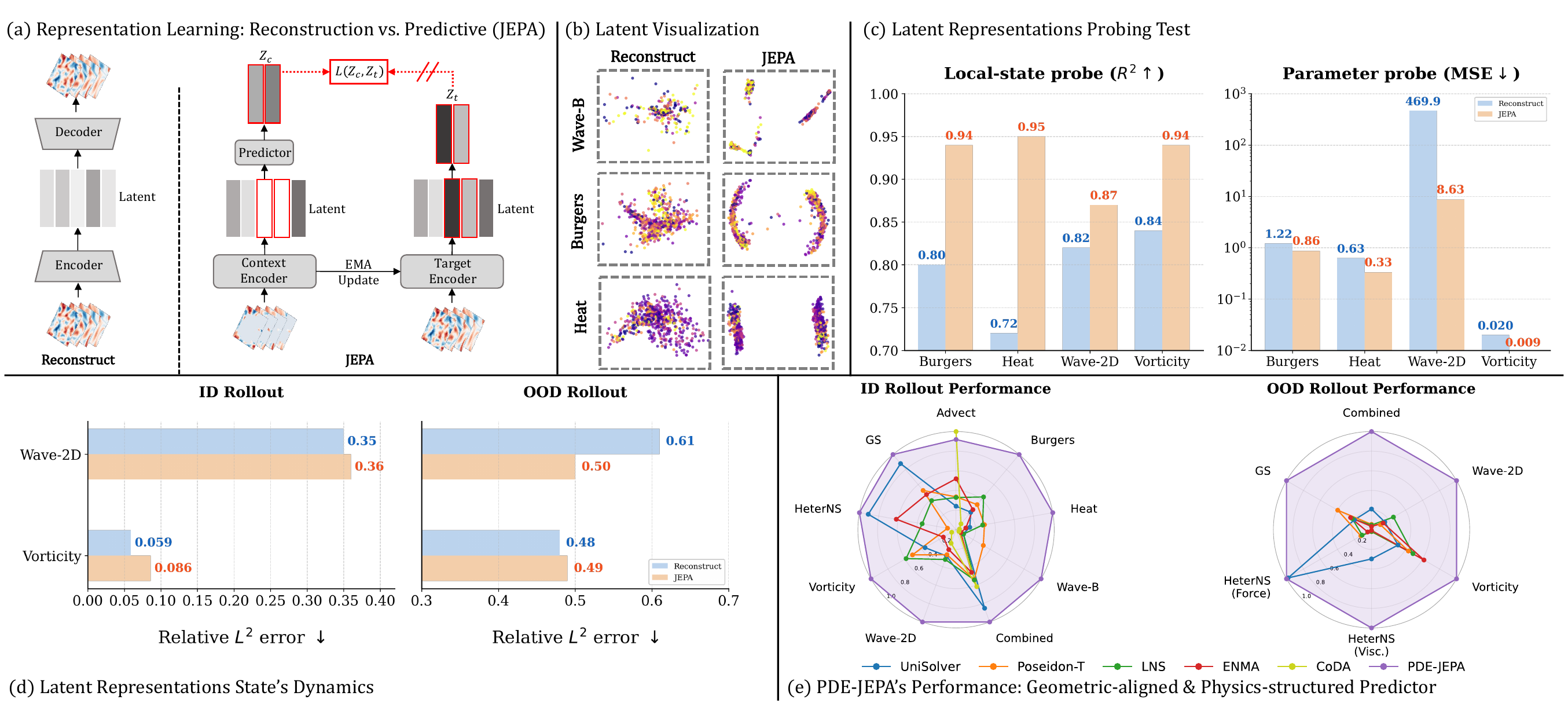}
    \caption{\textbf{Overview of our key observations.} Predictive representation (JEPA) learns physically informative representations, but informativeness alone does not ensure effective forecasting of parametric PDEs. Our approach, PDE-JEPA, improves PDE forecasting and extrapolation through geometry alignment and physics-structured latent prediction.}
    \label{fig:teaser}
\vspace{-15pt}
\end{figure}

We explore this question through JEPA-based predictive representation learning. As illustrated in Figure~\ref{fig:teaser}(a), we predict masked latent targets rather than reconstructing observations~\citep{assran2023self}.
While such objectives have achieved considerable success in world modeling~\citep{terver2025drives, mur2026v,klindt2026does, maes2026leworldmodel, yan2026forward} and demonstrated early promise in parameter probing~\citep{qu2026representation}, their utility for parametric PDE dynamics remains largely unexplored. We therefore conduct an in-depth dissection of reconstruction-based and predictive representations through frozen-encoder probes and autoregressive rollout evaluation. Figure~\ref{fig:teaser}(b--d) shows the results, with additional analyses in Appendix~\ref{app:repana}. Our analysis yields two key observations.

\textbf{Observation 1: Predictive learning yields more informative representations of physical dynamics.}
We first examine how reconstruction-based and predictive learning organize physical information in latent space. Figure~\ref{fig:teaser}(b) suggests clearer parameter-dependent organization in JEPA features than in reconstruction-based features. For example, in the Burgers dataset, the JEPA representation exhibits two branches associated with different parameter ranges. To assess the physical relevance of these features, we evaluate two complementary probes on frozen representations, as shown in Figure~\ref{fig:teaser}(c). A local-state probe measures how accurately instantaneous physical fields can be recovered from the features. A parameter probe evaluates how well the governing conditions can be inferred from them. Across the evaluated PDEs, JEPA achieves stronger performance on both probing tasks. These results demonstrate that predicting dynamics in latent space encourages the encoder to retain both fine-grained state information and global governing factors that drive physical evolution.


\textbf{Observation 2: Informative representations are not necessarily easy to evolve.}
Physical probing evaluates what can be recovered from representations of observed states. Autoregressive forecasting poses a different challenge: repeatedly evolving predicted states without access to future observations. Figure~\ref{fig:teaser}(d) reveals a mismatch between these two capabilities. Despite its stronger probing performance, the vanilla JEPA-based model produces higher ID rollout errors than the reconstruction-based baseline on both Wave-2D and Vorticity. Under parameter shifts, JEPA achieves lower OOD error on Wave-2D and comparable performance on Vorticity. These results suggest that predictive pretraining provides a physically informative starting point, but its representational advantages alone do not ensure accurate recursive evolution.

Taking these observations together, we define the central challenge as preserving the physical information captured during pretraining (\textbf{observation 1}) while adapting the latent organization to the demands of long-horizon evolution and parameter generalization (\textbf{observation 2}). In this paper, we introduce \textbf{PDE-JEPA}, a framework for learning evolvable state spaces for parametric PDEs. Specifically, we first view predictive pretraining as a foundation to build upon rather than a complete solution to latent dynamics modeling. Based on pretrained representations, we then introduce the Physics-Aligned Latent Geometry (\textbf{PAG}) module, a lightweight residual geometry projector that aligns latent trajectory geometry with the evolution geometry of physical fields to enhance latent evolution while preserving the information encoded by the original representation via an anchor loss. Finally, to further improve generalization to unseen governing conditions, we develop the Physics-Structured Latent Predictor (\textbf{PSP}) that decomposes the dynamics into parameter-independent evolution and parameter-dependent responses. This structure reflects the common form of parametric PDEs, in which governing parameters modulate specific dynamical components, and provides an explicit inductive bias for extrapolation under unseen governing conditions. 

We evaluate PDE-JEPA on 9 parametric PDE benchmarks spanning transport, diffusion, reaction--diffusion, wave propagation, and fluid dynamics. 
Our method achieves the lowest rollout error across most in-distribution benchmarks and all five benchmarks evaluated under out-of-distribution governing conditions. For example, as shown in Figure~\ref{fig:teaser} (e), PDE-JEPA achieves the most improvement on Heat against Poseidon-T for the ID setting.
Further analyses show that these gains are accompanied by more physically aligned latent trajectories and more consistent responses to changes in governing parameters. To sum up, our contributions are listed below:
\begin{itemize}
    \item To our knowledge, we are the first to systematically study JEPA for parametric PDEs and show that informative representations alone do not ensure accurate rollout. We thus propose \textbf{PDE-JEPA}, combining \textbf{PAG} and \textbf{PSP} for accurate forecasting and OOD extrapolation.

\item We propose \textbf{PAG} module, a lightweight geometry projector that aligns latent trajectory geometry with physical evolution to substantially improve rollout accuracy, while preserving the pretrained representation through an anchor loss.

    \item We further introduce the \textbf{PSP}, which incorporates PDE formulation inductive bias by separating parameter-independent evolution from parameter-dependent responses, improving OOD extrapolation.
\end{itemize}


\section{Related Work}
\subsection{Parametric PDE Solvers}

Learning-based PDE solvers broadly span physics-informed methods~\citep{karniadakis2021physics,toscano2025pinns} and data-driven neural operators~\citep{lu2021learning, wu2024transolver, tan2026points}. We focus on the latter for parametric forecasting, with the Fourier Neural Operator (FNO)~\citep{li2020fourier} providing a foundational framework. Parametric generalization has since been explored through explicit parameter conditioning~\citep{brandstetter2022message,takamoto2023learning,cho2024parameterized,berman2024colora,hagnberger2024vectorized}, multi-environment adaptation~\citep{yin2021leads,kirchmeyer2022generalizing,huang2022meta,kassai2024boosting} and in-context operator learning~\citep{yang2023context,yang2024pde,
serrano2024zebra,kassai2026enma,patel2026continuum}. More recently, general-purpose PDE solvers leverage large-scale multi-physics pretraining~\citep{mccabe2024multiple,hao2024dpot,herde2024poseidon,zhou2024unisolver, mccabe2025walrus, wang2026mixture, wu2026geopt}. Complementary work further explores operator decomposition~\citep{gopakumar2026learning} and joint parameter--boundary conditioning~\citep{li2026generalized}. Parametric PDE solvers are not the scope of this paper, we approach parametric forecasting through latent-state dynamics, asking how the learned state space should be structured for evolution and extrapolation across governing conditions.

\subsection{Latent Learning for PDE Dynamics}

Latent PDE models represent physical states in learned latent spaces and model temporal evolution directly in representation space~\citep{benner2015survey,wiewel2019latent,maulik2021reduced,han2022predicting}. A central design choice in this paradigm is how the latent state itself is learned. Many existing approaches obtain latent representations through reconstruction objectives, where the latent variables are optimized to recover the observed physical fields~\citep{chen2022crom,wu2022learning,li2025latent}. Furthermore, reconstruction objectives are often combined with additional losses that encourage latent dynamical predictability~\citep{regazzoni2024learning}. More recent approaches extend latent-space forecasting through autoregressive modeling over quantized or continuous latent representations~\citep{serrano2024zebra,kassai2026enma}, but the learned state space is still reconstruction oriented. In contrast, our encoder is pretrained through masked latent prediction rather than physical-field reconstruction, allowing the state representation to be learned from predictable spatiotemporal structure in representation space. We then explicitly align the latent trajectory geometry for downstream dynamics modeling.

\subsection{Predictive Learning for Physical Systems}

Predictive representation learning provides an alternative to reconstruction-based
self-supervision by learning features from predictable structure rather than directly
recovering observations. Joint-embedding predictive architectures (JEPAs) instantiate
this idea through latent prediction, beginning with I-JEPA~\citep{assran2023self} and
later extending to video with V-JEPA~\citep{bardes2024revisiting,mur2026v}.
Related physical self-supervision has explored Lie-symmetry-based invariance
~\citep{mialon2023self} and masked reconstruction~\citep{zhou2024masked}.
More recently, predictive representations have been shown to encode governing
physical factors more effectively~\citep{qu2026representation} and have been extended
to 3D aerodynamic fields with AeroJEPA~\citep{giral2026aerojepa}.
Separately, latent trajectory geometry has been explicitly shaped to facilitate
downstream dynamics~\citep{wang2026temporal}.
Our work studies whether predictive representations can serve as informative latent states for parametric dynamics, and how their geometry should be adapted for long-horizon evolution and extrapolation.


\section{Preliminaries and Problem Setup}
\subsection{Parametric PDE Dynamics}

We consider time-dependent physical systems governed by a family of PDEs,
\begin{equation}
\frac{\partial \mathbf{u}}{\partial t}
=
\mathcal{F}
\left(
t,\mathbf{x},\mathbf{u},
\nabla \mathbf{u},
\nabla^2 \mathbf{u},
\ldots;
\boldsymbol{\xi}
\right),
\qquad
\mathbf{x}\in\Omega,\;
t\in(0,T],
\end{equation}
subject to
\begin{equation}
\mathcal{B}_{\boldsymbol{\xi}}[\mathbf{u}](t,\mathbf{x})=0,
\qquad
\mathbf{x}\in\partial\Omega,
\qquad
\mathbf{u}(0,\mathbf{x})=\mathbf{u}_0(\mathbf{x}).
\end{equation}
Here, $\mathbf{u}(t,\mathbf{x})\in\mathbb{R}^{C}$ denotes the physical fields on the domain $\Omega$ with $C$ channels and $\mathcal{B}$ denotes a boundary operator. $\boldsymbol{\xi}$ characterizes the governing environment, including PDE coefficients, forcing terms, and boundary conditions. A fixed $\boldsymbol{\xi}$ therefore defines a particular dynamical system, while varying $\boldsymbol{\xi}$ induces a family of related but distinct physical evolutions.

\subsection{Joint-Embedding Predictive Learning}

Joint-Embedding Predictive Architectures (JEPAs) learn representations by predicting missing content in a learned latent space rather than reconstructing the observation itself~\citep{lecun2022path,assran2023self,bardes2024revisiting}. Let $\mathcal{C}$ and $\mathcal{M}$ denote the index sets of visible context tokens and masked target tokens, respectively. Given the visible context $\mathbf{x}_{\mathcal{C}}$ and target mask tokens $\mathbf{m}$, the context encoder $E_{\theta}$ produces context representations, and the predictor $P_{\phi}$ predicts the target representations at positions $i\in\mathcal{M}$ provided by an EMA target encoder $\bar{E}_{\theta}$.

\begin{equation}
\mathcal{L}_{\mathrm{JEPA}}
=
\frac{1}{|\mathcal{M}|}
\sum_{i\in\mathcal{M}}
\left\|
P_{\phi}\!\left(E_{\theta}(\mathbf{x}_{\mathcal{C}}),\mathbf{m}\right)_i
-
\operatorname{sg}\!\left(\bar{E}_{\theta}(\mathbf{x})_i\right)
\right\|_1 .
\end{equation}
By predicting directly in representation space, JEPA encourages latent features to capture predictable spatiotemporal structure. V-JEPAs demonstrate strong motion understanding and temporally consistent representations~\citep{mur2026v}, while recent studies on physical systems show that JEPA features encode governing physical factors effectively~\citep{qu2026representation}. These results motivate us to investigate JEPA as a state representation for parametric physical dynamics.

\begin{figure}[t]
    \centering
    \includegraphics[width=\linewidth]{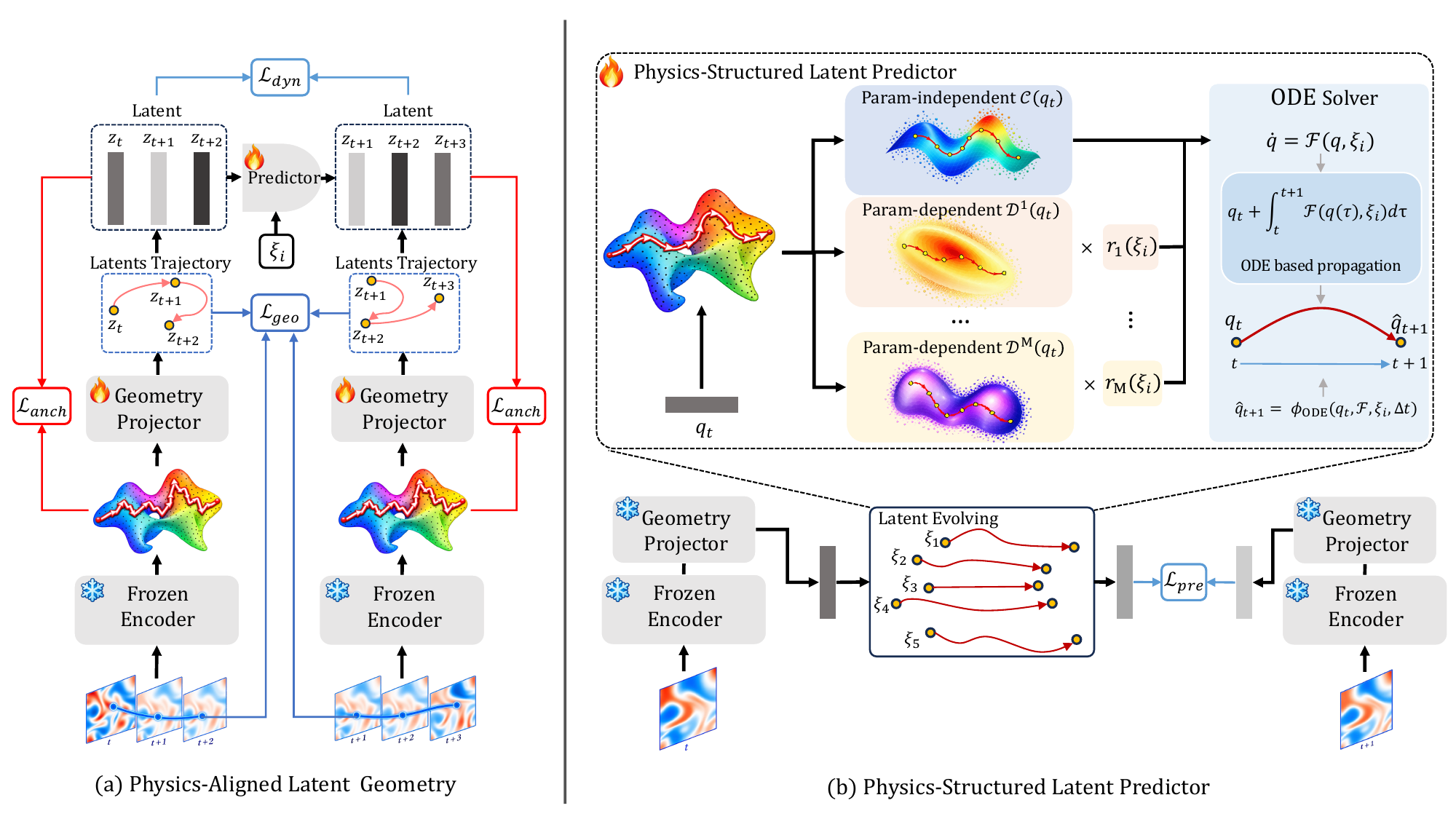}
    \caption{\textbf{Overview of PDE-JEPA.}
(a) We first adapt the frozen predictive representation initialized by JEPA with a light weight geometry projector that aligns latent trajectory geometry with physical evolution. (b) We then evolve the aligned latent states using a physics-structured latent predictor, which decomposes the dynamics into parameter-independent and parameter-dependent components and integrates their combined dynamics with an ODE solver.}
    \label{fig:pipeline}
\end{figure}

\section{Method}

\subsection{Predictive Physical States}
\label{sec:predictive_state}

We adopt the pretrained JEPA representation as the base latent space. Given a physical trajectory
$\mathbf{u}_{0:T}=\{\mathbf{u}_t\}_{t=0}^{T}$, we encode the full state trajectory as
\begin{equation}
    \mathbf{z}_{0:T} = E(\mathbf{u}_{0:T}), \qquad
    \mathbf{u}_{0:T} \in \mathbb{R}^{|\mathcal{X}|\times T\times C},
    \quad
    \mathbf{z}_{0:T} \in \mathbb{R}^{N \times T\times D},
\end{equation}
where $|\mathcal{X}|$ denotes the grid number and
$C$ the number of physical-field channels. The encoder maps $\mathbf{u}_{0:t}$ to $NT$ spatial latent tokens $\mathbf{z}_{0:T}$, each of dimension $D$. Although JEPA is pretrained on full trajectory, downstream states, such as those used for predictor training, are encoded frame-wise following~\citep{mur2026v}, ensuring that $\mathbf{z}_t$ contains no information from future observations.

Importantly, the parameter $\boldsymbol{\xi}$ is provided to neither the encoder nor the JEPA predictor during pretraining, encouraging the encoder to learn parameter-agnostic physical representations. We keep $E$ frozen throughout the subsequent geometry-alignment and dynamics-learning stages, and use the resulting latent trajectory $\{\mathbf{z}_t\}_{t=0}^{T}$ as the input to the geometry projector.

\subsection{Physics-Aligned Latent Geometry}
\label{sec:geometry_alignment}

Although predictive pretraining yields a physically informative state
space, its geometry is not explicitly optimized for temporal evolution.
We observe a clear mismatch between physical and latent trajectories:
latent states exhibit larger turning angles. For example, the mean turning angle increases from
$31.3^\circ$ to $60.5^\circ$ on Vorticity and from $18.2^\circ$ to
$59.4^\circ$ on Burgers.
Such excessive turning produces more zig-zag latent trajectories,
potentially complicating rollout propagation. We therefore align
latent trajectory geometry with physical evolution while preserving the
information already encoded by JEPA. The overview is presented in Figure~\ref{fig:pipeline} (a).

\paragraph{Residual geometry projection.}
Given latents $\mathbf{z}_{0:T}$, we introduce a token-wise projector
\begin{equation}
    \mathbf{q}_t
    =
    \mathbf{z}_t + G_{\phi}(\mathbf{z}_t),
    \qquad
    G_{\phi}(\mathbf{z})
    =
    W_2\sigma\!\left(W_1\mathrm{LN}(\mathbf{z})\right),
\end{equation}
where $\sigma$ is GELU, LN is layernorm, and $W_2$ is zero-initialized so that
$\mathbf{q}_t=\mathbf{z}_t$ initially. It therefore acts as a lightweight
coordinate correction rather than relearning the state representation.

\paragraph{Physical trajectory alignment.}
We align latent trajectory geometry with evolution measured in physical-field space. Define the normalized temporal directions
\begin{equation}
    \mathbf{d}^{u}_t
    =
    \frac{\mathbf{u}_{t+1}-\mathbf{u}_t}
    {\|\mathbf{u}_{t+1}-\mathbf{u}_t\|_2+\epsilon},
    \qquad
    \mathbf{d}^{q}_t
    =
    \frac{\mathbf{q}_{t+1}-\mathbf{q}_t}
    {\|\mathbf{q}_{t+1}-\mathbf{q}_t\|_2+\epsilon}.
\end{equation}
For temporal lag $\ell$, we measure trajectory turning by similarity between two directions
\begin{equation}
    s^{u}_{t,\ell}
    =
    \langle \mathbf{d}^{u}_t,\mathbf{d}^{u}_{t+\ell}\rangle,
    \qquad
    s^{q}_{t,\ell}
    =
    \langle \mathbf{d}^{q}_t,\mathbf{d}^{q}_{t+\ell}\rangle,
\end{equation}
and minimize
\begin{equation}
    \mathcal{L}_{\mathrm{geo}}
    =
    \sum_{\ell\in\{1,2,4\}}
    w_\ell\,
    \operatorname{SmoothL1}
    \left(
    s^{q}_{t,\ell},
    \operatorname{sg}(s^{u}_{t,\ell})
    \right).
\end{equation}
The multi-lag objective captures local and longer-range directional consistency.

\paragraph{Preserving informativeness.}
To prevent geometric alignment from distorting the pretrained state
information, we use an identity anchor
\begin{equation}
    \mathcal{L}_{\mathrm{anchor}}
    =
    \frac{\|\mathbf{q}-\mathbf{z}\|_2^2}
    {\|\mathbf{z}\|_2^2+\epsilon}.
\end{equation}
We further train an auxiliary conditional causal predictor to estimate
$\Delta\mathbf{q}_t=\mathbf{q}_{t+1}-\mathbf{q}_t$,
\begin{equation}
    \mathcal{L}_{dyn}
    =
    \frac{
    \|\widehat{\Delta\mathbf{q}}_t-\Delta\mathbf{q}_t\|_2^2
    }{
    \|\Delta\mathbf{q}_t\|_2^2+\epsilon
    },
\end{equation}
so that the aligned coordinates remain dynamically predictable.
The geometry stage optimizes
\begin{equation}
    \mathcal{L}_{\mathrm{align}}
    =
    \mathcal{L}_{dyn}
    +
    \lambda_{\mathrm{geo}}\mathcal{L}_{\mathrm{geo}}
    +
    \lambda_{\mathrm{anchor}}\mathcal{L}_{\mathrm{anchor}}.
\end{equation}
The JEPA encoder remains frozen. After alignment, we discard the auxiliary
predictor and freeze the projector before training the final
physics-structured dynamics model.

\subsection{Physics-Structured Latent Predictor}
\label{sec:physics_predictor}

The geometry-aligned representation supports temporal evolution, but OOD generalization still requires extrapolating across governing parameters. A standard conditional predictor can inject the parameter directly to the networks, allowing arbitrary state--parameter interactions that may fit the training range well but extrapolate unreliably beyond it. We instead explicitly structure how the governing parameter enters the latent dynamics with formulation inductive bias.
\paragraph{Structured latent vector field.}
We model $\mathbf q(t)$ as a continuous-time
parametric dynamical system,
\begin{equation}
    \frac{d\mathbf q}{dt}
    =
    F_{\theta}\!\left(\mathbf q;\boldsymbol{\xi}\right),
\end{equation}
rather than allowing $\boldsymbol{\xi}$ to interact arbitrarily with
the latent state throughout the dynamics network, we separate a shared
state evolution from a set of parameter-dependent responses:
\begin{equation}
    \frac{d\mathbf q}{dt}
    =
    C_{\theta}(\mathbf q)
    +
    \sum_{j=1}^{M}
    r_j(\boldsymbol{\xi})\,D_{\theta}^{(j)}(\mathbf q).
    \label{eq:structured_latent_ode}
\end{equation}
Here, $C_{\theta}$ captures shared evolution, while $D_{\theta}^{(j)}$ represents a state-dependent
response associated with the $j$-th parameter. $r_j(\boldsymbol{\xi})$ are normalized physical parameters and $M$ is the number of parameters. This decomposition reflects a common structure in parametric PDEs, where
governing coefficients modulate specific components.
For example, the Navier--Stokes equation in vorticity form is
\begin{equation}
    \frac{\partial \omega}{\partial t}
    =
    -(\mathbf{u}\cdot\nabla)\omega
    +
    \nu\nabla^2\omega ,
\end{equation}
where $\nu$ explicitly scales the viscous response. Motivated by this
structure, the Eq.~\ref{eq:structured_latent_ode} becomes
\begin{equation}
    \frac{d\mathbf q}{dt}
    =
    C_{\theta}(\mathbf q)
    +
    r(\nu)D_{\theta}(\mathbf q).
\end{equation}
We do not require either response to recover the exact analytical
PDE operators; the decomposition provides a physics-inspired
inductive bias on how governing parameters enter the latent dynamics. 

\paragraph{Continuous-time propagation.}
We propagate latent between observations using an ODE solver,
\begin{equation}
    \widehat{\mathbf q}_{t+1}
    =
    \Phi_{\mathrm{ODE}}
    \left(
        \mathbf q_t,
        F_{\theta},
        \boldsymbol{\xi},
        \Delta t
    \right),
\end{equation}
where $\Phi_{\mathrm{ODE}}$ integrates the learned vector field over one
observation interval. In practice, we use a fixed-step fourth-order
Runge--Kutta (RK4) solver. Long-horizon predictions are obtained by
recursively integrating the predicted states. We then supervise it with predictor objective
\begin{equation}
    \mathcal L_{\mathrm{pre}}
    =
    \frac{
    \|\widehat{\mathbf q}_{t+1}-\mathbf q_{t+1}\|_2^2
    }{
    \|\Delta\mathbf q_t\|_2^2+\epsilon
    }.
\end{equation}

\section{Experiments}

\begin{table*}[t]
\centering
\caption{
\textbf{ID rollout performance across PDE benchmarks.} All results are reported in Relative $L^2$ error, lower is better. Best results are \textbf{bold} and second-best results are \underline{underlined}.}
\label{tab:id_main}
\small
\setlength{\tabcolsep}{3.7pt}
\renewcommand{\arraystretch}{1.08}

\begin{tabular}{lccccccccc}
\toprule
Method
& Advect
& Burgers
& Heat
& Wave-B
& Combined
& Wave-2D
& Vorticity
& HeterNS
& GS \\
\midrule

\multicolumn{10}{l}{\textit{Parametric Solvers}} \\
\cmidrule(lr){1-10}
FNO
& 0.0390 & 0.4972 & 0.3449 & 0.9819 & 0.0374
& 0.9913 & 0.1411 & 0.0210 & 0.0547 \\

CAPE
& 0.0094 & 0.2230 & 0.2130 & 0.9780 & \underline{0.0085}
& -- & -- & -- & -- \\

CoDA
& \textbf{0.0068} & 0.5460 & 0.7670 & 1.0200 & 0.0120
& 0.7770 & 0.6780 & -- & -- \\

GEPS
& 0.0930 & 0.3989 & 0.6641 & 0.6317 & 0.0097
& 0.5138 & 0.0821 & 0.1032 & 0.0332 \\

\midrule
\multicolumn{10}{l}{\textit{In-Context Solvers}} \\
\cmidrule(lr){1-10}
ViT-\textit{in-context}
& 0.0902 & 0.4720 & 0.5820 & 0.4720 & 0.0885
& 0.3900 & 0.1730 & -- & 0.0690 \\

$[\mathrm{CLS}]$ ViT
& 0.1400 & 0.1360 & 0.1160 & 0.9710 & 0.0446
& 0.2710 & 0.9720 & -- & 0.0480 \\

Zebra
& 0.0079 & 0.1540 & 0.1150 & 0.2450 & 0.0096
& \underline{0.2070} & 0.1190 & -- & 0.0440 \\

\midrule
\multicolumn{10}{l}{\textit{Foundation Models}} \\
\cmidrule(lr){1-10}
UniSolver
& 0.0284 & 0.1838 & 0.1933 & 0.4049 & 0.0087
& 0.4009 & 0.0954 & \underline{0.0098} & \underline{0.0323} \\

MPP
& 0.0312 & 0.4547 & 0.5629 & 0.3856 & 0.0519
& 0.8815 & 0.4509 & 0.0347 & 0.2698 \\

DPOT-S
& 0.0390 & 0.6480 & 0.4638 & 0.4063 & 0.0365
& 0.3152 & 0.1686 & 0.0896 & 0.4159 \\

Poseidon-T
& 0.0203 & 0.1280 & \underline{0.0933} & \underline{0.1093} & 0.0137
& 0.4211 & 0.0679 & 0.1009 & 0.0544 \\

\midrule
\multicolumn{10}{l}{\textit{Latent Solvers}} \\
\cmidrule(lr){1-10}
LE-PDE
& 0.0212 & \underline{0.0869} & 0.1053 & 0.6927 & 0.0299
& 0.5140 & 0.5816 & 0.2380 & 0.1225 \\

LNS
& 0.0207 & 0.0982 & 0.0997 & 0.4002 & 0.0136
& 0.3539 & \underline{0.0592} & 0.0254 & 0.0734 \\

MAE-PDE
& 0.1943 & 0.4467 & 0.3726 & 0.5285 & 0.0667
& 0.7792 & 0.1327 & 0.1798 & 0.0406 \\

ENMA
& 0.0131 & 0.1607 & 0.2712 & 0.6295 & 0.0160
& 0.5285 & 0.2320 & 0.0144 & 0.0607 \\
\midrule
\textbf{Ours}
& \underline{0.0074}
& \textbf{0.0428}
& \textbf{0.0274}
& \textbf{0.0350}
& \textbf{0.0074}
& \textbf{0.1140}
& \textbf{0.0348}
& \textbf{0.0089}
& \textbf{0.0284} \\

\textit{Rel. Impr.}
& $-8.8\%$
& $50.7\%$
& $70.6\%$
& $68.0\%$
& $12.9\%$
& $44.9\%$
& $41.2\%$
& $9.2\%$
& $12.1\%$ \\

\bottomrule
\end{tabular}
\end{table*}

\subsection{Experiments Setup}
\paragraph{Datasets.}
We evaluate on nine parametric PDE benchmarks spanning transport, diffusion, waves, reaction--diffusion, and fluid dynamics. Seven follow Zebra~\citep{serrano2024zebra}: Advection varies the transport speed, Burgers and Heat vary diffusion and forcing, Wave-B varies boundary conditions, Combined varies three differential coefficients, Wave-2D varies wave celerity and damping, and Vorticity varies viscosity. We further include HeterNS from UniSolver~\citep{zhou2024unisolver} and Gray--Scott (GS) from ENMA~\citep{kassai2026enma}. Detailed equations, parameter ranges, and data splits and generation are provided in Appendix~\ref{app:datasets}.

\paragraph{Baselines.}
We compare against a broad set of PDE solvers covering four representative paradigms. \emph{Parametric solvers} include FNO~\citep{li2020fourier}, CAPE~\citep{takamoto2023learning}, CoDA~\citep{kirchmeyer2022generalizing}, and GEPS~\citep{kassai2024boosting}; \emph{in-context solvers} include ViT-\textit{in-context}, $[\mathrm{CLS}]$ ViT~\citep{peebles2023scalable}, and Zebra~\citep{serrano2024zebra}; \emph{foundation models} include UniSolver~\citep{zhou2024unisolver}, MPP~\citep{mccabe2024multiple}, DPOT-S~\citep{hao2024dpot}, and Poseidon-T~\citep{herde2024poseidon}; and \emph{latent solvers} include LE-PDE~\citep{wu2022learning}, LNS~\citep{li2025latent}, MAE-PDE~\citep{zhou2024masked}, and ENMA~\citep{kassai2026enma}. These baselines span direct operator learning, parameter-conditioned adaptation, in-context prediction, large-scale PDE pretraining, and latent-space dynamics modeling.

\paragraph{Metrics.}
We evaluate accuracy using the relative $L^2$ error over the full rollout trajectory, 
\begin{equation}
\mathcal{E}_{\mathrm{rel}}
=
\frac{1}{N_{\mathrm{test}}}
\sum_{j=1}^{N_{\mathrm{test}}}
\frac{\|\hat{\mathbf{u}}_{j}^{\,1:T}-\mathbf{u}_{j}^{\,1:T}\|_2}
{\|\mathbf{u}_{j}^{\,1:T}\|_2},
\end{equation}
where $\hat{\mathbf{u}}^{\,1:T}$ and $\mathbf{u}^{\,1:T}$ denote the predicted and ground-truth trajectories, respectively. Lower values indicate better long-horizon forecasting accuracy.

\subsection{In-Distribution Generalization}
\label{sec:id_generalization}

Table~\ref{tab:id_main} reports in-distribution rollout performance. Our method achieves the lowest relative $L^2$ on eight of nine benchmarks and ranks second on Advection. The gains are particularly large on Burgers, Heat, Wave-B, Wave-2D, and Vorticity. Although the strongest baseline varies across systems---from LE-PDE and Poseidon-T to Zebra and LNS---our method remains consistently strong, suggesting that its benefits extend across diverse dynamical regimes. Advection is the only case where our method is not best ($0.0074$ vs.\ $0.0068$ for CoDA). We attribute this small gap partly to the translation-dominated dynamics, where structures mainly move across the domain rather than change their shape, thus reconstruction ability matters most. Indeed, several methods already achieve errors below $10^{-2}$, indicating a near-saturated regime. In contrast, the larger gains on other non-linear systems suggest that our approach becomes more effective as the underlying dynamics grow more complex.

\begin{table*}[t]
\centering

\begin{minipage}[t]{0.34\textwidth}
\vspace{0pt}
\centering

\captionsetup{font=small,skip=3pt}
\captionof{table}{\textbf{Module ablation on rollout.}}
\label{tab:module_ablation}

\footnotesize
\setlength{\tabcolsep}{3.8pt}
\renewcommand{\arraystretch}{1.08}

\begin{tabular}{lcccc}
\toprule
&
\multicolumn{2}{c}{Vorticity}
&
\multicolumn{2}{c}{Wave-2D}
\\
\cmidrule(lr){2-3}
\cmidrule(lr){4-5}

Model
& ID & OOD
& ID & OOD
\\
\midrule

Vanilla
& .086 & .491 & .363 & .502 \\

+ PAG
& \underline{.040}
& \underline{.397}
& \underline{.143}
& \underline{.321} \\

+ PSP
& \textbf{.034}
& \textbf{.288}
& \textbf{.114}
& \textbf{.157} \\

\bottomrule
\end{tabular}

\end{minipage}
\hfill
\begin{minipage}[t]{0.64\textwidth}
\vspace{0pt}
\centering

\captionsetup{font=small,skip=3pt}
\captionof{table}{
\textbf{Representation geometry before and after alignment.}
}
\label{tab:geo_alignment}

\footnotesize
\setlength{\tabcolsep}{2.6pt}
\renewcommand{\arraystretch}{1.08}

\begin{tabular}{lccc}
\toprule
Dataset
& Angle MAE ($^\circ$)
& Sym. Acc. MAE
& Lag-1 Cos. MAE \\
\midrule

Vorticity
& $29.1\!\rightarrow\!\mathbf{6.3}$ {\scriptsize (-78\%)}
& $.46\!\rightarrow\!\mathbf{.10}$ {\scriptsize (-77\%)}
& $.36\!\rightarrow\!\mathbf{.05}$ {\scriptsize (-84\%)} \\
\addlinespace[1.5pt]

Wave-2D
& $35.5\!\rightarrow\!\mathbf{10.1}$ {\scriptsize (-71\%)}
& $.46\!\rightarrow\!\mathbf{.14}$ {\scriptsize (-68\%)}
& $.50\!\rightarrow\!\mathbf{.14}$ {\scriptsize (-74\%)} \\
\addlinespace[1.5pt]

Burgers
& $41.4\!\rightarrow\!\mathbf{20.8}$ {\scriptsize (-49\%)}
& $.73\!\rightarrow\!\mathbf{.37}$ {\scriptsize (-49\%)}
& $.43\!\rightarrow\!\mathbf{.17}$ {\scriptsize (-59\%)} \\
\addlinespace[1.5pt]

GS
& $28.8\!\rightarrow\!\mathbf{20.4}$ {\scriptsize (-29\%)}
& $.39\!\rightarrow\!\mathbf{.27}$ {\scriptsize (-30\%)}
& $.39\!\rightarrow\!\mathbf{.26}$ {\scriptsize (-32\%)} \\

\bottomrule
\end{tabular}

\end{minipage}

\end{table*}

\subsection{Out-of-Distribution Extrapolation}
\label{sec:ood_generalization}

\begin{wraptable}{r}{0.62\textwidth}
\vspace{-8pt}
\centering
\captionsetup{width=\linewidth}
\caption{\textbf{OOD rollout performance across PDE benchmarks.}
Relative $L^2$ error is reported, lower is better.
Best results are \textbf{bold} and second-best results are \underline{underlined}.}
\label{tab:ood_main}

\footnotesize
\setlength{\tabcolsep}{2.2pt}
\renewcommand{\arraystretch}{1.08}

\begin{tabular}{lccccc}
\toprule
Method
& Combined
& Wave-2D
& Vorticity
& \makecell{HeterNS\\Visc./Force}
& GS \\
\midrule

UniSolver
& \underline{0.038}
& 1.003
& 0.923
& \underline{0.037}/\underline{0.105}
& 0.1636 \\

Poseidon-T
& 0.146
& 1.511
& 0.665
& 0.560/0.821
& \underline{0.083} \\

LNS
& 0.1667
& \underline{0.610}
& 0.481
& 0.610/0.932
& 0.146 \\

ENMA
& 0.243
& 1.151
& 0.467
& 1.501/2.271
& 0.134 \\


\makecell[l]{Zebra}
& --
& 0.680
& \underline{0.320}
& --
& -- \\

\midrule
\textbf{Ours}
& \textbf{0.008}
& \textbf{0.157}
& \textbf{0.288}
& \textbf{0.011}/\textbf{0.103}
& \textbf{0.033} \\

\textit{Rel. Impr.}
& 77.9\%
& 74.2\%
& 9.7\%
& 69.5\%/1.5\%
& 59.7\% \\

\bottomrule
\end{tabular}

\vspace{-7pt}
\end{wraptable}

We next evaluate extrapolation to governing conditions outside the training distribution. Due to space constraints, Table~\ref{tab:ood_main} reports a representative subset of baselines, with the full comparison provided in the Appendix~\ref{app:full_ood}. Our method achieves the lowest error on all five benchmarks. The gains are particularly large on Combined, Wave-2D, and GS, improving over the strongest baselines by $77.9\%$, $74.2\%$, and $59.7\%$, respectively. Moreover, the ID-to-OOD degradation remains limited on these systems, increasing only from $0.0074$ to $0.0084$, $0.1140$ to $0.157$, and $0.0284$ to $0.0337$, respectively, indicating strong extrapolation beyond the training regimes.


\subsection{Effect of PAG}
\label{sec:geo_effect}

We first examine whether PAG facilitates long-horizon evolution. As shown in Table~\ref{tab:module_ablation}, adding the PAG reduces Vorticity error from $0.086$ to $0.040$ on ID data and from $0.491$ to $0.397$ on OOD data, corresponding to $53.4\%$ and $19.1\%$ improvements. To characterize this geometric change, we compare the original representation $\mathbf{z}$ and aligned representation $\mathbf{q}$ with the physical trajectory using \emph{Angle MAE} for local turning angles, \emph{Sym. Acc. MAE} for second-order temporal variation, and \emph{Lag-1 Cos. MAE} for directional consistency between consecutive increments. As shown in Table~\ref{tab:geo_alignment}, alignment reduces all three errors, with particularly large reductions on Vorticity ($78\%/77\%/84\%$) and Wave-2D ($71\%/68\%/74\%$), while Burgers and GS show smaller but consistent improvements. Since these metrics are computed directly on $\mathbf{z}$ and $\mathbf{q}$ without a dynamics predictor, they confirm that the projector itself aligns latent trajectory geometry more closely with physical-field evolution. More analyses are in Appendix~\ref{app:mechanisms}


\begin{figure}[t]
    \centering
    \includegraphics[width=\columnwidth]
    {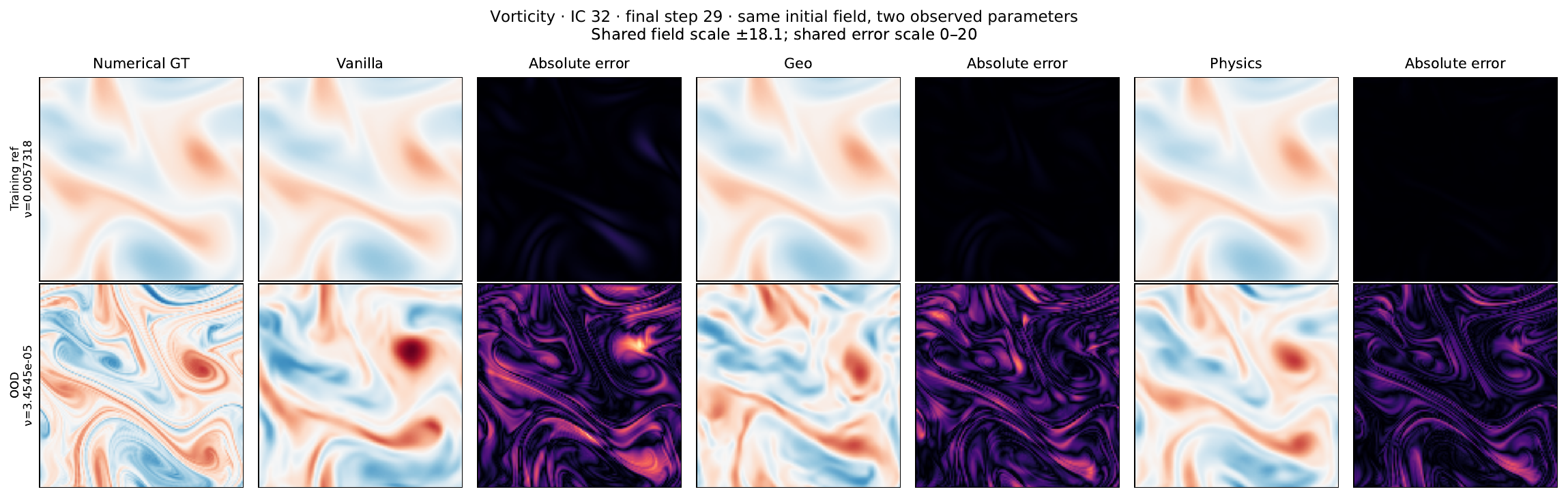}
    \caption{
    \textbf{Visualization on Vorticity under matched initial
    conditions on final timestep.} Prediction errors are shown for
    the vanilla, geometry-aligned and physics-structured models. More visualization can be found in Appendix~\ref{app:tra_vis}.
    }
    \label{fig:param_response_vorticity}
\end{figure}

\subsection{Parameter-Response Mechanism for PSP}
\label{sec:param_response}
\begin{wraptable}{r}{0.60\textwidth}
\vspace{-1.0\baselineskip}   
\centering
\captionsetup{width=\linewidth,font=small}
\caption{
\textbf{Parameter-response consistency under matched ICs.}}
\label{tab:param_response}

\footnotesize
\setlength{\tabcolsep}{2.0pt}
\renewcommand{\arraystretch}{1.08}

\resizebox{\linewidth}{!}{
\begin{tabular}{llcccc}
\toprule
& &
\multicolumn{2}{c}{Field}
&
\multicolumn{2}{c}{Latent}
\\
\cmidrule(lr){3-4}
\cmidrule(lr){5-6}

Dataset & Model
& Cos. $\uparrow$
& Amp. $\rightarrow 1$
& Cos. $\uparrow$
& Amp. $\rightarrow 1$
\\
\midrule

Vorticity
& PAG
& .626 & .846
& .913 & 1.142 \\

& + PSP
& \textbf{.708}
& \textbf{.987}
& \textbf{.924}
& \textbf{1.043} \\

\addlinespace[1.5pt]

Wave-2D
& PAG
& .958 & .979
& .939 & .990 \\

& + PSP
& \textbf{.986}
& \textbf{.996}
& \textbf{.947}
& \textbf{1.002} \\

\bottomrule
\end{tabular}
}
\vspace{-7pt}
\end{wraptable}

We then isolate the effect of the physics-structured predictor on top of the geometry-aligned representation. As shown in Table~\ref{tab:module_ablation}, adding the PSP further reduces OOD rollout error from $.397$ to $.288$ on Vorticity and from $.321$ to $.157$ on Wave-2D. To understand this gain, Table~\ref{tab:param_response} evaluates parameter-conditioned responses under matched initial conditions (ICs). The structured predictor consistently improves parameter-conditioned responses in both field and latent spaces. Here, Cos. measures the directional agreement between the predicted and reference parameter-induced changes, while Amp. measures whether the response magnitude is correct, with values closer to $1$ indicating better agreement. Figure~\ref{fig:param_response_vorticity} provides complementary qualitative evidence: when only the governing parameter is changed, the structured predictor more faithfully follows the corresponding numerical solution, particularly under the OOD condition. More analyses are in Appendix~\ref{app:mechanisms}.


\section{Conclusion}
In this paper, we study how learned state-space structure affects parametric PDE dynamics.
Predictive JEPA pretraining yields physically informative representations,
while explicit trajectory alignment further improves long-horizon evolution.
Building on this, PDE-JEPA combines physics-aligned latent geometry with a
structured predictor that separates shared evolution from
parameter-dependent responses. Across nine PDE benchmarks, it achieves
strong ID performance and substantially improves extrapolation to unseen
governing conditions. Further analyses show better physical trajectory
alignment and more consistent parameter-conditioned responses. Overall,
generalizable physical forecasting depends not only on the evolution model,
but also on learning a state-space geometry suited for evolution and
extrapolation.

\subsection*{AI use statement}

Generative AI was employed only as an auxiliary tool during the development
of this work. Its use was limited to tasks such as improving written
presentation, reorganizing portions of the manuscript, assisting with draft
preparation, and supporting code implementation and troubleshooting. The
scientific content of the paper, including the derivation and presentation of equations, the
analysis of experimental results, and the interpretation of the findings,
was produced and determined by the authors. No generative AI
system was used to create experimental observations, determine reported
results, or make scientific conclusions on behalf of the authors. Any code produced with AI assistance was manually inspected, executed, and
validated against the intended implementation and experimental behavior.
Similarly, AI-assisted prose was subsequently edited by the authors, and all
numerical values reported in the manuscript were verified against the
corresponding experimental records. The authors remain fully responsible for
the accuracy, validity, and integrity of all material presented in this work.

\bibliography{references}
\bibliographystyle{arxiv}

\newpage
\appendix
\clearpage

\etocsetlocaltop.toc{part}
\etocsettocdepth{subsubsection}
\etocsettocstyle
    {\section*{Appendix Contents}}
    {}
\localtableofcontents
\newpage
\section{Dataset details}
\label{app:datasets}

We consider nine PDE datasets: Advection, Burgers, Heat, Wave-B, Combined Equation, Vorticity, HeterNS, Wave-2D, and Gray--Scott. Advection, Wave-B, Combined Equation, Wave-2D, and Vorticity are followed from Zebra~\citep{serrano2024zebra}. Burgers and Heat are generated following MP-PDE~\citep{brandstetter2022message}, with an additional forcing coefficient, while all other settings are kept consistent with Zebra. Gray--Scott is adopted from ENMA~\citep{kassai2026enma}, and HeterNS from Unisolver~\citep{zhou2024unisolver}. A summary of the dataset configurations is provided in Table~\ref{tab:dataset_inventory} and ~\ref{tab:id_ood_ranges}, with further details given below.

\begin{table*}[ht]
\centering
\caption{\textbf{Datasets overview.} Counts denote trajectory number. A dash means that no separate OOD
split is specified for the audited dataset version. One-dimensional shapes
use $(T,X)$; two-dimensional shapes use $(C,X,Y,T)$.}
\label{tab:dataset_inventory}
\scriptsize
\renewcommand{\arraystretch}{1.12}

\begin{tabularx}{\textwidth}{
@{}
l
c
>{\centering\arraybackslash}X
>{\centering\arraybackslash}X
>{\centering\arraybackslash}X
>{\centering\arraybackslash}X
@{}
}
\toprule
\textbf{Dataset} & \textbf{Shape per trajectory} & \textbf{Train}
& \textbf{Val.} & \textbf{Test} & \textbf{OOD} \\
\midrule

\textbf{Advection}
& $140\times256$ & 12,000 & 120 & 120 & -- \\

\textbf{Burgers}
& $250\times256$ & 12,000 & 120 & 120 & -- \\

\textbf{Heat}
& $250\times256$ & 12,000 & 120 & 120 & -- \\

\textbf{Wave-B}
& $250\times256$ & 12,000 & 120 & 120 & -- \\

\textbf{Combined Equation}
& $140\times256$ & 12,000 & 120 & 120 & 120 \\

\textbf{Vorticity}
& $1\times128\times128\times30$
& 12,000 & 1,200 & 1,200 & 120 \\

\textbf{Wave-2D}
& $2\times64\times64\times30$
& 12,000 & 1,200 & 1,200 & 120 \\

\textbf{Gray--Scott}
& $2\times32\times32\times20$
& 12,000 & 1,200 & 1,200 & 120 \\

\textbf{HeterNS}
& $1\times64\times64\times20$
& 15,000 & 1,500 & 1,500 & 1,500 \\
\bottomrule
\end{tabularx}
\end{table*}

\begin{table*}[ht]
\centering
\caption{\textbf{In-distribution (In-D) and out-of-distribution (Out-D)
parameter settings for each dataset.}}
\label{tab:id_ood_ranges}
\scriptsize
\renewcommand{\arraystretch}{1.15}

\begin{tabularx}{\textwidth}{
@{}
l
c
>{\centering\arraybackslash}X
>{\centering\arraybackslash}X
@{}
}
\toprule
\textbf{Dataset}
& \textbf{Parameter}
& \textbf{In-D}
& \textbf{Out-D} \\
\midrule

\multirow{3}{*}{\textbf{Combined}}
& $\alpha$
& $\mathcal{U}([0,1])$
& $[1.0682,1.7581]$ \\

& $\beta$
& $\mathcal{U}([0,0.4])$
& same as In-D \\

& $\gamma$
& $\mathcal{U}([0,1])$
& same as In-D \\
\midrule

\textbf{Vorticity}
& $\nu$
& $[10^{-3},10^{-2}]$
& $[10^{-5},10^{-4}]$ \\
\midrule

\multirow{2}{*}{\textbf{HeterNS}}
& $\nu$
& $\{10^{-5},5{\times}10^{-5},10^{-4},
5{\times}10^{-4},10^{-3}\}$
& $\begin{gathered}
\text{Interp.: }(\{2,3,4,6,7,8,9\}{\times}10^{-5})\\
{}\cup(\{2,3,4,6,7,8,9\}{\times}10^{-4})\\
\text{Extra.: }\{2,3,4,5,6,7,8,9\}{\times}10^{-3}
\end{gathered}$ \\

& $m$
& $\{1,2,3\}$
& $\begin{gathered}
\{1,2,3\}\;(\text{viscosity OOD})\\
\{0.5,1.5,2.5,3.5\}\;(\text{forcing OOD})
\end{gathered}$ \\
\midrule

\multirow{2}{*}{\textbf{Wave-2D}}
& $c$
& $[100,\;500]$
& $[500,\;550]$ \\

& $k$
& $[0,\;50]$
& $[50,\;60]$ \\
\midrule

\multirow{2}{*}{\textbf{Gray--Scott}}
& $F$
& $\mathcal{U}([0.023,0.045])$
& $\mathcal{U}([0.045,0.0467])$ \\

& $k$
& $\mathcal{U}([0.0590,0.0640])$
& $\mathcal{U}([0.0570,0.0590])$ \\
\bottomrule
\end{tabularx}
\end{table*}

\subsection{Advection}
\label{app:advection}
\lead{Equation and environments.}
We solve the constant-speed transport equation
\begin{equation}
  \partial_t u + \beta\,\partial_x u=0,
  \qquad x\in[0,L),\qquad L=128,
\end{equation}
with periodic boundaries. The sole environment parameter is the speed
$\beta\sim\Unif[0,4]$. We draw 1,200 training environments and 12 environments
for each of validation and testing, with ten independently seeded initial
conditions per environment.

\lead{Initial conditions.}
Each trajectory starts from a random superposition of three cosine modes,
\begin{equation}
\begin{split}
 &u_0(x)=\sum_{j=1}^{3} A_j
 \cos\!\left(\frac{2\pi\ell_j x}{L}+\phi_j\right),
 \\
 A_j\sim\Unif[-0.5,&0.5],\quad
 \phi_j\sim\Unif[0,2\pi],\quad
 \ell_j\sim\Unif\{1,2,3,4,5\}.
\end{split}
\end{equation}

The mode coefficients are sampled independently for each initial condition.

\begin{figure}[H]
    \centering
    \includegraphics[width=\linewidth]{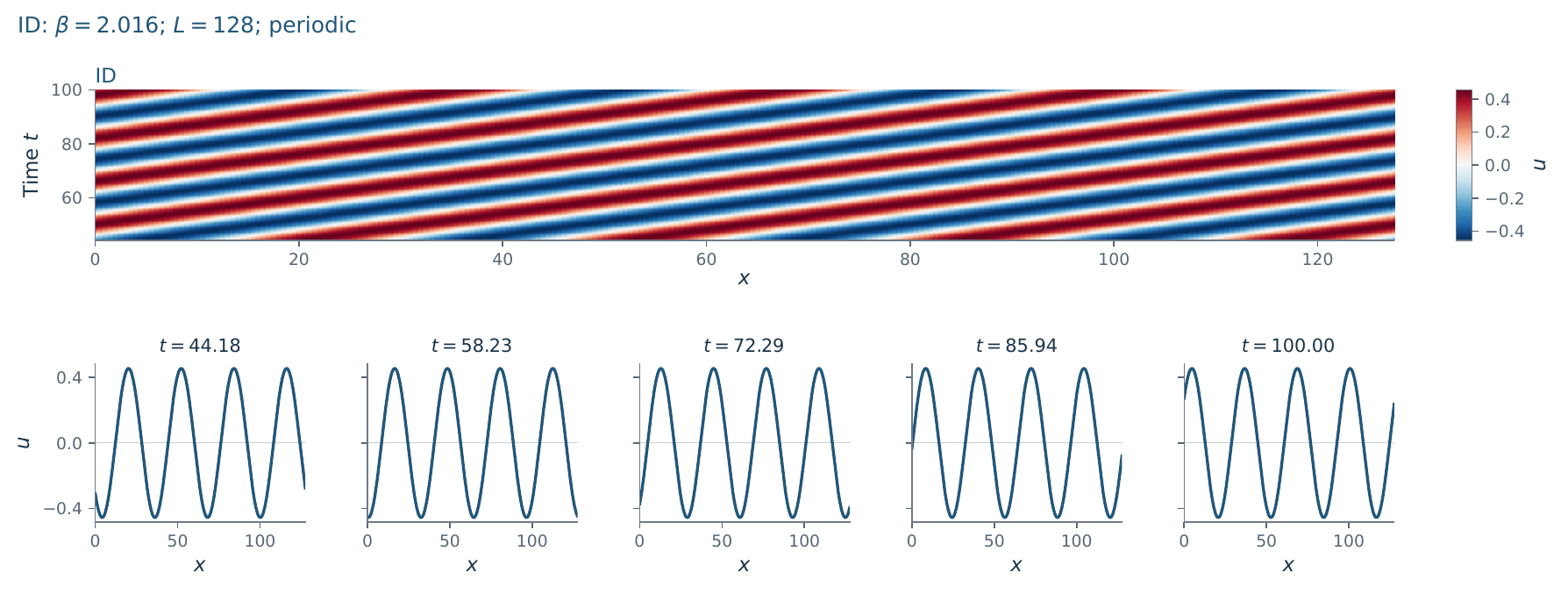}
    \caption{\textbf{Advection.}
Transport speed $\beta=2.01597$, within the training support $[0,4]$,
with periodic boundaries on a domain of length 128. The stored
segment spans $t\simeq44.18$ to $100$.}
    \label{fig:advect}
\end{figure}

\lead{Numerical generation.}
We use the adapter provided by ENMA to generate trajectories analytically via periodic translation,
\[
u(x,t)=u_0\left((x-\beta t)\bmod L\right),
\]
on a 256-point spatial grid. The solution is evaluated at 250 uniformly spaced times over $t\in[0,100]$, with the last 140 frames retained, corresponding approximately to $t\in[44.18,100]$. The visualization is shown in Figure~\ref{fig:advect}.

\subsection{Burgers equation}
\label{app:burgers}
\lead{Equation and environments.}
On a periodic interval of length $L=16$, we solve
\begin{equation}
 \partial_t u+\partial_x\!\left(\tfrac12u^2-\beta\,\partial_xu\right)
 = F(t,x),
 \qquad \beta\sim\LogUnif[10^{-3},5],
 \label{eq:burgers_dataset}
\end{equation}
where the forcing is
\begin{align}
 F(t,x)&=\sum_{j=1}^{5} A_j
 \sin\!\left(\omega_j t+\frac{2\pi\ell_jx}{L}+\phi_j\right),
 \label{eq:forcing_dataset}\\
 A_j\sim\Unif[-0.5,0.5],\quad
 \omega_j&\sim\Unif[-0.4,0.4],\quad
 \ell_j\sim\Unif\{1,2,3\},\quad
 \phi_j\sim\Unif[0,2\pi].\nonumber
\end{align}
Here log-uniform sampling means that $\log\beta$ is uniform between the
logarithms of the two endpoints. Diffusivity and all 20 forcing coefficients
are fixed within an environment and vary between environments.

\lead{Initial conditions.}
For each trajectory, we independently draw
\begin{equation}
 u_0(x)=\sum_{j=1}^{5}\widetilde A_j
 \sin\!\left(\frac{2\pi\widetilde\ell_jx}{L}+\widetilde\phi_j\right),
 \label{eq:forced_ic}
\end{equation}
using the amplitude, integer mode, and phase distributions in
Eq.~\eqref{eq:forcing_dataset}. These initial-condition coefficients are
independent of the environment's forcing coefficients. The split contains
1,200/12/12 environments for training/validation/testing, with ten trajectories
per environment. All three splits use the same
parameter support.

\begin{figure}[H]
    \centering
    \includegraphics[width=\linewidth]{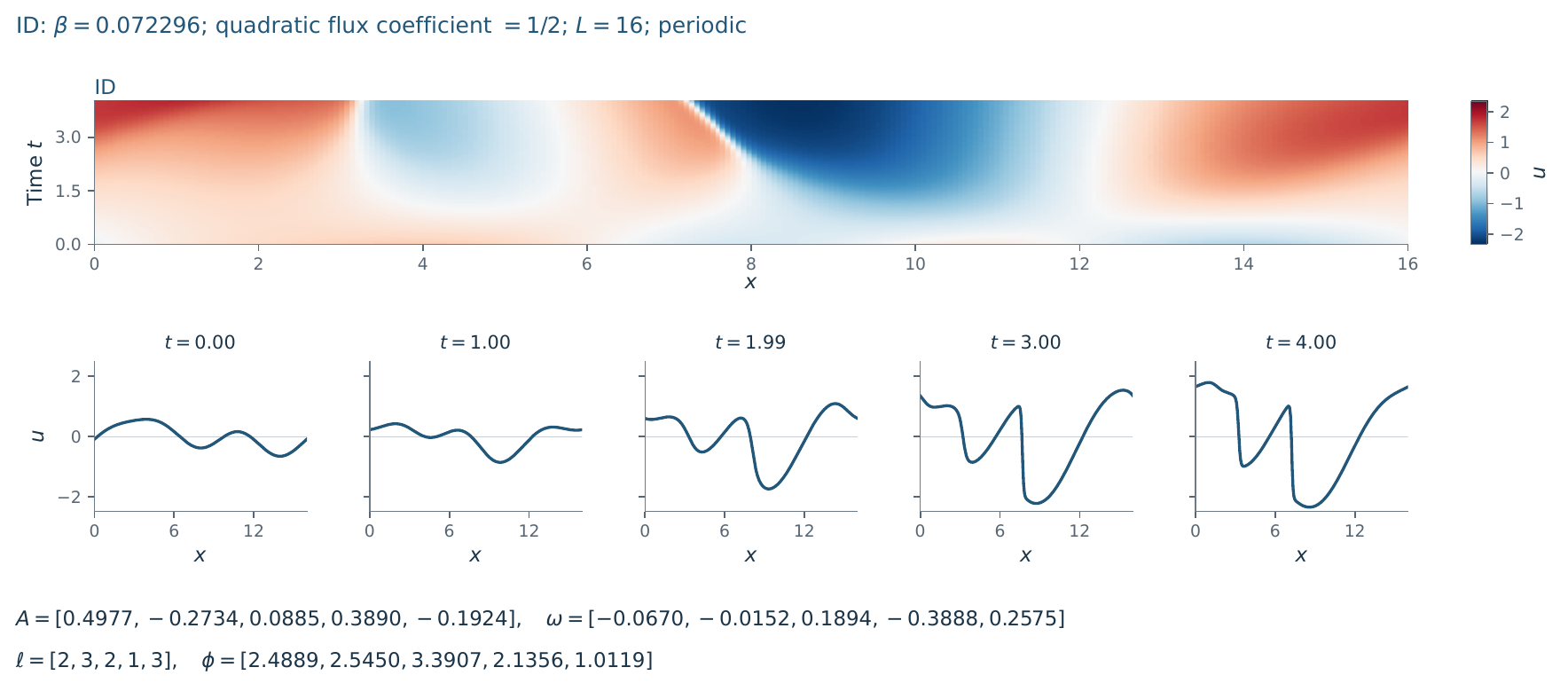}
    \caption{\textbf{Burgers.}
Viscosity $\beta=0.072296$, within the training support $[10^{-3},5]$,
with periodic boundaries on a domain of length 16. The forcing is
$F(t,x)=\sum_{j=1}^{5}A_j\sin(\omega_jt+2\pi\ell_jx/16+\phi_j)$;
its coefficients $A_j$, $\omega_j$, $\ell_j$, and $\phi_j$ are listed
beneath.}
    \label{fig:burgers}
\end{figure}

\lead{Numerical generation.}
The local MP-PDE implementation uses WENO reconstruction~\citep{jiang1996efficient} with Godunov flux splitting~\citep{godounov1959difference} for the nonlinear term, finite differences for diffusion, and an adaptive Dormand--Prince 4/5 integrator~\citep{dormand1980family} with tolerances of $10^{-5}$. Trajectories are generated on 256 spatial points over $t\in[0,4]$, with 250 snapshots evaluated and every tenth frame retained to form 25-frame sequences. We follow the implementation-specific spatial discretization and forcing range $[-0.4,0.4]$ used by the released generator. The visualization is shown in Figure~\ref{fig:burgers}.

\subsection{Heat equation}
\label{app:heat}

\lead{Equation and environments.}
We solve
\begin{equation}
 \partial_tu=\beta\,\partial_{xx}u+F(t,x),
 \qquad x\in[0,16],\qquad \beta\sim\LogUnif[10^{-3},5],
\end{equation}
with periodic differentiation. The forcing is the five-mode process in
Eq.~\eqref{eq:forcing_dataset}; 

\lead{Initial conditions.} Initial conditions are independently sampled
from Eq.~\eqref{eq:forced_ic}. Following Burgers, one environment fixes the
diffusivity and the complete forcing realization, while its ten trajectories
have independent initial conditions.

\begin{figure}[H]
    \centering
    \includegraphics[width=\linewidth]{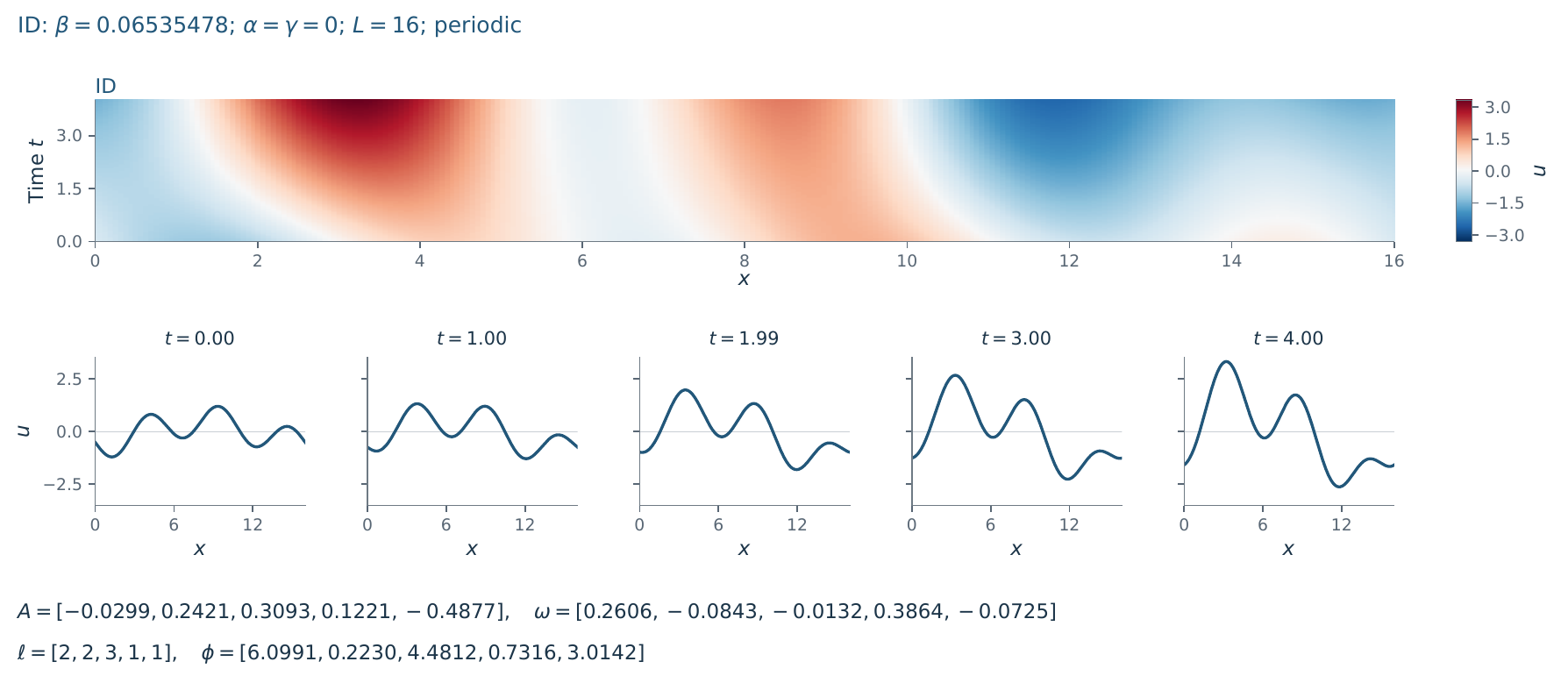}
    \caption{\textbf{Heat.}
Diffusivity $\beta=0.0653548$, within the training support $[10^{-3},5]$.
The trajectory solves $\partial_tu=\beta\partial_{xx}u+F(t,x)$ on a
periodic interval of length 16, with the five-mode forcing coefficients
shown beneath.}
    \label{fig:heat}
\end{figure}

\lead{Numerical generation.} We generate the heat trajectories using the local MP-PDE implementation with the nonlinear and dispersive terms disabled, leaving a finite-difference diffusion operator and an adaptive Dormand--Prince 4/5 integrator~\citep{dormand1980family} with tolerances of $10^{-5}$. Each trajectory is evaluated on 256 spatial points at 250 uniformly spaced times over $t\in[0,4]$. Every tenth frame is retained to form 25-frame sequences; we use 1,200/12/12 train/validation/test environments with ten trajectories per environment. The visualization is shown in Figure~\ref{fig:heat}.

\subsection{Wave-B: one-dimensional waves with varying boundaries}
\label{app:waveb}
\lead{Equation and environments.}
We consider
\begin{equation}
 \partial_{tt}u-c^2\partial_{xx}u=0,
 \qquad x\in[-8,8],\quad c=2.
\end{equation}
Each endpoint independently uses either homogeneous Dirichlet conditions
($u=0$, denoted D) or homogeneous Neumann conditions ($\partial_xu=0$,
denoted N). This defines four environments: DD, DN, ND, and NN. Wave speed is fixed and is not an
environment variable.

\lead{Initial conditions.}
For each trajectory, a pulse center $s\sim\Unif[-4,4]$ determines both initial
displacement and initial velocity:
\begin{equation}
 u(x,0)=\exp[-(x-s)^2],\qquad
 \partial_tu(x,0)=-2c(x-s)\exp[-(x-s)^2].
\end{equation}
The four environments each contain 3,000 training trajectories, 30 validation
trajectories, and 30 test trajectories.
All four boundary combinations occur in every split: this split tests
generalization to new initial conditions, not unseen boundary types.

\begin{figure}[H]
    \centering
    \includegraphics[width=\linewidth]{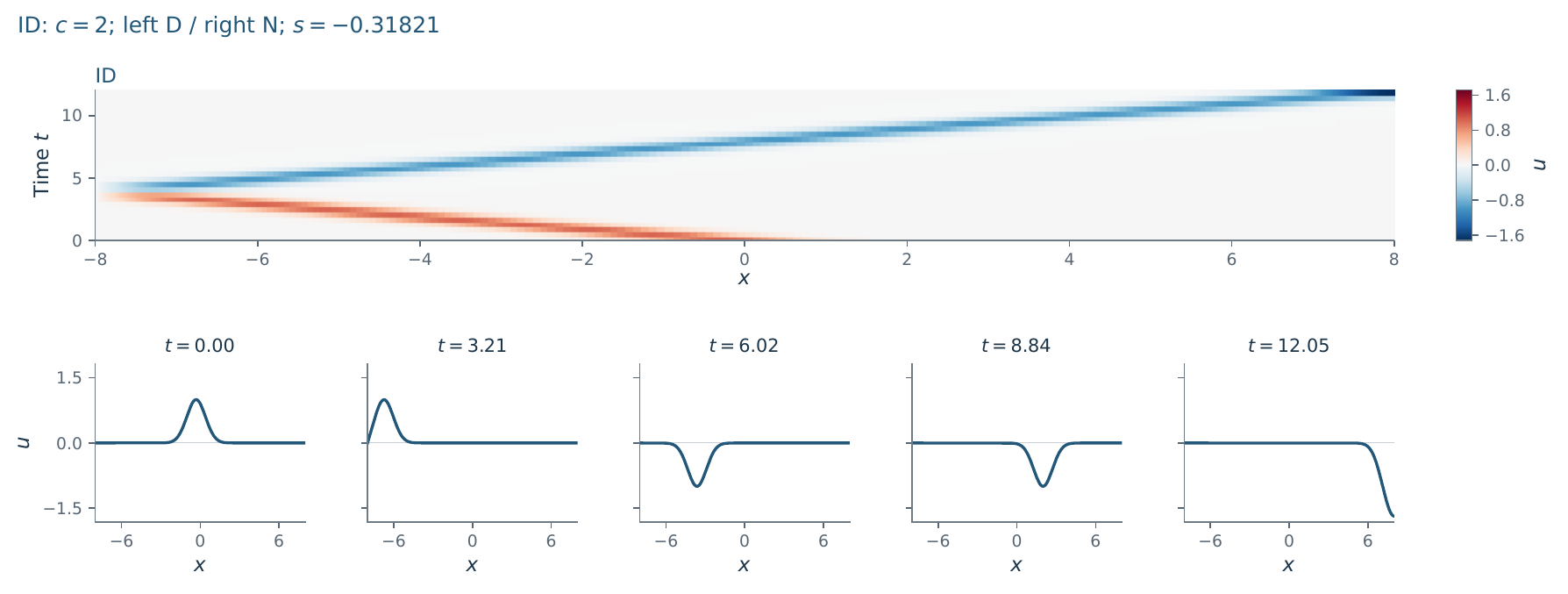}
    \caption{\textbf{Wave-B.}
Wave speed $c=2$, a left Dirichlet and a right Neumann boundary, and
initial pulse center $s=-0.31821$. The initial displacement and velocity
are $u_0(x)=e^{-(x-s)^2}$ and $v_0(x)=-2c(x-s)u_0(x)$. The plot shows $t\simeq0$--$12.05$ after restoring
forward time from the reversed storage order, retaining the original
nonuniform Chebyshev grid.}
    \label{fig:waveb}
\end{figure}

\lead{Numerical generation.}
We generate the trajectories using the MP-PDE Chebyshev differentiation operator~\citep{trefethen2000spectral} on a 256-point nonuniform grid, with an implicit Radau integrator~\citep{wanner1996solving} over $t\in[0,100]$ and tolerances of $10^{-3}$. Each trajectory is evaluated at 250 output times. For model input, the sequence is reordered into forward physical time and every tenth frame is retained, yielding 25-frame trajectories. The visualization is shown in Figure~\ref{fig:waveb}.

\subsection{Combined Equation}
\label{app:combined}

\lead{Equation and environments.}
The Combined Equation follows the setting of Zebra, which adopts the benchmark
introduced by MP-PDE et al.~\citep{brandstetter2022message} without the forcing
term. The dynamics are governed by
\begin{equation}
\partial_t u +
\partial_x\left(
\alpha u^2
-\beta \partial_x u
+\gamma \partial_{xx}u
\right)=0,
\label{eq:combined}
\end{equation}
with initial conditions given by random finite sums of sinusoidal modes,
\begin{equation}
u_0(x)=\sum_{j=1}^{J}
A_j\sin\left(\frac{2\pi \ell_j x}{L}+\phi_j\right).
\end{equation}

For training, 1,200 parameter triplets are sampled uniformly from
$\alpha\in[0,1]$, $\beta\in[0,0.4]$, and $\gamma\in[0,1]$, with ten
trajectories generated for each parameter setting, yielding 12,000 training
trajectories. An additional 120 trajectories are used for testing. Following
Zebra, solutions are generated using the MP-PDE solver~\citep{brandstetter2022message}
on 256 spatial points over $t\in[0,10]$, with 140 temporal snapshots.
Temporal downsampling by a factor of ten gives trajectories of shape
$256\times14$.

\begin{figure}[H]
    \centering
    \includegraphics[width=\linewidth]{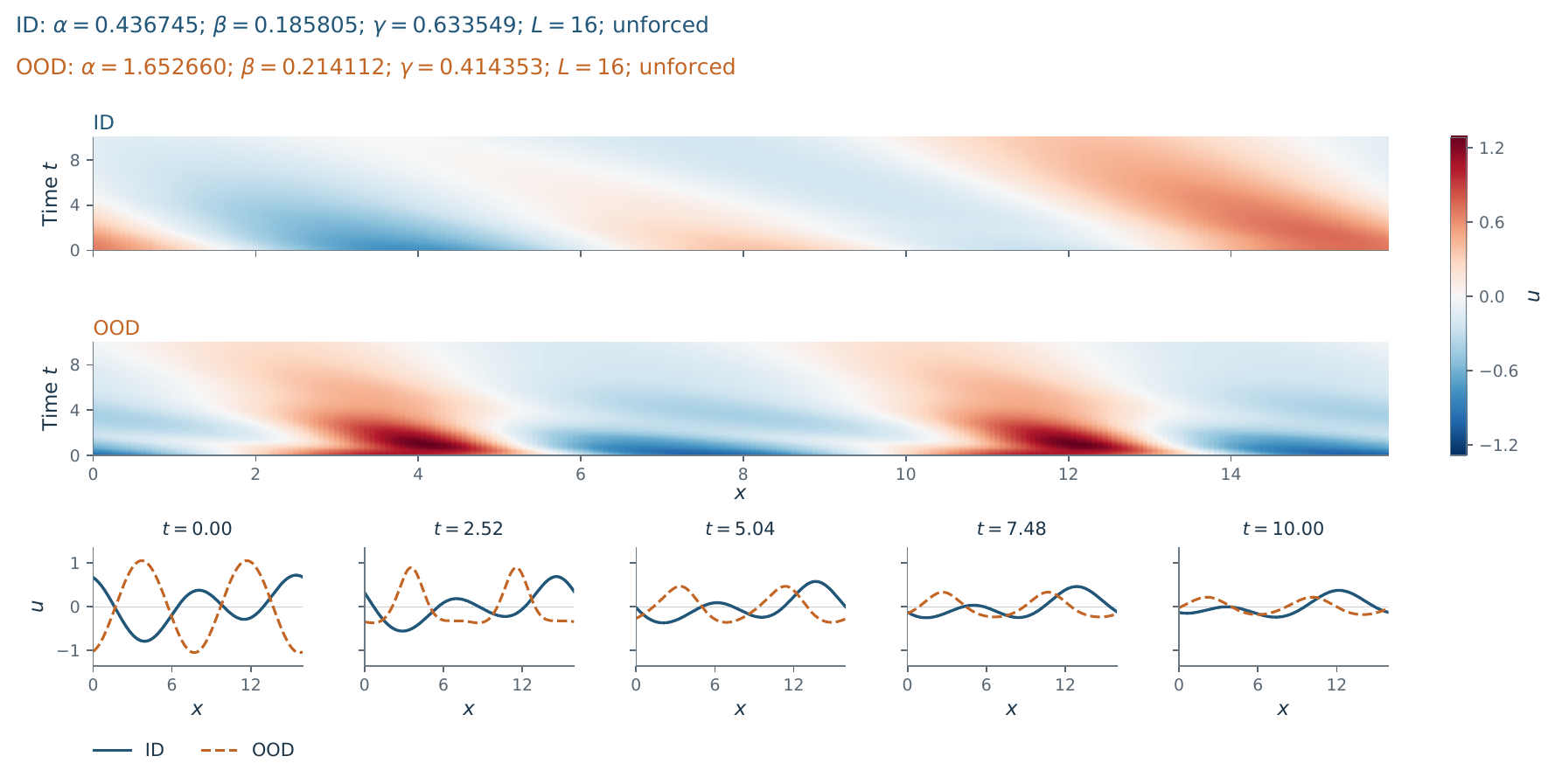}
    \caption{\textbf{Combined Equation.}
ID: $(\alpha,\beta,\gamma)=(0.436745,0.185805,0.633549)$;
OOD: $(\alpha,\beta,\gamma)=(1.652660,0.214112,0.414353)$.
The OOD nonlinear-transport coefficient exceeds the ID interval $[0,1]$,
while $\beta$ and $\gamma$ remain within their ID ranges
$[0,0.4]$ and $[0,1]$. Both trajectories use
periodic boundaries, no external forcing, and the same saved coordinates
on $x\in[0,16)$ and $t\in[0,10]$. Their initial fields and all three
coefficients differ, so every trajectories have different initial conditions.}
    \label{fig:combine}
\end{figure}

\lead{OOD data.}
The OOD split contains 120 trajectories with $\alpha\in[1.0682,1.7581]$,
extending beyond the ID range $\alpha\in[0,1]$, while $\beta$ and $\gamma$
remain within their respective ID ranges. Separate OOD training and validation
splits contain ten trajectories each and are excluded from the ID data. The visualization is shown in Figure~\ref{fig:combine}.

\subsection{Vorticity: two-dimensional incompressible flow}
\label{app:vorticity}
\lead{Equation and environments.}
The dataset contains unforced two-dimensional incompressible flow in
vorticity form on a periodic square:
\begin{equation}
 \partial_t\omega+(\mathbf u\!\cdot\!\nabla)\omega
 =\nu\Delta\omega,\qquad \nabla\!\cdot\!\mathbf u=0,
\end{equation}
where velocity is recovered from a streamfunction through a Poisson solve.
Viscosity $\nu$ is the environment parameter. The ID support is
$[10^{-3},10^{-2}]$, with 1,200 training and 120 validation/test environments
per split, each containing ten trajectories. The OOD release contains 12
viscosities on a linear grid spanning $[10^{-5},10^{-4}]$, again with ten
trajectories per viscosity.

\lead{Initial conditions.}
The initial vorticity fields are generated from the prescribed energy spectrum
\begin{equation}
E(k)=\frac{4}{3}\sqrt{\pi}
\left(\frac{k}{k_0}\right)^4
\frac{1}{k_0}
\exp\left[-\left(\frac{k}{k_0}\right)^2\right],
\end{equation}
with the corresponding vorticity spectrum
\begin{equation}
\omega(k)=\sqrt{\frac{E(k)}{\pi k}}.
\end{equation}

\begin{figure}[H]
    \centering
    \includegraphics[width=\linewidth]{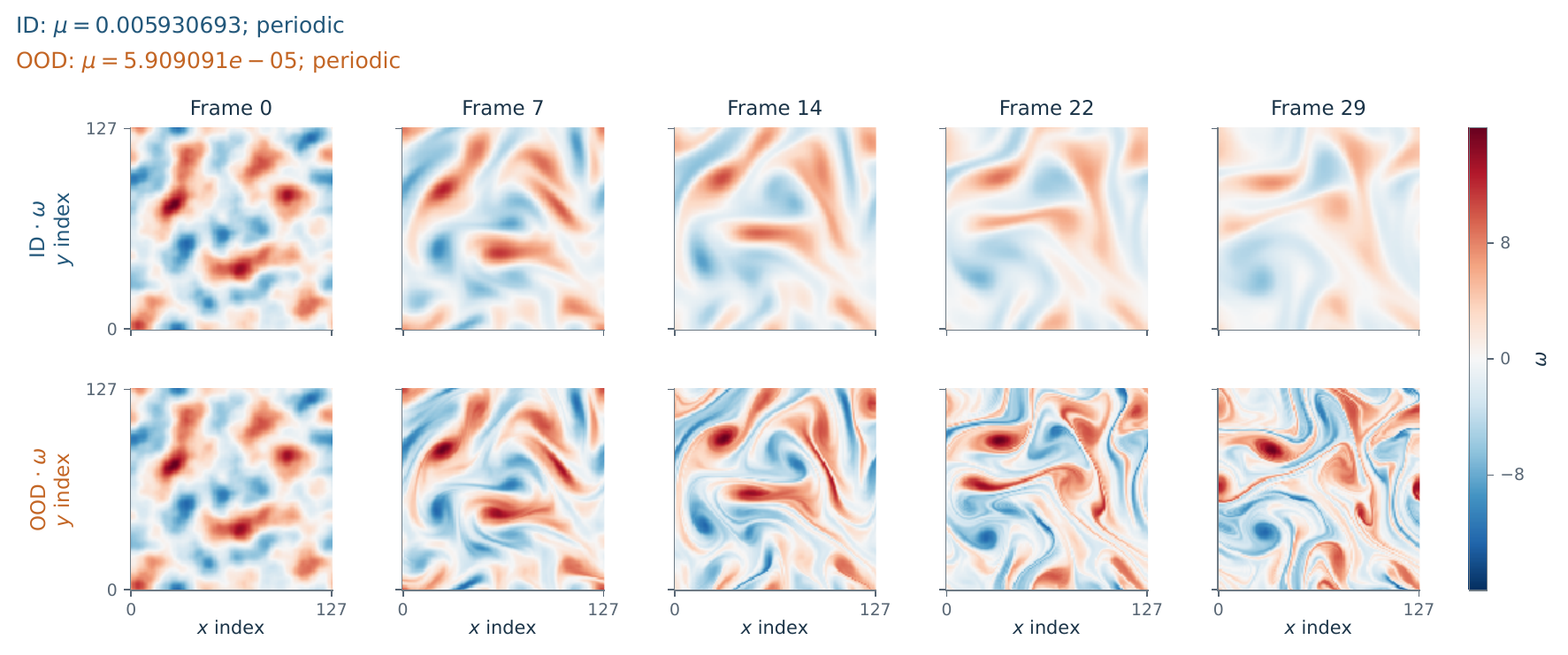}
    \caption{\textbf{Vorticity.}
ID: $\nu=0.00593069$; OOD: $\nu=5.90909\times10^{-5}$,
below the training support $[10^{-3},10^{-2}]$.
ID and OOD both use initial-condition, and their first saved vorticity fields are exactly equal.
Rows compare the same five saved-frame positions, with periodic boundaries
and a shared symmetric color scale.}
    \label{fig:vorticity}
\end{figure}

\lead{Numerical generation.}
We follow the numerical solver used in Zebra~\citep{serrano2024zebra}, combining a five-point
finite-difference Laplacian, the Arakawa Jacobian~\citep{arakawa1997computational}, an FFT-based Poisson
solver~\citep{cooley1965algorithm}, and fourth-order Runge--Kutta integration~\citep{butcher2016numerical}. The simulation is
performed on a $512\times512$ grid over $t\in[0,2]$, and the resulting
trajectories are spatially and temporally subsampled to $128\times128$
resolution with 30 frames. The visualization is shown in Figure~\ref{fig:vorticity}.

\subsection{HeterNS}
\label{app:heterns}

\begin{figure}[H]
    \centering
    \includegraphics[width=\linewidth]{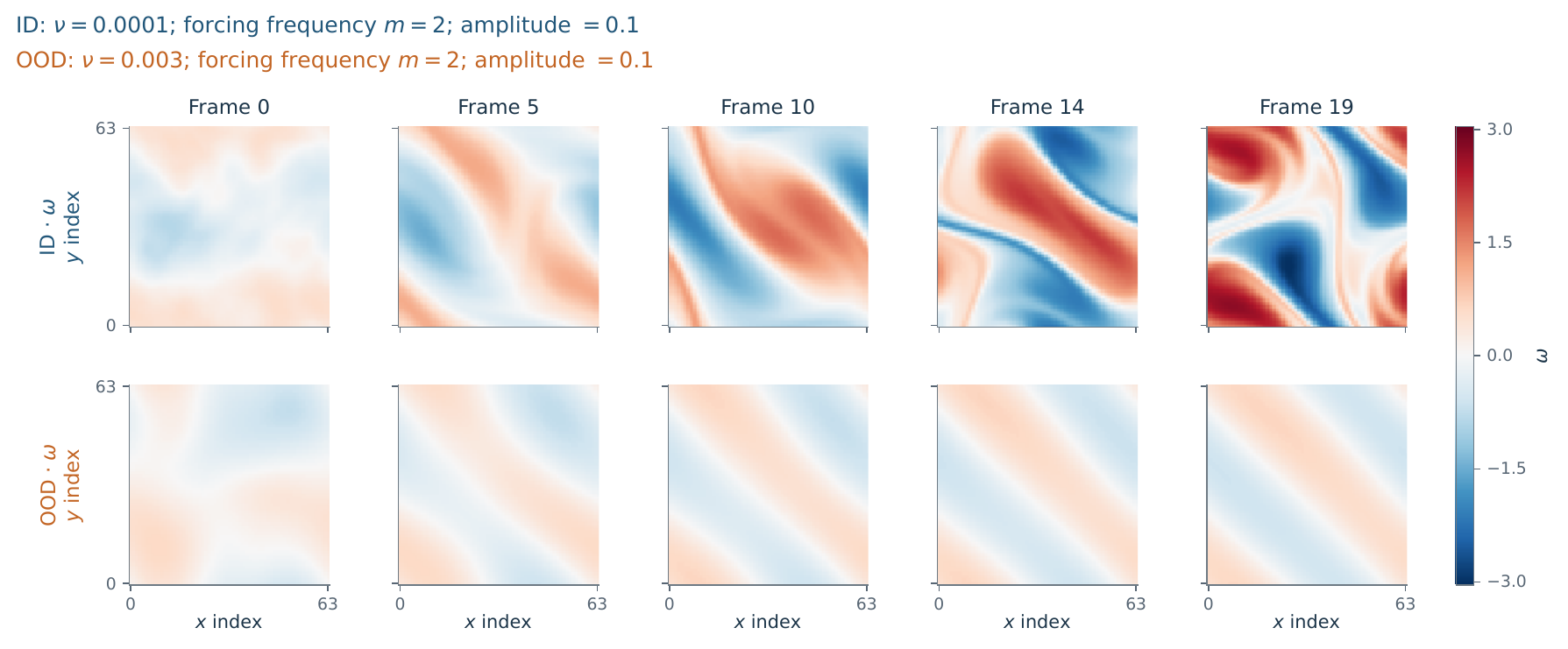}
    \caption{\textbf{HeterNS.}
ID: $\nu\simeq10^{-4}$; OOD: $\nu\simeq3\times10^{-3}$, above the
training range $[10^{-5},10^{-3}]$.
Both use $F(x,y)=0.1[\sin(2\pi(x+y))+\cos(2\pi(x+y))]$.}
    \label{fig:heterns}
\end{figure}

\lead{Equation and environments.}
HeterNS considers the two-dimensional incompressible Navier--Stokes equation
in vorticity form on the unit torus. All settings are identical to those used in UniSolver~\citep{zhou2024unisolver}; we include this description only for completeness.
\begin{equation}
\begin{aligned}
\partial_t \omega + \mathbf{u}\cdot\nabla\omega
&= \nu\Delta\omega + f(\mathbf{x}), \\
\nabla\cdot\mathbf{u} &= 0,
\end{aligned}
\end{equation}
where $\nu$ is the viscosity coefficient and the forcing is
\begin{equation}
f(\mathbf{x})
=
0.1\left[
\sin\left(\omega_f\pi(x_1+x_2)\right)
+
\cos\left(\omega_f\pi(x_1+x_2)\right)
\right].
\end{equation}
The PDE environment is therefore determined by the viscosity $\nu$ and
forcing frequency $\omega_f$. The training environments use $\nu\in
\left\{
10^{-5},\,5\times10^{-5},\,10^{-4},\,5\times10^{-4},\,10^{-3}
\right\},
\omega_f\in\{1,2,3\}$,
giving 15 distinct PDE configurations. Each configuration contains 1,000
trajectories, yielding 15,000 training trajectories in total. The ID test
set keeps the same PDE configurations while using unseen initial conditions. The visualization is shown in Figure~\ref{fig:heterns}.

\lead{OOD data.}
We consider three OOD settings. For viscosity interpolation, we use
$\nu\in
\left(\{2,3,4,6,7,8,9\}\times10^{-5}\right)
\cup
\left(\{2,3,4,6,7,8,9\}\times10^{-4}\right)$,
with $m\in\{1,2,3\}$. For viscosity extrapolation, we use
$\nu\in\{2,3,4,5,6,7,8,9\}\times10^{-3}$,
again with $m\in\{1,2,3\}$. For forcing OOD, we fix $\nu=10^{-5}$ and use
unseen forcing frequencies $m\in\{0.5,1.5,2.5,3.5\}$. We sample 1,500 OOD
trajectories in total, consisting of 900 viscosity-interpolation trajectories,
510 viscosity-extrapolation trajectories, and 90 forcing-OOD trajectories.

\subsection{Wave-2D: damped two-dimensional waves}
\label{app:wave2d}
\lead{Equation and environments.}
The released data represent
\begin{equation}
 \partial_{tt}u=c^2\Delta u-k\partial_tu,
 \label{eq:wave2d}
\end{equation}
with two stored channels, $(u,\partial_tu)$. Direct inspection establishes
$c \in [100,500]$ and $k \in [0,50]$ for the ID data. Training uses 1,200 distinct
parameter pairs selected from a $100\times100$ Cartesian parameter grid;
validation and test each contain 120 pairs. Each pair has 10 initial
conditions. 

\lead{Initial conditions.}
The initial condition is constructed as a sum of five Gaussian functions,
\begin{equation}
\omega_0(x,y)
=
\sum_{i=1}^{5}
\exp\left(
-\frac{(x-x_i)^2+(y-y_i)^2}{2\sigma_i^2}
\right),
\end{equation}
where $x_i,y_i\sim\mathcal{U}([0,1])$ and
$\sigma_i\sim\mathcal{U}([0.025,0.1])$. Each Gaussian has unit amplitude.

\begin{figure}[H]
    \centering
    \includegraphics[width=\linewidth]{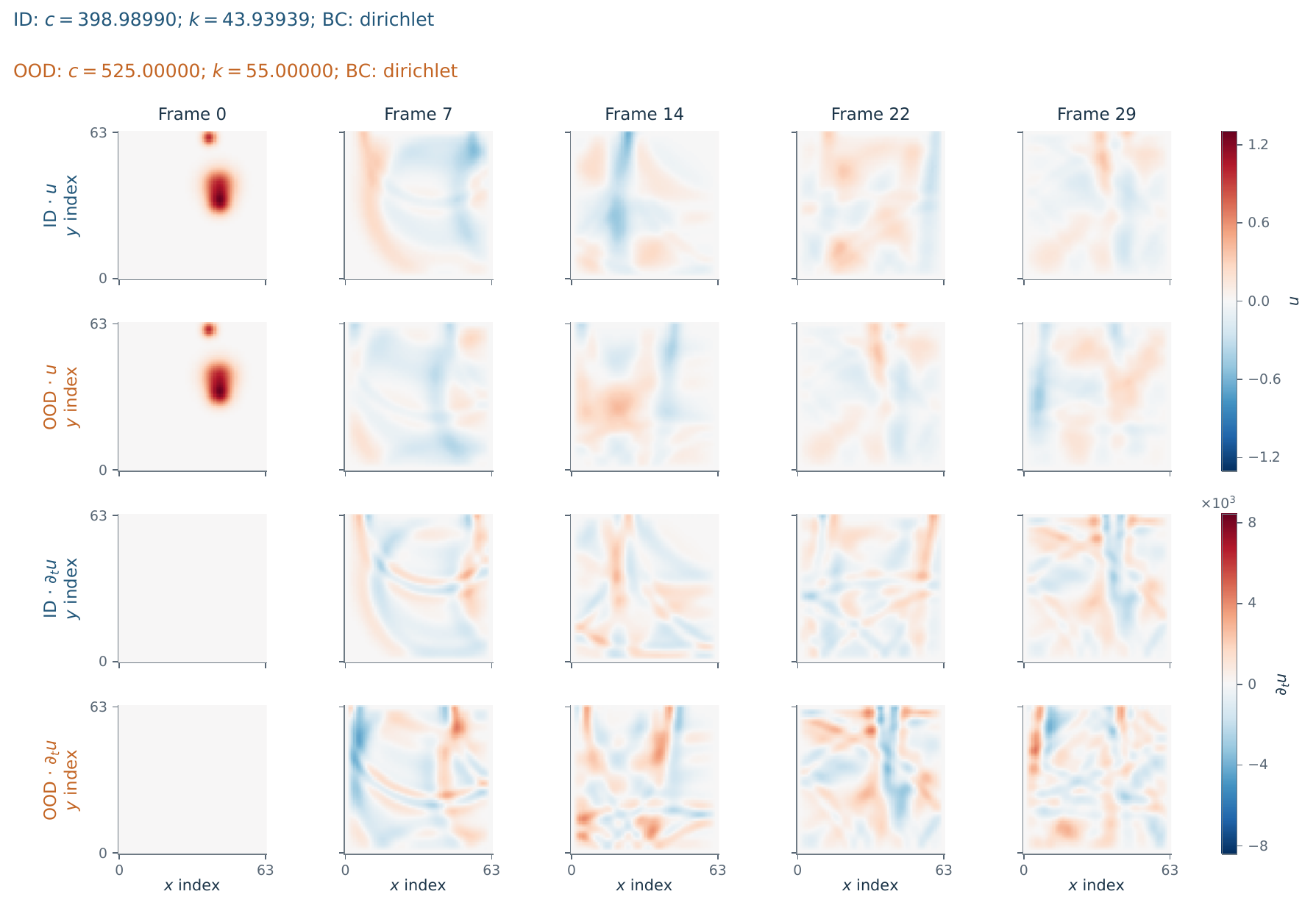}
    \caption{\textbf{Wave-2D:.}
ID: $(c,k)=(398.98990,43.93939)$; OOD: $(c,k)=(525,55)$,
outside the training ranges $c\in[100,500]$ and $k\in[0,50]$.}
    \label{fig:wave2d}
\end{figure}

\lead{Numerical generation.}
Following Zebra~\citep{serrano2024zebra}, the spatial domain is discretized
on a $64\times64$ grid using a $5\times5$ discrete Laplacian operator with
dirichlet boundary conditions. The dynamics are integrated using
a fourth-order Runge--Kutta scheme with time step
$\Delta t=6.25\times10^{-6}$ over $t\in[0,5\times10^{-3}]$. The visualization is shown in Figure~\ref{fig:wave2d}.

\lead{OOD data.}
The designated OOD release has 12 parameter pairs from the $5\times5$ grid
$c\in[500,550]$, $k\in[50,60]$, with ten trajectories per pair. Eleven pairs
are outside the ID support, while $(c,k)=(500,50)$ lies on its boundary.
Thus this file contains 110 strict parameter-extrapolation trajectories
and ten boundary-support trajectories.

\subsection{Gray--Scott Equation}
\label{app:grayscott}
\begin{figure}[H]
    \centering
    \includegraphics[width=\linewidth]{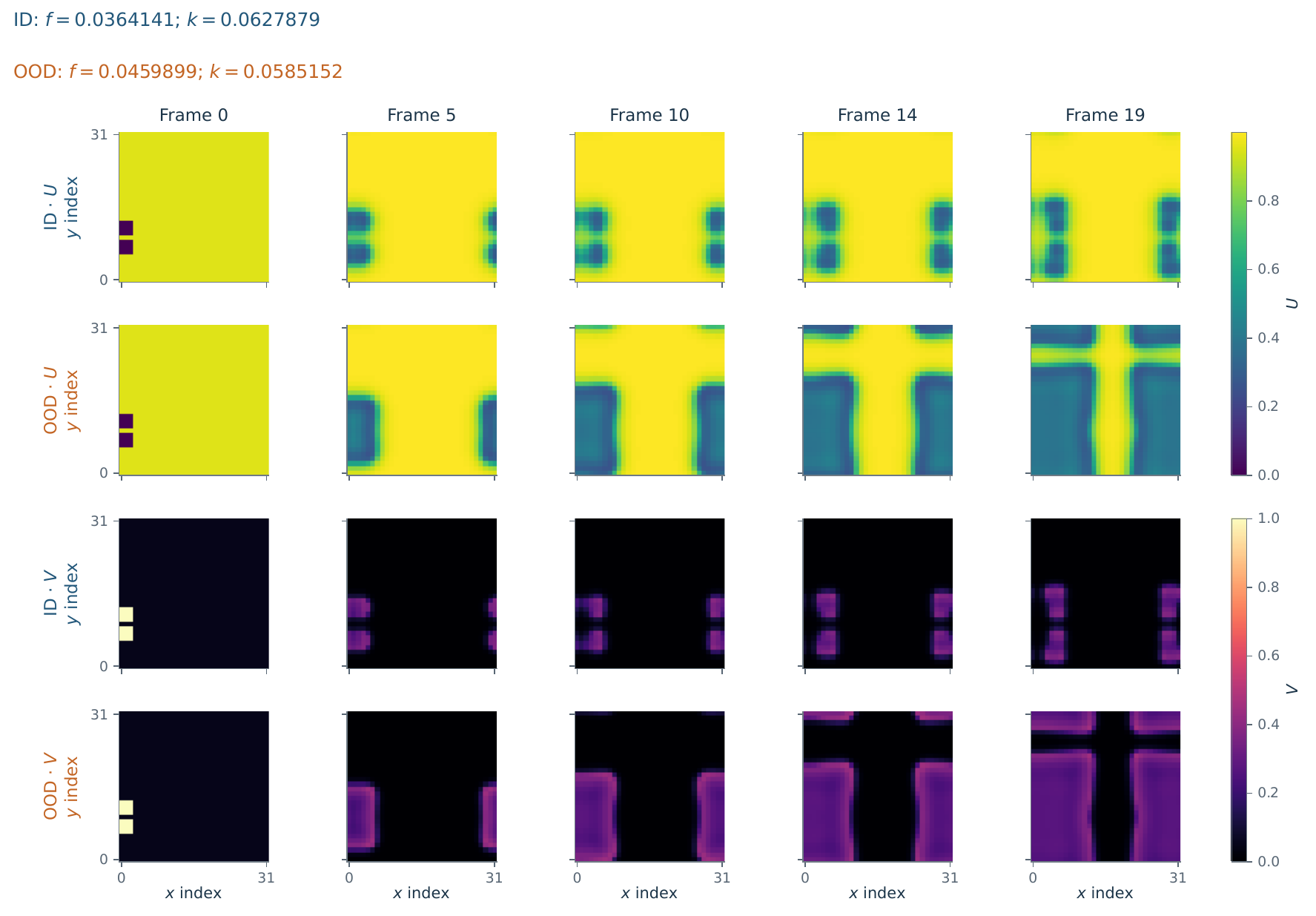}
    \caption{\textbf{Gray--Scott.}
ID: $(f,k)=(0.0364141,0.0627879)$;
OOD: $(f,k)=(0.0459899,0.0585152)$, outside the training ranges
$f\in[0.023,0.045]$ and $k\in[0.059,0.064]$.}
    \label{fig:gray}
\end{figure}

\lead{Equation and environments.}
The Gray--Scott dataset describes a two-dimensional reaction--diffusion
system governed by
\begin{equation}
\begin{aligned}
\frac{\partial u}{\partial t}
&= D_u \Delta u - uv^2 + F(1-u), \\
\frac{\partial v}{\partial t}
&= D_v \Delta v - uv^2 - (F+k)v ,
\end{aligned}
\end{equation}
where periodic boundary conditions are imposed and the diffusion
coefficients are fixed to $D_u=0.102$ and $D_v=0.204$.
For ID trajectories, the reaction parameters are sampled as
$F\sim\mathcal{U}([0.023,0.045])$ and
$k\sim\mathcal{U}([0.0590,0.0640])$.
For OOD evaluation, following ENMA~\citep{kassai2026enma}, we use
$F\sim\mathcal{U}([0.045,0.0467])$ and
$k\sim\mathcal{U}([0.0570,0.0590])$.
The spatial domain is discretized on a $32\times32$ grid with
spatial resolution $\Delta s=2$. The visualization is shown in Figure~\ref{fig:gray}.

\section{Architecture Details}

Our framework is trained in four stages: predictive representation pretraining, geometry alignment, latent predictor learning, and decoder training. We describe the architecture and tensor flow of each stage below. Let an input trajectory be

$$
\mathbf{X}\in\mathbb{R}^{B\times T\times C\times H\times W},
$$

where \(B\), \(T\), \(C\), \(H\), and \(W\) denote the batch size, temporal length, number of physical channels, and spatial resolution, respectively.

\subsection{Pretraining Stage}

\paragraph{Encoder.}
The pretraining stage learns physical representations using a joint-embedding predictive objective (JEPA). Given a trajectory

\[
\mathbf{X}
=
[\mathbf{x}_1,\ldots,\mathbf{x}_T],
\qquad
\mathbf{x}_t\in\mathbb{R}^{C\times H\times W},
\]

each frame is partitioned into non-overlapping spatial patches and projected into \(D\)-dimensional tokens. With a temporal tubelet size of one, this patch embedding does not mix information across adjacent frames, yielding \(N\) spatial tokens per time step.

The resulting spatiotemporal tokens are then processed jointly by a Vision Transformer. Specifically, the \(T\times N\) tokens are arranged as a single sequence and augmented with spatial and temporal positional information before being passed through a stack of multi-head self-attention and feed-forward blocks. The encoder therefore produces

\[
\mathbf{Z}
=
f_{\theta}(\mathbf{X})
\in
\mathbb{R}^{B\times T\times N\times D},
\]

where each temporal slice

\[
\mathbf{z}_t
=
[\mathbf{Z}]_t
\in
\mathbb{R}^{N\times D}
\]

is obtained in the context of the full trajectory rather than by independently encoding \(\mathbf{x}_t\).

\paragraph{Masked latent prediction.}
During pretraining, a subset of latent tokens is masked from the context encoder. A predictor receives the visible context representations together with embeddings corresponding to the masked target positions and predicts their latent representations. The prediction target is produced by a momentum-updated target encoder.

Let

$$
\mathbf{Z}^{c}\in\mathbb{R}^{B\times N_c\times D}
$$

denote the visible context tokens and

$$
\mathbf{Z}^{t}\in\mathbb{R}^{B\times N_t\times D}
$$

the latent representations of the target tokens, where \(N_c\) and \(N_t\) denote the number of context and target tokens, respectively. The latent predictor maps the context representation to

$$
\widehat{\mathbf{Z}}^{t}
=
p_{\psi}(\mathbf{Z}^{c})
\in
\mathbb{R}^{B\times N_t\times D}.
$$

The pretraining objective aligns the predicted target representations with those generated by the target encoder,

$$
\mathcal{L}_{\mathrm{pre}}
=
\mathcal{D}
\left(
\widehat{\mathbf{Z}}^{t},
\mathbf{Z}^{t}
\right),
$$

where \(\mathcal{D}\) denotes the latent-space prediction loss. After pretraining, the encoder \(f_{\theta}\) is retained as the representation model for the following stages.


\subsection{Geometry Alignment Stage}

The pretrained representation preserves predictive information about the physical trajectory, but its latent coordinates are not explicitly constrained to reflect the evolution geometry of the underlying physical states. We therefore introduce a lightweight geometry projector to adapt the latent state space before learning the dynamics model.

\paragraph{Geometry projector.}
The pretrained encoder is frozen during this stage. For each latent token

$$
\mathbf{z}_{t,n}\in\mathbb{R}^{D},
$$

the geometry projector applies a token-wise residual transformation,

$$
\mathbf{q}_{t,n}
=
\mathbf{z}_{t,n}
+
g_{\phi}(\mathbf{z}_{t,n}),
$$

where

$$
g_{\phi}:
\mathbb{R}^{D}
\rightarrow
\mathbb{R}^{D}
$$

is a multilayer perceptron consisting of normalization, channel expansion to an intermediate dimension \(D_g\), nonlinear activation, and projection back to \(D\).

The projector therefore preserves the complete latent tensor shape,

$$
\mathbf{Q}
=
\mathbf{Z}+g_{\phi}(\mathbf{Z})
\in
\mathbb{R}^{B\times T\times N\times D}.
$$

The transformation is independently applied to every token and time step. In particular, the projector does not perform temporal or spatial-token mixing, and only adjusts the latent channel coordinates.

\paragraph{Temporal geometry alignment.}
To characterize the temporal geometry of the physical trajectory, each physical state is flattened as

$$
\widetilde{\mathbf{x}}_t
\in
\mathbb{R}^{CHW},
$$

while each projected latent state is flattened across its token and channel dimensions,

$$
\widetilde{\mathbf{q}}_t
\in
\mathbb{R}^{ND}.
$$

Temporal displacement vectors are then computed independently in the two state spaces,

$$
\Delta\mathbf{x}_t
=
\widetilde{\mathbf{x}}_{t+1}
-
\widetilde{\mathbf{x}}_t,
$$

and

$$
\Delta\mathbf{q}_t
=
\widetilde{\mathbf{q}}_{t+1}
-
\widetilde{\mathbf{q}}_t.
$$

Although the physical and latent states have different feature dimensions, i.e.,

$$
CHW \neq ND,
$$

their trajectory geometry can be compared through dimension-independent directional statistics. For a temporal offset \(\ell\), we compute

$$
s^{x}_{t,\ell}
=
\frac{
\Delta\mathbf{x}_{t}^{\top}
\Delta\mathbf{x}_{t+\ell}
}{
\|\Delta\mathbf{x}_{t}\|_2
\|\Delta\mathbf{x}_{t+\ell}\|_2
},
$$

and

$$
s^{q}_{t,\ell}
=
\frac{
\Delta\mathbf{q}_{t}^{\top}
\Delta\mathbf{q}_{t+\ell}
}{
\|\Delta\mathbf{q}_{t}\|_2
\|\Delta\mathbf{q}_{t+\ell}\|_2
}.
$$

Both quantities are scalar directional similarities, allowing the trajectory geometry of the two spaces to be directly aligned. The geometry loss aggregates the discrepancy across a set of temporal offsets \(\mathcal{S}\),

$$
\mathcal{L}_{\mathrm{geo}}
=
\sum_{\ell\in\mathcal{S}}
w_{\ell}
\mathcal{D}_{\mathrm{geo}}
\left(
s^{q}_{t,\ell},
s^{x}_{t,\ell}
\right),
$$

where \(w_{\ell}\) controls the contribution of each temporal scale.

To prevent the projector from unnecessarily altering the predictive representation, we additionally constrain the projected states to remain close to the pretrained latent states,

$$
\mathcal{L}_{\mathrm{anchor}}
=
\mathcal{D}_{\mathrm{anchor}}
(\mathbf{Q},\mathbf{Z}).
$$

The geometry-alignment stage optimizes only the projector parameters while keeping the pretrained encoder fixed.

\paragraph{Auxiliary conditional causal predictor.}
In addition to geometric alignment, we introduce an auxiliary causal predictor to encourage the projected space to remain compatible with parameter-dependent dynamics. Given the projected latent history and the physical parameter \(\boldsymbol{\xi}\), the predictor estimates the next latent displacement,

$$
\widehat{\Delta\mathbf{q}}_t
=
P_{\psi}
\left(
\mathbf{q},
\boldsymbol{\xi}
\right),
$$

where \(\boldsymbol{\xi}\) is embedded as a conditioning token and supplied to the causal predictor. Importantly, the geometry projector itself is parameter-independent and takes only the pretrained latent representation as input; parameter information enters exclusively through the auxiliary dynamics predictor.

The corresponding prediction objective is

$$
\mathcal{L}_{dyn}
=
\mathcal{D}_{\mathrm{pred}}
\left(
\widehat{\Delta\mathbf{q}}_t,
\Delta\mathbf{q}_t
\right),
$$

where

$$
\Delta\mathbf{q}_t
=
\mathbf{q}_{t+1}-\mathbf{q}_t.
$$

Together with the geometry and anchor objectives, the alignment-stage loss is

$$
\mathcal{L}_{\mathrm{align}}
=
\mathcal{L}_{dyn}
+
\lambda_{\mathrm{geo}}\mathcal{L}_{\mathrm{geo}}
+
\lambda_{\mathrm{anchor}}\mathcal{L}_{\mathrm{anchor}}.
$$

The pretrained encoder remains frozen, while the geometry projector and auxiliary predictor are jointly optimized during this stage.

\subsection{Predictor Learning Stage}

After geometry alignment, physical evolution is modeled directly in the aligned latent state space. Given a projected latent state

$$
\mathbf{q}
\in
\mathbb{R}^{N\times D}
$$

\paragraph{Physics-structured latent predictor.}
The latent transition is decomposed into a parameter-independent evolution component and a parameter-dependent response component. we model its evolution through a parameter-structured latent vector field,
\[
\dot q_t
=
\mathcal{F}_\theta(q_t,\xi)
=
\mathcal{F}_{\mathrm{evo}}(q_t)
+
\mathcal{F}_{\mathrm{par}}(q_t,\xi),
\]

The evolution branch

$$
\mathcal{F}_{\mathrm{evo}}:
\mathbb{R}^{N\times D}
\rightarrow
\mathbb{R}^{N\times D}
$$

models dynamics shared across different governing conditions. The parameter-dependent branch

$$
\mathcal{F}_{\mathrm{par}}:
\mathbb{R}^{N\times D}
\times
\mathbb{R}^{D_{\xi}}
\rightarrow
\mathbb{R}^{N\times D}
$$

captures the change in latent evolution induced by the governing parameters.

Both branches preserve the spatial-token structure of the representation, so that their outputs can be directly combined with the current latent state. For continuous-time predictors, the next latent state is obtained by
numerically integrating the vector field over one observation interval,
\[
\hat q_{t+1}
=
\Phi_{\rm ODE}(q_t,\mathcal{F}_\theta,\xi,\Delta t).
\]

\paragraph{Autoregressive rollout.}
For long-horizon prediction, the predicted latent state is recursively used as the input to the next transition,

$$
\widehat{\mathbf{q}}_{t+k}
=
\mathcal{P}
\left(
\widehat{\mathbf{q}}_{t+k-1},
\boldsymbol{\xi}
\right),
$$

where \(\mathcal{P}\) denotes the complete latent transition operator.

The dynamics are therefore evolved entirely in the learned latent space. Physical predictions are recovered only when needed through a reconstruction head

$$
\mathcal{D}_{\omega}:
\mathbb{R}^{N\times D}
\rightarrow
\mathbb{R}^{C\times H\times W}.
$$

This avoids repeatedly mapping autoregressive predictions between the physical and latent domains during rollout.

\subsection{Decoder Learning Stage}
\label{app:decoder}

The decoder is trained after the geometry projector and latent predictor have been learned. During this stage, the encoder, geometry projector, and latent predictor are all frozen, and only the decoder parameters are optimized.

\paragraph{Rollout-based latent generation.}
Given a physical trajectory

$$
\mathbf{X}
=
[\mathbf{x}_0,\mathbf{x}_1,\ldots,\mathbf{x}_{T-1}]
\in
\mathbb{R}^{B\times T\times C\times H\times W},
$$

only the initial physical state \(\mathbf{x}_0\) is used to construct the latent input. It is first mapped by the frozen encoder and geometry projector as

$$
\mathbf{z}_0
=
f_{\theta}(\mathbf{x}_0)
\in
\mathbb{R}^{B\times N\times D},
$$

$$
\mathbf{q}_0
=
\mathbf{z}_0 +
g_{\phi}(\mathbf{z}_0)
\in
\mathbb{R}^{B\times N\times D}.
$$

Starting from $q_0$, the frozen physics-structured latent predictor
recursively generates the remaining latent states by integrating the
learned latent vector field:
\[
\hat q_{t+1}
=
\Phi_{\mathrm{ODE}}
(\hat q_t,F_\theta,\xi,\Delta t),
\qquad
\hat q_0=q_0.
\]
In our implementation, $\Phi_{\mathrm{ODE}}$ is instantiated using
fourth-order Runge--Kutta integration with four substeps.

This produces the complete latent rollout

$$
\widehat{\mathbf{Q}}
=
[
\mathbf{q}_0,
\widehat{\mathbf{q}}_1,
\ldots,
\widehat{\mathbf{q}}_{T-1}
]
\in
\mathbb{R}^{B\times T\times N\times D}.
$$

Thus, future ground-truth states
\(\mathbf{x}_1,\ldots,\mathbf{x}_{T-1}\)
are never encoded to construct the decoder input. They are used only as reconstruction targets. This exposes the decoder during training to the same autoregressive latent distribution encountered at inference time, including deviations accumulated during latent rollout.

\paragraph{Decoder.}
Before decoding, the temporal and spatial-token dimensions of the rollout are combined,

$$
\widehat{\mathbf{Q}}
\in
\mathbb{R}^{B\times T\times N\times D}
\rightarrow
\mathbb{R}^{B\times TN\times D}.
$$

The latent tokens are first normalized and projected from the latent dimension \(D\) to an initial convolutional feature dimension \(D_0\). Since each latent state contains a two-dimensional token grid with

$$
N=N_hN_w,
$$

the token sequence is rearranged into frame-wise spatial feature maps,

$$
\mathbb{R}^{B\times TN\times D_0}
\rightarrow
\mathbb{R}^{(BT)\times D_0\times N_h\times N_w}.
$$

The decoder then progressively reconstructs the physical spatial resolution. Starting from the coarse latent token grid, residual convolutional blocks refine the features, followed by a sequence of spatial upsampling stages,

$$
\widehat{\mathbf{Q}}^{(s+1)}
=
\mathcal{R}^{(s)}
\left(
\operatorname{Up}_2
\left(
\widehat{\mathbf{Q}}^{(s)}
\right)
\right),
$$

where \(\operatorname{Up}_2(\cdot)\) denotes a factor-of-two spatial interpolation and
\(\mathcal{R}^{(s)}\) denotes the residual convolutional refinement at stage \(s\). After \(S\) stages, the feature resolution is restored from

$$
N_h\times N_w
$$

to

$$
H\times W.
$$

A final convolutional head maps the reconstructed features to the \(C\) physical channels. The frame dimension is then restored, yielding

$$
\widehat{\mathbf{X}}
=
\mathcal{D}_{\omega}
\left(
\widehat{\mathbf{Q}}
\right)
\in
\mathbb{R}^{B\times T\times C\times H\times W}.
$$

Although the full latent rollout is provided jointly as the decoder input tensor, the spatial reconstruction is performed frame-wise: the temporal dimension is folded into the batch dimension before the convolutional decoding blocks. Consequently, the decoder itself does not model temporal evolution.

\paragraph{Training objective.}
The complete ground-truth physical trajectory serves as the reconstruction target,

$$
\mathcal{L}_{\mathrm{dec}}
=
\mathcal{D}_{\mathrm{phy}}
\left(
\widehat{\mathbf{X}},
\mathbf{X}
\right).
$$

During this stage  $
f_{\theta},
g_{\phi},
\mathcal{P}_{\psi}
$remain frozen, and gradients are propagated only through
\(\mathcal{D}_{\omega}\).

\section{Implementation Details}
\label{App:Imp}

All codes are written in Pytorch~\citep{paszke2019pytorch}. All experiments are conducted on 4 RTX PRO 6000 Blackwell 96G with approximately total of 9000 GPU hours.

\subsection{PDE-JEPA Implementation}

The hyperparameter settings for PDE-JEPA's architecture across
all datasets are summarized in Table~\ref{tab:method-four-stage-configuration}. All stages use AdamW with $(\beta_1,\beta_2)=(0.9,0.999)$, except structured predictors, which use $(0.9,0.95)$.

\subsubsection{Physics-Structured Latent Predictor Implementations}

Following the general formulation in Section~\ref{sec:physics_predictor},
the predictor first maps the current latent state through a shared backbone,
\[
\mathbf H_\theta(\mathbf q)
=
\mathcal H_\theta(\mathbf q),
\]
and applies independent component heads to the shared features,
\[
C_\theta(\mathbf q)
=
h_{C,\theta}\!\left(\mathbf H_\theta(\mathbf q)\right),
\qquad
D_\theta^{(j)}(\mathbf q)
=
h_{j,\theta}\!\left(\mathbf H_\theta(\mathbf q)\right),
\quad j=1,\ldots,M.
\]
Each component has the same spatial and channel dimensions as the latent
state, i.e.,
\[
C_\theta(\mathbf q),\,
D_\theta^{(j)}(\mathbf q)
\in\mathbb{R}^{N\times D}.
\]

The latent vector field is then parameterized as
\[
\dot{\mathbf q}
=
C_{\theta}(\mathbf q)
+
\sum_{j=1}^{M}
r_j(\boldsymbol{\xi})\,D_{\theta}^{(j)}(\mathbf q)
,
\]
where $r_j(\boldsymbol{\xi})$ denotes the normalized coefficient associated
with the $j$-th parameter-dependent component.
We instantiate this general form according to the governing structure of
each PDE.

For a training set of $N_{\rm tr}$ trajectories, we denote
\[
\left\langle g(a)\right\rangle_{\rm tr}
=
\frac{1}{N_{\rm tr}}
\sum_{i=1}^{N_{\rm tr}} g(a_i).
\]

\paragraph{Vorticity.}
For Vorticity, the latent vector field is
\[
\dot{\mathbf q}
=
\mathbf C_{\theta}(\mathbf q)
+
r_{\nu}\mathbf D^\nu_{\theta}(\mathbf q),
\qquad
r_{\nu}
=
\frac{\nu-\nu_0}{s_{\nu}},
\]
where
\[
\nu_0
=
\frac{\nu_{\min}^{\rm tr}+\nu_{\max}^{\rm tr}}{2},
\qquad
s_{\nu}
=
\frac{\nu_{\max}^{\rm tr}-\nu_{\min}^{\rm tr}}{2}.
\]
Here $\nu_{\min}^{\rm tr}$ and $\nu_{\max}^{\rm tr}$ denote the minimum
and maximum viscosities in the ID training split.

\paragraph{Wave-2D.}
For Wave-2D, we use
\[
\dot{\mathbf q}
=
\mathbf C_{\theta}(\mathbf q)
+
r_{c^2}\mathbf D^c_{\theta}(\mathbf q)
-
r_k\mathbf D^k_{\theta}(\mathbf q),
\]
where
\[
r_{c^2}
=
\frac{c^2}{s_{c^2}},
\qquad
r_k
=
\frac{k}{s_k^{\rm Wave}},
\]
with
\[
s_{c^2}
=
\sqrt{
\left\langle c^4\right\rangle_{\rm tr}
},
\qquad
s_k^{\rm Wave}
=
\sqrt{
\left\langle k^2\right\rangle_{\rm tr}
}.
\]

\paragraph{Gray--Scott.}
For Gray--Scott, the predictor is
\[
\dot{\mathbf q}
=
\mathbf C_{\theta}(\mathbf q)
+
\frac{F}{s_F}\mathbf D^F_{\theta}(\mathbf q)
+
\frac{k}{s_k^{\rm GS}}\mathbf D^k_{\theta}(\mathbf q),
\]
where
\[
s_F
=
\left\langle |F| \right\rangle_{\rm tr},
\qquad
s_k^{\rm GS}
=
\left\langle |k| \right\rangle_{\rm tr}.
\]

\paragraph{Combined Equation.}
For the combined equation, we associate separate latent vector fields
with the transport, diffusion, and dispersion coefficients:
\[
\dot{\mathbf q}
=
-
\frac{\alpha}{s_{\alpha}}
\mathbf D^\alpha_{\theta}(\mathbf q)
+
\frac{\beta}{s_{\beta}}
\mathbf D^\beta_{\theta}(\mathbf q)
-
\frac{\gamma}{s_{\gamma}}
\mathbf D^\gamma_{\theta}(\mathbf q),
\]
where the coefficient scales are
\[
s_p
=
\sqrt{
\left\langle p^2\right\rangle_{\rm tr}
},
\qquad
p\in\{\alpha,\beta,\gamma\}.
\]

\paragraph{HeterNS.}
For HeterNS, viscosity and the spatial forcing field are modeled
separately:
\[
\dot{\mathbf q}
=
\mathbf C_{\theta}(\mathbf q)
+
r_{\nu}\mathbf D^\nu_{\theta}(\mathbf q)
+
m \mathbf D^{m}_{\theta}(\mathbf q),
\]
where
\[
r_{\nu}
=
\frac{\nu-\nu_0}{s_{\nu}},
\]
and
\[
\nu_0
=
\frac{\nu_{\min}^{\rm tr}+\nu_{\max}^{\rm tr}}{2},
\qquad
s_{\nu}
=
\frac{\nu_{\max}^{\rm tr}-\nu_{\min}^{\rm tr}}{2}.
\]

\subsubsection{Latent time integration.}
For predictors formulated as continuous latent vector fields, we obtain the
next latent state by numerically integrating
\[
\dot{\mathbf q}
=
\mathcal{F}_{\theta}(\mathbf q,\boldsymbol{\xi}),
\]
where $\mathcal{F}_{\theta}$ denotes the corresponding
physics-structured vector field defined above. Given the current latent state
$\mathbf q_t$ and the physical interval $\Delta t$ between two consecutive
frames, we use the classical fourth-order Runge--Kutta (RK4) scheme with
$M$ substeps. Let $h=\Delta t/M$ and
$\mathbf q_t^{(0)}=\mathbf q_t$. For each substep $m=0,\ldots,M-1$,
\[
\begin{aligned}
\mathbf k_1 &=
\mathcal{F}_{\theta}
\left(\mathbf q_t^{(m)},\boldsymbol{\xi}\right),\\
\mathbf k_2 &=
\mathcal{F}_{\theta}
\left(\mathbf q_t^{(m)}+\frac{h}{2}\mathbf k_1,
      \boldsymbol{\xi}\right),\\
\mathbf k_3 &=
\mathcal{F}_{\theta}
\left(\mathbf q_t^{(m)}+\frac{h}{2}\mathbf k_2,
      \boldsymbol{\xi}\right),\\
\mathbf k_4 &=
\mathcal{F}_{\theta}
\left(\mathbf q_t^{(m)}+h\mathbf k_3,
      \boldsymbol{\xi}\right),
\end{aligned}
\]
and the latent state is updated as
\[
\mathbf q_t^{(m+1)}
=
\mathbf q_t^{(m)}
+
\frac{h}{6}
\left(
\mathbf k_1+2\mathbf k_2+2\mathbf k_3+\mathbf k_4
\right).
\]
After $M$ substeps, the prediction for the next frame is
\[
\widehat{\mathbf q}_{t+1}
=
\mathbf q_t^{(M)}.
\]
We use $M=4$ RK4 substeps for each latent transition.

\begingroup

\fontsize{7.2}{7.8}\selectfont
\setlength{\tabcolsep}{0.8pt}
\renewcommand{\arraystretch}{1.00}

\setlength{\LTleft}{0pt}
\setlength{\LTright}{0pt}
\setlength{\LTpost}{0pt}

\begin{xltabular}{\textwidth}{
@{}
>{\raggedright\arraybackslash}p{0.19\textwidth}
*{9}{>{\centering\arraybackslash}X}
@{}
}

\caption{\textbf{Implementation and training hyperparameters of PDE-JEPA.}}
\label{tab:method-four-stage-configuration}\\

\toprule
Hyperparameter
& Combined
& Advection
& Burgers
& Heat
& Wave-B
& GS
& Wave2D
& Vorticity
& HeterNS \\
\midrule
\endfirsthead

\multicolumn{10}{c}{
\tablename\ \thetable{} -- continued from previous page
}\\
\toprule
Hyperparameter
& Combined
& Advection
& Burgers
& Heat
& Wave-B
& GS
& Wave2D
& Vorticity
& HeterNS \\
\midrule
\endhead

\midrule
\multicolumn{10}{r}{Continued on next page}\\
\endfoot

\bottomrule
\endlastfoot

Model input grid
& $256$
& $256$
& $256$
& $256$
& $256$
& $32{\times}32$
& $64{\times}64$
& $128{\times}128$
& $64{\times}64$ \\

Sampled frames / stride
& 14 / 10
& 14 / 5
& 25 / 10
& 25 / 10
& 25 / 10
& 20 / 1
& 30 / 1
& 30 / 1
& 20 / 1 \\

Channels / patch size
& $1/4$
& $1/4$
& $1/4$
& $1/4$
& $1/4$
& $2$/$4{\times}4$
& $2$/$4{\times}4$
& $1$/$8{\times}8$
& $1$/$4{\times}4$ \\

Latent dimension
& 192 & 192 & 192 & 192 & 192
& 192 & 192 & 192 & 192 \\

\midrule
\multicolumn{10}{l}{\textbf{Stage 1: Pretraining}} \\

Encoder depth
& 12 & 12 & 12 & 12 & 12
& 24 & 12 & 12 & 12 \\

Predictor depth
& 12 & 12 & 12 & 12 & 12
& 24 & 12 & 12 & 12 \\

Predictor hidden size
& 384 & 384 & 384 & 384 & 384
& 384 & 384 & 384 & 384 \\

Encoder attention heads
& 3 & 3 & 3 & 3 & 3
& 3 & 3 & 3 & 3 \\

Predictor attention heads
& 12 & 12 & 12 & 12 & 12
& 12 & 12 & 12 & 12 \\

MLP ratio
& 4 & 4 & 4 & 4 & 4
& 4 & 4 & 4 & 4 \\

Dropout
& 0 & 0 & 0 & 0 & 0
& 0 & 0 & 0 & 0 \\

Normalization type
& LayerNorm & LayerNorm & LayerNorm & LayerNorm & LayerNorm
& LayerNorm & LayerNorm & LayerNorm & LayerNorm \\

Activation
& GELU & GELU & GELU & GELU & GELU
& GELU & GELU & GELU & GELU \\

Positional encoding
& RoPE & RoPE & RoPE & RoPE & RoPE
& RoPE & RoPE & RoPE & RoPE \\

Epoch
& 15 + 20
& 15 + 30
& 20 + 50
& 20 + 50
& 20 + 50
& 20 + 50
& 20 + 50
& 20 + 50
& 20 + 50 \\

Global batch
& 32 & 32 & 32 & 32 & 32
& 32 & 32 & 32 & 32 \\

Peak LR
& $3.5{\times}10^{-4}$
& $3.5{\times}10^{-4}$
& $3.5{\times}10^{-4}$
& $3.5{\times}10^{-4}$
& $3.5{\times}10^{-4}$
& $3.5{\times}10^{-4}$
& $3.5{\times}10^{-4}$
& $3.5{\times}10^{-4}$
& $3.5{\times}10^{-4}$ \\

Final LR
& $10^{-6}$ & $10^{-6}$ & $10^{-6}$
& $10^{-6}$ & $10^{-6}$ & $10^{-6}$
& $10^{-6}$ & $10^{-6}$ & $10^{-6}$ \\

Weight decay
& 0.04 & 0.04 & 0.04 & 0.04 & 0.04
& 0.04 & 0.04 & 0.04 & 0.04 \\

Warmup epochs
& 2 / 0 & 2 / 0 & 2 / 0 & 2 / 0 & 2 / 0
& 2 / 0 & 2 / 0 & 2 / 0 & 2 / 0 \\

EMA momentum
& 0.99925 & 0.99925 & 0.99925
& 0.99925 & 0.99925 & 0.99925
& 0.99925 & 0.99925 & 0.99925 \\

Spatial mask scales
& .15 / .70
& .12 / .60
& .15 / .70
& .15 / .70
& .15 / .70
& .15 / .70
& .12 / .60
& .15 / .70
& .15 / .70 \\

\midrule
\multicolumn{10}{l}{\textbf{Stage 2: Geometry Alignment}} \\

Projector MLP layers
& 2 & 2 & 2 & 2 & 2
& 2 & 2 & 2 & 2 \\

Projector hidden width
& 384 & 384 & 384 & 384 & 384
& 384 & 384 & 384 & 384 \\

Projector residual
& True & True & True & True & True
& True & True & True & True \\

Zero output initialization
& True & True & True & True & True
& True & True & True & True \\

Causal predictor depth
& 24 & 24 & 24 & 24 & 24
& 24 & 24 & 24 & 24 \\

Causal predictor width
& 384 & 384 & 384 & 384 & 384
& 384 & 384 & 384 & 384 \\

Causal predictor head
& 12 & 12 & 12 & 12 & 12
& 12 & 12 & 12 & 12 \\

Normalization
& LN & LN & LN & LN & LN
& LN & LN & LN & LN \\

Activation
& GELU & GELU & GELU & GELU & GELU
& GELU & GELU & GELU & GELU \\

Epoch
& 50 & 50 & 50 & 50 & 50
& 50 & 50 & 50 & 50 \\

Global batch
& 16 & 16 & 16 & 16 & 16
& 16 & 16 & 16 & 16 \\

Peak LR
& $5{\times}10^{-5}$
& $5{\times}10^{-5}$
& $5{\times}10^{-5}$
& $5{\times}10^{-5}$
& $5{\times}10^{-5}$
& $5{\times}10^{-5}$
& $5{\times}10^{-5}$
& $5{\times}10^{-5}$
& $5{\times}10^{-5}$ \\

Final LR
& $10^{-6}$ & $10^{-6}$ & $10^{-6}$
& $10^{-6}$ & $10^{-6}$ & $10^{-6}$
& $10^{-6}$ & $10^{-6}$ & $10^{-6}$ \\

Weight decay
& 0.04 & 0.04 & 0.04 & 0.04 & 0.04
& 0.04 & 0.04 & 0.04 & 0.04 \\

Warmup epochs
& 5 & 5 & 5 & 5 & 5
& 5 & 5 & 5 & 5 \\

Loss weights
$\left(\lambda_{\rm geom},\lambda_{\rm anchor}\right)$
& .1 / .01
& .1 / .01
& .1 / .01
& .1 / .01
& .1 / .01
& .1 / .01
& .1 / .01
& .1 / .01
& .1 / .01 \\

\midrule
\multicolumn{10}{l}{\textbf{Stage 3: Predictor Learning}} \\

Dynamics / integrator
& RK4 $\times4$
&RK4 $\times4$
& RK4 $\times4$
& RK4 $\times4$
& RK4 $\times4$
& RK4 $\times4$
& RK4 $\times4$
& RK4 $\times4$
& RK4 $\times4$ \\

Predictor depth
& 24 & 24 & 24 & 24 & 24
& 8 & 24 & 24 & 24 \\
Hidden size
& 384
& 384
& 384
& 384
& 384
& 384
& 384
& 384
& 384 \\

Attention heads
& 12
& 12
& 12
& 12
& 12
& 12
& 12
& 12
& 12 \\

MLP ratio
& 4
& 4
& 4
& 4
& 3.90104
& 4
& 3.85938
& 4
& 4 \\

Feed-forward width
& 1536
& 1536
& 1536
& 1536
& 1498
& 1536
& 1482
& 1536
& 1536 \\

Normalization
& LN
& LN
& LN
& LN
& LN
& LN
& LN
& LN
& LN \\

Activation
& GELU
& GELU
& GELU
& GELU
& GELU
& GELU
& GELU
& GELU
&GELU \\

Dropout
& 0
& 0
& 0
& 0
& 0
& 0
& 0
& 0
&0  \\

Positional encoding
& RoPE
& RoPE
& RoPE
& RoPE
& RoPE
& sin/cos
& RoPE
& sin/cos
& RoPE\\

Dynamics state dimension
& 192
& 192
& 192
& 192
& 192
& 192
& 192
& 192
& 192 \\

Condition dimension
& 3
& 1
& 7
& 7
& 4
& 2
& 2
& 1
& 2\\

Epoch
& 50 & 50 & 50 & 50 & 50
& 50 & 50 & 50 & 100\\

Global batch
& 16 & 16 & 16 & 16
& 64 & 64 & 64 & 64 & 64 \\

Peak LR
& $5{\times}10^{-5}$
& $5{\times}10^{-5}$
& $5{\times}10^{-5}$
& $5{\times}10^{-5}$
& $10^{-4}$
& $10^{-4}$
& $10^{-4}$
& $10^{-4}$
& $10^{-4}$ \\

Final LR
& $10^{-6}$ & $10^{-6}$ & $10^{-6}$
& $10^{-6}$ & $10^{-6}$ & $10^{-6}$
& $10^{-6}$ & $10^{-6}$ & $10^{-6}$ \\

Weight decay
& 0.04 & 0.04 & 0.04 & 0.04
& 0.05 & 0.05 & 0.05 & 0.05 & 0.05 \\

Warmup
& 5 & 5 & 5 & 5 & 5
& 5 & 5 & 5 & 5 \\

\midrule
\multicolumn{10}{l}{\textbf{Stage 4: Decoder}} \\

Latent input
& Rollout
& Rollout
& Rollout
& Rollout
& Rollout
& Rollout
& Rollout
& Rollout
& Rollout \\

Convolutional channel widths
& \shortstack{204/148\\103/54}
& \shortstack{128/96\\64/32}
& \shortstack{192/128\\96}
& \shortstack{192/128\\96}
& \shortstack{248/168\\112}
& \shortstack{192/128\\64}
& \shortstack{192/128\\96/64}
& \shortstack{384/256\\192/128}
& \shortstack{424/288\\112} \\
Residual blocks & 1 & 1 & 1 & 1 & 2 & 1 & 1 & 1 & 2 \\
Normalization type & LN + GN(8) & LN + GN(8) & LN + GN(8) & LN + GN(8) & LN + GN(8) & LN + GN(8) & LN + GN(8) & LN + GN(8) & LN + GN(8) \\
Activation & GELU & GELU & GELU & GELU & GELU & GELU & GELU & GELU & GELU \\
Spatial scale factors & 2/2 & 2/2 & 2/2 & 2/2 & 2/2 & 2/2 & 1/2/2 & 2/2/2 & 2/2 \\
Spatial padding & Circular & Circular & Circular & Circular & Zeros & Circular & Reflect & Circular & Circular \\
Reconstructed channels & 1 & 1 & 1 & 1 & 1 & 2 & 2 & 1 & 1 \\
Epoch
& 2000
& 2000
& 200
& 200
& 200
& 200
& 200
& 200
& 1000 \\

Global batch
& 96
& 64
& 16
& 16
& 32
& 16
& 32
& 16
& 32 \\

Peak LR
& $2{\times}10^{-4}$
& $10^{-4}$
& $10^{-4}$
& $10^{-4}$
& $2{\times}10^{-4}$
& $10^{-4}$
& $10^{-4}$
& $10^{-4}$
& $3{\times}10^{-4}$ \\

Final LR
& $10^{-5}$
& $10^{-6}$
& $10^{-6}$
& $10^{-6}$
& $10^{-6}$
& $10^{-6}$
& $10^{-6}$
& $10^{-6}$
& $0$ \\

Weight decay
& $10^{-4}$
& 0.04
& 0.04
& 0.04
& $10^{-4}$
& 0.04
& 0.04
& 0.04
& 0.01 \\

Warmup epochs
& 50
& 50
& 10
& 10
& 10
& 10
& 10
& 10
& 50 \\

\end{xltabular}

\endgroup

\subsection{Baseline Implementations}

We provide the implementation details of the baselines used in our experiments. We compare our method against FNO~\citep{li2020fourier}, CAPE~\citep{takamoto2023learning}, CoDA~\citep{kirchmeyer2022generalizing}, GEPS~\citep{kassai2024boosting}, ViT-\textit{in-context}, $[\mathrm{CLS}]$ ViT~\citep{peebles2023scalable}, Zebra~\citep{serrano2024zebra}, UniSolver~\citep{zhou2024unisolver}, MPP~\citep{mccabe2024multiple}, DPOT-S~\citep{hao2024dpot}, Poseidon-T~\citep{herde2024poseidon}, LE-PDE~\citep{wu2022learning}, LNS~\citep{li2025latent}, MAE-PDE~\citep{zhou2024masked}, and ENMA~\citep{kassai2026enma}. For CAPE, CoDA, Zebra, $[\mathrm{CLS}]$ ViT, and ViT-\textit{in-context}, we directly adopt the results reported in Zebra~\citep{serrano2024zebra}. All remaining baselines are implemented and evaluated under our experimental setting, and their implementation details are described below. All reimplemented baselines use the same evaluation trajectories / metric / rollout protocol.

\lead{FNO.}
We concatenated the temporal history along the channel dimension and appended spatial coordinates to the input. We stacked four Fourier layers, each combining a spectral convolution with a pointwise convolution and GELU activation. The 1D models used 32 Fourier modes and a hidden width of 195, whereas the 2D models used 12 modes per spatial axis and a width of 66. We used a two-frame history for the all datasets and generated trajectories autoregressively. The configuration is in Table~\ref{tab:baseline-fno}.

\begin{table*}[h]
\centering
\footnotesize
\setlength{\tabcolsep}{4pt}
\renewcommand{\arraystretch}{1.12}
\caption{\textbf{FNO implementation hyperparameters.}}
\label{tab:baseline-fno}
\resizebox{\ifdim\width>\textwidth\textwidth\else\width\fi}{!}{%
\begin{tabular}{lccccccccc}
\toprule
Hyperparameter & Combined & Advection & Burgers & Heat & Wave-B & GS & Wave2D & Vorticity & HeterNS \\
\midrule
Model input grid & $256$ & $256$ & $256$ & $256$ & $256$ & $32\times32$ & $64\times64$ & $128\times128$ & $64\times64$ \\
State channels & 1 & 1 & 1 & 1 & 1 & 2 & 2 & 1 & 1 \\
Fourier layers & 4 & 4 & 4 & 4 & 4 & 4 & 4 & 4 & 4 \\
Hidden width & 195 & 195 & 195 & 195 & 195 & 66 & 66 & 66 & 66 \\
Modes per spatial axis & 32 & 32 & 32 & 32 & 32 & 12 & 12 & 12 & 12 \\
Input history (frames) & 2 & 2 & 2 & 2 & 2 & 2 & 2 & 2 & 2 \\
Layer normalization & False & False & False & False & False & False & False & False & False \\
Input instance normalization & True & True & True & True & True & True & True & True & True \\
Activation & GELU & GELU & GELU & GELU & GELU & GELU & GELU & GELU & GELU \\
Coordinate features & Grid & Grid & Grid & Grid & Grid & Grid & Grid & Grid & Grid \\
Temporal stride & 10 & 10 & 10 & 10 & 10 & 1 & 1 & 1 & 1 \\
Training epochs (budget) & 500 & 500 & 500 & 500 & 500 & 500 & 500 & 500 & 500 \\
Batch size / GPU & 32 & 32 & 32 & 32 & 32 & 32 & 32 & 16 & 32 \\
Training rollout steps & 1 & 1 & 1 & 1 & 1 & 1 & 2 & 1 & 1 \\
Optimizer & Adam & Adam & Adam & Adam & Adam & Adam & Adam & Adam & Adam \\
Peak learning rate & 0.001 & 0.001 & 0.001 & 0.001 & 0.001 & 0.001 & $10^{-4}$ & 0.001 & 0.001 \\
Weight decay & $10^{-6}$ & $10^{-6}$ & $10^{-6}$ & $10^{-6}$ & $10^{-6}$ & $10^{-6}$ & $10^{-6}$ & $10^{-6}$ & $10^{-6}$ \\
LR schedule & OneCycle & OneCycle & OneCycle & OneCycle & OneCycle & OneCycle & OneCycle & OneCycle & OneCycle \\
Warmup epochs & 40 & 40 & 40 & 40 & 40 & 40 & 20 & 40 & 40 \\
Gradient clipping & 10000 & 10000 & 10000 & 10000 & 10000 & 10000 & 1 & 10000 & 10000 \\
\bottomrule
\end{tabular}}
\par\vspace{5pt}
\begin{minipage}{\textwidth}
\footnotesize
Adam uses $(\beta_1,\beta_2)=(0.9,0.9)$; both OneCycle division factors are $10^4$.
\end{minipage}
\end{table*}

\lead{GEPS.}
We used an FNO backbone whose weights and activations were modulated by a learned four-dimensional environment code. The 1D models contained eight Fourier layers with width 64 and 16 modes; the 2D models contained four Fourier layers with width 16 and 12 modes per spatial axis. The local convolution branches used kernel sizes of 1 in 1D and $7\times7$ in 2D. We included spatial coordinates and used code-conditioned Swish activations. At evaluation, the environment code was adapted using a support trajectory while the shared network parameters remained frozen. The configuration is in Table~\ref{tab:baseline-geps}.

\begingroup
\scriptsize
\setlength{\tabcolsep}{2pt}
\renewcommand{\arraystretch}{1.12}
\setlength\LTleft{0pt}
\setlength\LTright{0pt}
\setlength{\LTpost}{0pt}

\begin{xltabular}{\textwidth}{
@{}
>{\raggedright\arraybackslash}p{0.18\textwidth}
*{9}{>{\centering\arraybackslash}X}
@{}
}
\caption{\textbf{GEPS implementation hyperparameters.}}
\label{tab:baseline-geps}\\

\toprule
Hyperparameter & Combined & Advection & Burgers & Heat & Wave-B &
GS & Wave2D & Vorticity & HeterNS \\
\midrule
\endfirsthead

\multicolumn{10}{c}{%
\tablename\ \thetable{} -- continued from previous page}\\
\toprule
Hyperparameter & Combined & Advection & Burgers & Heat & Wave-B &
GS & Wave2D & Vorticity & HeterNS \\
\midrule
\endhead

\midrule
\multicolumn{10}{r}{Continued on next page}\\
\endfoot

\bottomrule
\endlastfoot

Model input grid
& $128$ & $128$ & $128$ & $128$ & $128$
& $32\times32$ & $64\times64$ & $64\times64$ & $64\times64$ \\

State channels
& 1 & 1 & 1 & 1 & 1 & 2 & 2 & 1 & 1 \\

Fourier layers
& 8 & 8 & 8 & 8 & 8 & 4 & 4 & 4 & 4 \\

Hidden width
& 64 & 64 & 64 & 64 & 64 & 16 & 16 & 16 & 16 \\

Modes per spatial axis
& 16 & 16 & 16 & 16 & 16 & 12 & 12 & 12 & 12 \\

Local convolution kernel
& 1 & 1 & 1 & 1 & 1 & 7 & 7 & 7 & 7 \\

Environment code size
& 4 & 4 & 4 & 4 & 4 & 4 & 4 & 4 & 4 \\

Adaptation factor
& 1 & 1 & 1 & 1 & 1 & 1 & 1 & 1 & 1 \\

Activation
& Swish & Swish & Swish & Swish & Swish
& Swish & Swish & Swish & Swish \\

Coordinate features
& Grid & Grid & Grid & Grid & Grid
& Grid & Grid & Grid & Grid \\

Model input frames
& 1 & 1 & 1 & 1 & 1 & 1 & 1 & 1 & 1 \\

Training epochs
& 20000 & 20000 & 20000 & 20000 & 20000 & 20000 & 20000 & 20000 & 20000 \\

Batch size
& 64 & 64 & 64 & 64 & 64 & 64 & 64 & 64 & 256 \\

Optimizer
& Adam & Adam & Adam & Adam & Adam
& Adam & Adam & Adam & Adam \\

Shared-weight LR
& 0.001 & 0.001 & 0.001 & 0.001 & 0.001
& 0.001 & 0.001 & 0.001 & 0.001 \\

Environment-code LR
& 0.01 & 0.01 & 0.01 & 0.01 & 0.01
& 0.01 & 0.01 & 0.01 & 0.01 \\

Min. learning rate
& $10^{-5}$ & $10^{-5}$ & $10^{-5}$ & $10^{-5}$ & $10^{-5}$
& $10^{-5}$ & $10^{-5}$ & $10^{-5}$ & $10^{-5}$ \\

LR schedule
& Cosine & Cosine & Cosine & Cosine & Cosine
& Cosine & Cosine & Cosine & Cosine \\

Rollout curriculum (steps)
& 1/2/4 & 1/2/4 & 1/2/4 & 1/2/4 & 1/2/4
& 1/2/4 & 1/2/4 & 1/2/4 & 1/2/4 \\

Gradient clipping
& 1 & 1 & 1 & 1 & 1 & 1 & 1 & 1 & 1 \\

Adaptation epochs
& 500 & 500 & 500 & 500 & 500 & 500 & 500 & 500 & 500 \\

Adaptation batch size
& 32 & 32 & 12 & 12 & 32 & 32 & 32 & 120 & 64 \\

Adaptation learning rate
& 0.01 & 0.01 & 0.01 & 0.01 & 0.01
& 0.01 & 0.01 & 0.01 & 0.01 \\

Adaptation rollout steps
& 4 & 4 & 4 & 4 & 4 & 4 & 4 & 4 & 4 \\

\end{xltabular}
\noindent
{\footnotesize Adam uses $(\beta_1,\beta_2)=(0.9,0.9)$.}

\endgroup

\lead{Unisolver.}
We used a PDE-conditioned transformer with ten blocks, a hidden size of 256, eight attention heads, and a feed-forward width of 1008. Domain-level PDE parameters and pointwise geometric features modulated the transformer through separate AdaLN-Zero conditioning branches. Geometry was represented by distances to a reference grid. We used patches of 16 points for 1D inputs, $2\times2$ patches for Gray--Scott, and $4\times4$ patches for Wave2D and Vorticity. Given the current field, the model predicted a residual update that was added to the input state for autoregressive forecasting. The configuration is in Table~\ref{tab:baseline-unisolver}.

\begingroup

\scriptsize
\setlength{\tabcolsep}{2pt}
\renewcommand{\arraystretch}{1.08}

\setlength{\LTleft}{0pt}
\setlength{\LTright}{0pt}
\setlength{\LTpost}{0pt}

\begin{xltabular}{\textwidth}{
@{}
>{\raggedright\arraybackslash}p{0.20\textwidth}
*{8}{>{\centering\arraybackslash}X}
@{}
}

\caption{\textbf{UniSolver implementation hyperparameters.}}
\label{tab:baseline-unisolver}\\

\toprule
Hyperparameter
& Combined
& Advection
& Burgers
& Heat
& Wave-B
& GS
& Wave2D
& Vorticity \\
\midrule
\endfirsthead

\multicolumn{9}{c}{
\tablename\ \thetable{} -- continued from previous page
}\\
\toprule
Hyperparameter
& Combined
& Advection
& Burgers
& Heat
& Wave-B
& GS
& Wave2D
& Vorticity \\
\midrule
\endhead

\midrule
\multicolumn{9}{r}{Continued on next page}\\
\endfoot

\bottomrule
\endlastfoot

Model input grid
& $128$ & $128$ & $128$ & $128$ & $128$
& $32\times32$ & $64\times64$ & $64\times64$ \\

Spatial patch size
& $16$ & $16$ & $16$ & $16$ & $16$
& $2\times2$ & $4\times4$ & $4\times4$ \\

State channels
& 1 & 1 & 1 & 1 & 1 & 2 & 2 & 1 \\

Transformer depth
& 10 & 10 & 10 & 10 & 10 & 10 & 10 & 10 \\

Hidden size
& 256 & 256 & 256 & 256 & 256 & 256 & 256 & 256 \\

MLP ratio
& 3.9375 & 3.9375 & 3.9375 & 3.9375
& 3.9375 & 3.9375 & 3.9375 & 3.9375 \\

Attention heads
& 8 & 8 & 8 & 8 & 8 & 8 & 8 & 8 \\

Dropout
& 0 & 0 & 0 & 0 & 0 & 0 & 0 & 0 \\

Block normalization
& AdaLN & AdaLN & AdaLN & AdaLN
& AdaLN & AdaLN & AdaLN & AdaLN \\

MLP activation
& GELU & GELU & GELU & GELU
& GELU & GELU & GELU & GELU \\

Position / geometry
& Ref. dist. & Ref. dist. & Ref. dist. & Ref. dist.
& Ref. dist. & Ref. dist. & Ref. dist. & Ref. dist. \\

Reference grid
& $16$ & $16$ & $16$ & $16$ & $16$
& $8\times8$ & $4\times4$ & $4\times4$ \\

Input history (frames)
& 1 & 1 & 1 & 1 & 1 & 1 & 1 & 1 \\

Residual prediction
& True & True & True & True
& True & True & True & True \\

Training epochs
& 100 & 100 & 100 & 100 & 100 & 100 & 100 & 100 \\

Batch size / GPU
& 64 & 64 & 64 & 64 & 64 & 64 & 64 & 64 \\

Optimizer
& AdamW & AdamW & AdamW & AdamW
& AdamW & AdamW & AdamW & AdamW \\

Peak learning rate
& $3\!\times\!10^{-4}$
& $3\!\times\!10^{-4}$
& $3\!\times\!10^{-4}$
& $3\!\times\!10^{-4}$
& $3\!\times\!10^{-4}$
& $3\!\times\!10^{-4}$
& $3\!\times\!10^{-4}$
& $3\!\times\!10^{-4}$ \\

Weight decay
& $10^{-4}$ & $10^{-4}$ & $10^{-4}$ & $10^{-4}$
& $10^{-4}$ & $10^{-4}$ & $10^{-4}$ & $10^{-4}$ \\

LR schedule
& Cosine & Cosine & Cosine & Cosine
& Cosine & Cosine & Cosine & Cosine \\

Warmup epochs
& 5 & 5 & 5 & 5 & 5 & 5 & 5 & 5 \\

Gradient clipping
& 1 & 1 & 1 & 1 & 1 & 1 & 1 & 1 \\

\end{xltabular}

\endgroup

\lead{MPP.}
We used the AViT-B~\citep{muller2022instant} architecture with twelve blocks that alternated temporal attention and axial spatial attention. The model used a hidden size of 768, twelve attention heads, an MLP ratio of 4, and $16\times16$ spatial patches. A learned projection embedded the physical state variables, and a patch decoder reconstructed the next field from up to sixteen history frames. For 1D inputs, we repeated the field along a synthetic axis of length 16 to preserve the 2D patch architecture. We used relative positional biases, LayerNorm on queries and keys, and a maximum drop-path rate of 0.1. All reported runs were trained from scratch. We use DAdaptAdan, a family of Adan~\citep{defazio2023learning} as optimizer. The configuration is in Table~\ref{tab:baseline-mpp}.

\begin{table*}[h]
\centering
\footnotesize
\setlength{\tabcolsep}{2pt}
\renewcommand{\arraystretch}{1.15}
\caption{MPP implementation hyperparameters.}
\label{tab:baseline-mpp}

\resizebox{\ifdim\width>\textwidth\textwidth\else\width\fi}{!}{%
\begin{tabular}{lccccccccc}
\toprule
Hyperparameter & Combined & Advection & Burgers & Heat & Wave-B & GS & Wave2D & Vorticity & HeterNS \\
\midrule

Spatial patch size
& $16$ & $16$ & $16$ & $16$ & $16$
& $16\times16$ & $16\times16$ & $16\times16$ & $16\times16$ \\

Space--time blocks
& 12 & 12 & 12 & 12 & 12 & 12 & 12 & 12 & 12 \\

Hidden size
& 768 & 768 & 768 & 768 & 768 & 768 & 768 & 768 & 768 \\

MLP ratio
& 4 & 4 & 4 & 4 & 4 & 4 & 4 & 4 & 4 \\

Attention heads
& 12 & 12 & 12 & 12 & 12 & 12 & 12 & 12 & 12 \\

Attention dropout
& 0 & 0 & 0 & 0 & 0 & 0 & 0 & 0 & 0 \\

Max. drop-path rate
& 0.1 & 0.1 & 0.1 & 0.1 & 0.1 & 0.1 & 0.1 & 0.1 & 0.1 \\

QK normalization
& LN & LN & LN & LN & LN & LN & LN & LN & LN \\

Spatial norm denominator
& std-div. & std-div. & std-div. & std-div. & std-div.
& RMS & std-div. & std-div. & std-div. \\

Temporal normalization
& IN & IN & IN & IN & IN & IN & IN & IN & IN \\

Activation
& GELU & GELU & GELU & GELU & GELU & GELU & GELU & GELU & GELU \\

Positional bias
& Relative & Relative & Relative & Relative & Relative
& Relative & Relative & Relative & Relative \\

Max. input history
& 16 & 16 & 16 & 16 & 16 & 16 & 16 & 16 & 16 \\

State channels
& 1 & 1 & 1 & 1 & 1 & 2 & 2 & 1 & 1 \\

Initialization
& Scratch & Scratch & Scratch & Scratch & Scratch
& Scratch & Scratch & Scratch & Scratch \\

Training epochs
& 500 & 500 & 500 & 500 & 500 & 500 & 500 & 500 & 500 \\

Batch size / process
& 16 & 16 & 16 & 16 & 16 & 16 & 16 & 8 & 16 \\

Global batch size
& 16 & 16 & 16 & 16 & 16 & 16 & 16 & 16 & 16 \\

Steps per epoch
& 125 & 125 & 125 & 125 & 125 & 125 & 125 & 125 & 125 \\

Gradient accumulation
& 1 & 1 & 1 & 1 & 1 & 1 & 1 & 1 & 1 \\

Optimizer
& DAdaptAdan & DAdaptAdan & DAdaptAdan & DAdaptAdan & DAdaptAdan
& DAdaptAdan & DAdaptAdan & DAdaptAdan & DAdaptAdan \\

LR multiplier
& 1 & 1 & 1 & 1 & 1 & 1 & 1 & 1 & 1 \\

Weight decay
& 0.001 & 0.001 & 0.001 & 0.001 & 0.001
& 0.001 & 0.001 & 0.001 & 0.001 \\

LR schedule
& Cosine & Cosine & Cosine & Cosine & Cosine
& Cosine & Cosine & Cosine & Cosine \\

Applied warmup steps
& 50 & 50 & 50 & 50 & 50 & 50& 50 & 50 & 50 \\

\bottomrule
\end{tabular}
}

\vspace{2pt}
\begin{minipage}{\textwidth}
\footnotesize
LN denotes LayerNorm and IN denotes InstanceNorm.
\end{minipage}
\end{table*}

\lead{DPOT-S.}
We fine-tuned the pretrained DPOT-S backbone on the target PDE datasets. We retained six adaptive Fourier neural operator blocks with a hidden size of 1024, eight Fourier channel groups, and an MLP ratio of 1. Inputs were represented on a $128\times128$ grid and embedded using $8\times8$ spatial patches. Learned positional embeddings and temporal aggregation combined the ten-frame input buffer before Fourier processing. The model used GroupNorm and GELU, followed by a convolutional upsampling head, and was fine-tuned using the target solution trajectories. The configuration is in Table~\ref{tab:baseline-dpot}.

\begingroup

\scriptsize
\setlength{\tabcolsep}{2pt}
\renewcommand{\arraystretch}{1.08}

\setlength{\LTleft}{0pt}
\setlength{\LTright}{0pt}
\setlength{\LTpost}{0pt}

\begin{xltabular}{\textwidth}{
@{}
>{\raggedright\arraybackslash}p{0.20\textwidth}
*{9}{>{\centering\arraybackslash}X}
@{}
}

\caption{\textbf{DPOT implementation hyperparameters.}}
\label{tab:baseline-dpot}\\

\toprule
Hyperparameter
& Combined
& Advection
& Burgers
& Heat
& Wave-B
& GS
& Wave2D
& Vorticity
& HeterNS \\
\midrule
\endfirsthead

\multicolumn{10}{c}{
\tablename\ \thetable{} -- continued from previous page
}\\
\toprule
Hyperparameter
& Combined
& Advection
& Burgers
& Heat
& Wave-B
& GS
& Wave2D
& Vorticity
& HeterNS \\
\midrule
\endhead

\midrule
\multicolumn{10}{r}{Continued on next page}\\
\endfoot

\bottomrule
\endlastfoot

Model input grid
& $128\times128$
& $128\times128$
& $128\times128$
& $128\times128$
& $128\times128$
& $128\times128$
& $128\times128$
& $128\times128$
& $128\times128$ \\

Spatial patch size
& $8\times8$
& $8\times8$
& $8\times8$
& $8\times8$
& $8\times8$
& $8\times8$
& $8\times8$
& $8\times8$
& $8\times8$ \\

AFNO depth
& 6 & 6 & 6 & 6 & 6 & 6 & 6 & 6 & 6 \\

Hidden size
& 1024 & 1024 & 1024 & 1024 & 1024
& 1024 & 1024 & 1024 & 1024 \\

MLP ratio
& 1 & 1 & 1 & 1 & 1 & 1 & 1 & 1 & 1 \\

Fourier blocks
& 8 & 8 & 8 & 8 & 8 & 8 & 8 & 8 & 8 \\

Fourier modes (configured)
& 32 & 32 & 32 & 32 & 32 & 32 & 32 & 32 & 32 \\

Padded state channels
& 4 & 4 & 4 & 4 & 4 & 4 & 4 & 4 & 4 \\

History buffer (frames)
& 10 & 10 & 10 & 10 & 10 & 10 & 10 & 10 & 10 \\

Initial true frames
& 1 & 1 & 1 & 1 & 1 & 1 & 1 & 1 & 1 \\

Block normalization
& GN & GN & GN & GN & GN & GN & GN & GN & GN \\

Normalization groups
& 8 & 8 & 8 & 8 & 8 & 8 & 8 & 8 & 8 \\

Activation
& GELU & GELU & GELU & GELU & GELU
& GELU & GELU & GELU & GELU \\

Positional embedding
& Learned & Learned & Learned & Learned & Learned
& Learned & Learned & Learned & Learned \\

Initialization
& DPOT-S & DPOT-S & DPOT-S & DPOT-S & DPOT-S
& DPOT-S & DPOT-S & DPOT-S & DPOT-S \\

Training epochs
& 200 & 200 & 200 & 200 & 200 & 200 & 200 & 200 & 200 \\

Batch size / GPU
& 16 & 16 & 16 & 16 & 16 & 16 & 16 & 16 & 16 \\

Optimizer
& Adam & Adam & Adam & Adam & Adam
& Adam & Adam & Adam & Adam \\

Peak learning rate
& 0.001 & 0.001 & 0.001 & 0.001 & 0.001
& 0.001 & 0.001 & 0.001 & 0.001 \\

Weight decay
& $10^{-6}$ & $10^{-6}$ & $10^{-6}$ & $10^{-6}$ & $10^{-6}$
& $10^{-6}$ & $10^{-6}$ & $10^{-6}$ & $10^{-6}$ \\

LR schedule
& OneCycle & OneCycle & OneCycle & OneCycle & OneCycle
& OneCycle & OneCycle & OneCycle & OneCycle \\

Warmup epochs
& 40 & 40 & 40 & 40 & 40 & 40 & 40 & 40 & 40 \\

Gradient clipping
& 10000 & 10000 & 10000 & 10000 & 10000
& 10000 & 10000 & 10000 & 10000 \\

\end{xltabular}

\noindent
{\footnotesize
DPOT-S is initialized from the saved pretrained checkpoint
\texttt{model\_S.pth}. All tasks are adapted to a $128\times128$
input with four state slots; 1D signals are lifted to the 2D input
representation. The ten-frame network buffer is initialized by repeating
the single available initial frame. The eight Fourier blocks are
AFNO~\citep{guibas2021efficient} channel groups, not attention heads.
The nominal mode count is 32. Adam uses $(0.9,0.9)$, and both
OneCycle division factors are $10^4$.
}

\endgroup

\lead{Poseidon-T.}
We initialized the ScOT encoder--decoder from the pretrained Poseidon-T checkpoint and fine-tuned it on the target PDE datasets. The model used four resolution stages in each branch, with four transformer blocks per stage and skip connections between corresponding stages. The stage widths H was $48,96,192,384$, with $3,6,12,24$ attention heads, respectively. We retained $4\times4$ patches, an MLP ratio of 4, shifted-window attention with a configured window size of 16, and time-conditioned LayerNorm. Input and output projections were adapted to the required channels, and the transformer backbone remained trainable during fine-tuning.
 The configuration is in Table~\ref{tab:baseline-poseidon}.

\begingroup
\scriptsize
\setlength{\tabcolsep}{2pt}
\renewcommand{\arraystretch}{1.12}

\setlength{\LTleft}{0pt}
\setlength{\LTright}{0pt}
\setlength{\LTpost}{2pt}

\begin{xltabular}{\textwidth}{
@{}
>{\raggedright\arraybackslash}p{0.20\textwidth}
*{9}{>{\centering\arraybackslash}X}
@{}
}
\caption{\textbf{Poseidon implementation hyperparameters.}}
\label{tab:baseline-poseidon}\\

\toprule
Hyperparameter
& Combined
& Advection
& Burgers
& Heat
& Wave-B
& GS
& Wave2D
& Vorticity
& HeterNS \\
\midrule
\endfirsthead

\multicolumn{10}{c}{%
\tablename\ \thetable{} -- continued from previous page}\\
\toprule
Hyperparameter
& Combined
& Advection
& Burgers
& Heat
& Wave-B
& GS
& Wave2D
& Vorticity
& HeterNS \\
\midrule
\endhead

\midrule
\multicolumn{10}{r}{Continued on next page}\\
\endfoot

\bottomrule
\endlastfoot

Model input grid
& $64\times64$ & $64\times64$ & $64\times64$ & $64\times64$ & $64\times64$
& $64\times64$ & $64\times64$ & $128\times128$ & $64\times64$ \\

Spatial patch size
& $4\times4$ & $4\times4$ & $4\times4$ & $4\times4$ & $4\times4$
& $4\times4$ & $4\times4$ & $4\times4$ & $4\times4$ \\

Encoder / decoder stages
& 4 / 4 & 4 / 4 & 4 / 4 & 4 / 4 & 4 / 4
& 4 / 4 & 4 / 4 & 4 / 4 & 4 / 4 \\

Depths per branch
& 4/4/4/4 & 4/4/4/4 & 4/4/4/4 & 4/4/4/4 & 4/4/4/4
& 4/4/4/4 & 4/4/4/4 & 4/4/4/4 & 4/4/4/4 \\

Stage hidden sizes
& H & H & H
& H & H & H
& H & H & H \\

Stage attention heads
& 3/6/12/24 & 3/6/12/24 & 3/6/12/24
& 3/6/12/24 & 3/6/12/24 & 3/6/12/24
& 3/6/12/24 & 3/6/12/24 & 3/6/12/24 \\

MLP ratio
& 4 & 4 & 4 & 4 & 4 & 4 & 4 & 4 & 4 \\

Window size (configured)
& 16 & 16 & 16 & 16 & 16 & 16 & 16 & 16 & 16 \\

Input channels (incl. PDE)
& 4 & 2 & 3 & 3 & 6 & 4 & 4 & 2 & 3 \\

Output channels
& 1 & 1 & 1 & 1 & 1 & 2 & 2 & 1 & 1 \\

Attention / hidden dropout
& 0 / 0 & 0 / 0 & 0 / 0 & 0 / 0 & 0 / 0
& 0 / 0 & 0 / 0 & 0 / 0 & 0 / 0 \\

Drop-path rate
& 0 & 0 & 0 & 0 & 0 & 0 & 0 & 0 & 0 \\

QK normalization
& L2 cosine & L2 cosine & L2 cosine & L2 cosine & L2 cosine
& L2 cosine & L2 cosine & L2 cosine & L2 cosine \\

Block normalization
& Cond. LN & Cond. LN & Cond. LN & Cond. LN & Cond. LN
& Cond. LN & Cond. LN & Cond. LN & Cond. LN \\

Activation
& GELU & GELU & GELU & GELU & GELU
& GELU & GELU & GELU & GELU \\

Positional bias
& Cont. rel. & Cont. rel. & Cont. rel. & Cont. rel. & Cont. rel.
& Cont. rel. & Cont. rel. & Cont. rel. & Cont. rel. \\

Initialization
& Poseidon-T & Poseidon-T & Poseidon-T & Poseidon-T & Poseidon-T
& Poseidon-T & Poseidon-T & Poseidon-T & Poseidon-T \\

Training epochs
& 100 & 100 & 100 & 100 & 100
& 100 & 100 & 100 & 100 \\

Batch size
& 64 & 64 & 64 & 64 & 64
& 64 & 64 & 64 & 64 \\

Optimizer
& AdamW & AdamW & AdamW & AdamW & AdamW
& AdamW & AdamW & AdamW & AdamW \\

Backbone / head LR
& $5\!\times\!10^{-4}$ & $5\!\times\!10^{-4}$ & $5\!\times\!10^{-4}$
& $5\!\times\!10^{-4}$ & $5\!\times\!10^{-4}$ & $5\!\times\!10^{-4}$
& $5\!\times\!10^{-4}$ & $5\!\times\!10^{-4}$ & $5\!\times\!10^{-4}$ \\

Weight decay
& $10^{-6}$ & $10^{-6}$ & $10^{-6}$ & $10^{-6}$ & $10^{-6}$
& $10^{-6}$ & $10^{-6}$ & $10^{-6}$ & $10^{-6}$ \\

LR schedule
& Cosine & Cosine & Cosine & Cosine & Cosine
& Cosine & Cosine & Cosine & Cosine \\

Warmup fraction
& 0.05 & 0.05 & 0.05 & 0.05 & 0.05
& 0.05 & 0.05 & 0.05 & 0.05 \\

Gradient clipping
& 5 & 5 & 5 & 5 & 5
& 5 & 5 & 5 & 5 \\

\end{xltabular}

\noindent
\begin{minipage}{\textwidth}
\footnotesize
All models are initialized from the pretrained Poseidon-T checkpoint and
fine-tuned on the corresponding target PDE dataset. L2 cosine denotes query/key unit normalization in SwinV2
attention~\citep{liu2022swin}. Cont. rel. denotes continuous relative positional
bias, with absolute positional embeddings disabled. Cond. LN denotes
time-conditioned LayerNorm. The same learning rate is used for the backbone
and task-specific input/output projections, and the transformer backbone
remains trainable during fine-tuning. The 256-point 1D signals are arranged
on a $16\times16$ serpentine grid and resized to $64\times64$ before being
passed to the model.
\end{minipage}

\endgroup
\lead{LE-PDE.}
We used a convolutional autoencoder with four downsampling blocks, GroupNorm as GN, and ELU activations~\citep{clevert2015fast}. The dynamic encoder maps the input history to a 256-dimensional latent, while a context encoder extracts a 64-dimensional static code from another trajectory in the same environment. Their concatenation is evolved by a five-layer residual MLP with hidden width 256, and the decoder reconstructs the physical field. The configuration is given in Table~\ref{tab:baseline-lepde}.

\begingroup
\scriptsize
\setlength{\tabcolsep}{2pt}
\renewcommand{\arraystretch}{1.12}

\setlength{\LTleft}{0pt}
\setlength{\LTright}{0pt}
\setlength{\LTpost}{2pt}

\begin{xltabular}{\textwidth}{
@{}
>{\raggedright\arraybackslash}p{0.20\textwidth}
*{9}{>{\centering\arraybackslash}X}
@{}
}
\caption{\textbf{LE-PDE implementation hyperparameters.}}
\label{tab:baseline-lepde}\\

\toprule
Hyperparameter
& Combined
& Advection
& Burgers
& Heat
& Wave-B
& GS
& Wave2D
& Vorticity
& HeterNS \\
\midrule
\endfirsthead

\multicolumn{10}{c}{%
\tablename\ \thetable{} -- continued from previous page}\\
\toprule
Hyperparameter
& Combined
& Advection
& Burgers
& Heat
& Wave-B
& GS
& Wave2D
& Vorticity
& HeterNS \\
\midrule
\endhead

\midrule
\multicolumn{10}{r}{Continued on next page}\\
\endfoot

\bottomrule
\endlastfoot

Model input grid
& $256$ & $256$ & $256$ & $256$ & $256$
& $32\times32$ & $64\times64$ & $64\times64$ & $64\times64$ \\

State channels
& 1 & 1 & 1 & 1 & 1 & 2 & 2 & 1 & 1 \\

Input history (frames)
& 1 & 1 & 1 & 1 & 1 & 1 & 1 & 1 & 10 \\

Context trajectory frames
& 10 & 10 & 10 & 10 & 15 & 7 & 10 & 10 & 20 \\

Encoder downsampling blocks
& 4 & 4 & 4 & 4 & 4 & 4 & 4 & 4 & 4 \\

Base convolution width
& 16 & 16 & 16 & 16 & 16 & 16 & 16 & 16 & 16 \\

Dynamic latent size
& 256 & 256 & 256 & 256 & 256 & 256 & 256 & 256 & 256 \\

Static latent size
& 64 & 64 & 64 & 64 & 64 & 64 & 64 & 64 & 64 \\

Evolution MLP linear layers
& 5 & 5 & 5 & 5 & 5 & 5 & 5 & 5 & 5 \\

Evolution hidden size
& 256 & 256 & 256 & 256 & 256 & 256 & 256 & 256 & 256 \\

Activation
& ELU & ELU & ELU & ELU & ELU & ELU & ELU & ELU & ELU \\

AE normalization
& GN & GN & GN & GN & GN
& GN & GN & GN & GN \\

Latent noise amplitude
& $10^{-5}$ & $10^{-5}$ & $10^{-5}$ & $10^{-5}$ & $10^{-5}$
& $10^{-5}$ & $10^{-5}$ & $10^{-5}$ & $10^{-5}$ \\

One-shot context
& True & True & True & True & True & True & True & True & True \\

AE pretraining epochs
& 20 & 20 & 20 & 20 & 20 & 20 & 20 & 20 & 20 \\

Subsequent training epochs
& 200 & 200 & 200 & 200 & 200 & 200 & 200 & 200 & 200 \\

Initially frozen AE epochs
& 20 & 20 & 20 & 20 & 20 & 20 & 20 & 20 & 20 \\

Batch size
& 64 & 64 & 64 & 64 & 64 & 20 & 20 & 20 & 12 \\

Optimizer
& Adam & Adam & Adam & Adam & Adam & Adam & Adam & Adam & Adam \\

Dynamics learning rate
& 0.001 & 0.001 & 0.001 & 0.001 & 0.001
& 0.001 & 0.001 & 0.001 & 0.001 \\

AE fine-tuning LR
& $10^{-5}$ & 0.001 & 0.001 & $10^{-5}$ & $10^{-5}$
& $10^{-5}$ & $10^{-5}$ & $10^{-5}$ & $10^{-5}$ \\

Weight decay
& 0 & 0 & 0 & 0 & 0 & 0 & 0 & 0 & 0 \\

LR schedule
& Cosine & Cosine & Cosine & Cosine & Cosine
& Cosine & Cosine & Cosine & Cosine \\

Gradient clipping
& 1 & 1 & 1 & 1 & 1 & 1 & 1 & 1 & 1 \\

Prediction loss weights
& 1/.1/.1/.1 & 1/.1/.1/.1 & 1/.1/.1/.1
& 1/.1/.1/.1 & 1/.1/.1/.1 & 1/.1/.1/.1
& 1/.1/.1/.1 & 1/.1/.1/.1 & 1/.1/.1/.1 \\

Latent loss weights
& 1/1/1/1 & 1/1/1/1 & 1/1/1/1
& 1/1/1/1 & 1/1/1/1 & 1/1/1/1
& 1/1/1/1 & 1/1/1/1 & 1/1/1/1 \\

\end{xltabular}

\noindent
\begin{minipage}{\textwidth}
\footnotesize
The encoder uses four spatial downsampling blocks, GroupNorm as GN, and ELU.
The latent evolution network has three Linear--ELU pairs followed by two additional linear layers.
A different trajectory from the same environment supplies the one-shot context.
The prediction and latent losses use four rollout steps with the listed weights;
reconstruction and consistency coefficients are both 1.
\end{minipage}

\endgroup

\lead{LNS.}
We used a convolutional autoencoder with a spatially structured latent space and a parameter-conditioned convolutional propagator. The 1D encoder maps 256 points to 16 latent sites with four channels, while the 2D encoder downsamples each axis by 8 with 64 latent channels. A three-block conditional residual propagator evolves the latent state using Fourier-embedded PDE parameters, with the pretrained autoencoder frozen during dynamics training. The configuration is given in Table~\ref{tab:baseline-lns}.

\begingroup
\scriptsize
\setlength{\tabcolsep}{2pt}
\renewcommand{\arraystretch}{1.12}

\setlength{\LTleft}{0pt}
\setlength{\LTright}{0pt}
\setlength{\LTpost}{0pt}

\begin{xltabular}{\textwidth}{
@{}
>{\raggedright\arraybackslash}p{0.20\textwidth}
*{9}{>{\centering\arraybackslash}X}
@{}
}
\caption{\textbf{LNS implementation hyperparameters.}}
\label{tab:baseline-lns}\\

\toprule
Hyperparameter
& Combined
& Advection
& Burgers
& Heat
& Wave-B
& GS
& Wave2D
& Vorticity
& HeterNS \\
\midrule
\endfirsthead

\multicolumn{10}{c}{%
\tablename\ \thetable{} -- continued from previous page}\\
\toprule
Hyperparameter
& Combined
& Advection
& Burgers-F
& Heat
& Wave-B
& GS
& Wave2D
& Vorticity
& HeterNS \\
\midrule
\endhead

\midrule
\multicolumn{10}{r}{Continued on next page}\\
\endfoot

\bottomrule
\endlastfoot

Model input grid
& $256$ & $256$ & $256$ & $256$ & $256$
& $32\times32$ & $64\times64$ & $128\times128$ & $64\times64$ \\

State channels
& 1 & 1 & 1 & 1 & 1 & 2 & 2 & 1 & 1 \\

Latent spatial grid
& $8$ & $8$ & $8$ & $8$ & $8$
& $4\times4$ & $8\times8$ & $8\times8$ & $8\times8$ \\

Latent channels
& 16 & 16 & 16 & 16 & 16 & 64 & 64 & 64 & 64 \\

Encoder channel schedule
& E1 & E1 & E1 & E1 & E1 & E2 & E2 & E2 & E2 \\

Decoder channel schedule
& D1 & D1 & D1 & D1 & D1 & D2 & D2 & D2 & D2 \\

AE residual blocks / level
& 1 & 1 & 1 & 1 & 1 & 1 & 1 & 1 & 1 \\

AE attention heads
& 8 & 8 & 8 & 8 & 8 & 8 & 8 & 8 & 8 \\

AE attention head size
& 64 & 64 & 64 & 64 & 64 & 64 & 64 & 64 & 64 \\

Dynamics blocks
& 3 & 3 & 3 & 3 & 3 & 3 & 3 & 3 & 3 \\

Dynamics hidden size
& 128 & 128 & 128 & 128 & 128 & 128 & 128 & 128 & 128 \\

Condition input size
& 3 & 1 & 3 & 21 & 2 & 2 & 2 & 1 & 2 \\

Condition embedding size
& 96 & 64 & 84 & 84 & 64 & 64 & 64 & 64 & 64 \\

Dynamics dilation
& 2 & 2 & 2 & 2 & 2 & 2 & 2 & 2 & 2 \\

Periodic geometry
& True & True & True & True & False & True & False & True & True \\

AE / dynamics epochs
& 100 / 100 & 100 / 100 & 100 / 100 & 100 / 100 & 100 / 100
& 100 / 100 & 100 / 100 & 100 / 100 & 100 / 100 \\

AE batch / GPU
& 8 & 8 & 8 & 8 & 8 & 8 & 8 & 8 & 8 \\

AE frames / trajectory
& 14 & 14 & 25 & 25 & 25 & 20 & 30 & 30 & 20 \\

Dynamics batch / GPU
& 100 & 100 &100 & 100 & 100 & 100 & 100 & 100 & 100 \\

Optimizer
& Adam & Adam & Adam & Adam & Adam & Adam & Adam & Adam & Adam \\

AE learning rate
& $3\!\times\!10^{-4}$
& $3\!\times\!10^{-4}$
&$3\!\times\!10^{-4}$
& $3\!\times\!10^{-4}$
& $3\!\times\!10^{-4}$
& $3\!\times\!10^{-4}$
& $3\!\times\!10^{-4}$
& $3\!\times\!10^{-4}$
& $3\!\times\!10^{-4}$ \\

Dynamics learning rate
& $5\!\times\!10^{-4}$
& $5\!\times\!10^{-4}$
& $5\!\times\!10^{-4}$
& $5\!\times\!10^{-4}$
& $5\!\times\!10^{-4}$
& $5\!\times\!10^{-4}$
& $5\!\times\!10^{-4}$
& $5\!\times\!10^{-4}$
& $5\!\times\!10^{-4}$ \\

Minimum learning rate
& $10^{-6}$ & $10^{-6}$ & $10^{-6}$ & $10^{-6}$ & $10^{-6}$
& $10^{-6}$ & $10^{-6}$ & $10^{-6}$ & $10^{-6}$ \\

Weight decay
& 0 & 0 & 0 & 0 & 0 & 0 & 0 & 0 & 0 \\

LR schedule
& Cosine & Cosine & Cosine & Cosine & Cosine
& Cosine & Cosine & Cosine & Cosine \\

Gradient clipping
& 1 & 1 & 1 & 1 & 1 & 1 & 1 & 1 & 1 \\

\end{xltabular}

\noindent
{\footnotesize
E1=$[64,64,64,128,128,128]$;
D1=$[128,128,128,64,64]$;
E2=$[64,64,64,128,128]$;
D2=$[128,128,64,64]$.
AE Adam betas are $(0.5,0.9)$; dynamics use Adam defaults $(0.9,0.999)$.
Evaluation is recursive from one true initial frame.
}

\endgroup

\lead{MAE-PDE.} we pretrained a masked autoencoder on five-frame solution windows. Its ViT encoder used four layers, hidden size 192, six attention heads, an MLP ratio of 2, and learned positional embeddings. The time--space patches were $1\times8$ in 1D, $1\times4\times4$ for Gray--Scott, and $1\times8\times8$ for the other 2D datasets. We masked 75\% of tokens in 1D and 90\% in 2D and used a two-layer reconstruction decoder with width 64 and four heads. For forecasting, we froze the encoder and projected its representation to a 64-dimensional condition for an FNO with four layers and width 48. The forecaster used 24 modes in 1D, eight per axis for Gray--Scott, and twelve per axis for the other 2D datasets, predicting residual updates from the five-frame history. The configuration is given in Table~\ref{tab:baseline-maepde}.

 \begingroup

\scriptsize
\setlength{\tabcolsep}{2pt}
\renewcommand{\arraystretch}{1.08}

\setlength{\LTleft}{0pt}
\setlength{\LTright}{0pt}
\setlength{\LTpost}{0pt}

\begin{xltabular}{\textwidth}{
@{}
>{\raggedright\arraybackslash}p{0.20\textwidth}
*{9}{>{\centering\arraybackslash}X}
@{}
}

\caption{\textbf{MAE-PDE implementation hyperparameters.}}
\label{tab:baseline-maepde}\\

\toprule
Hyperparameter
& Combined
& Advection
& Burgers
& Heat
& Wave-B
& GS
& Wave2D
& Vorticity
& HeterNS \\
\midrule
\endfirsthead

\multicolumn{10}{c}{
\tablename\ \thetable{} -- continued from previous page
}\\
\toprule
Hyperparameter
& Combined
& Advection
& Burgers
& Heat
& Wave-B
& GS
& Wave2D
& Vorticity
& HeterNS \\
\midrule
\endhead

\midrule
\multicolumn{10}{r}{Continued on next page}\\
\endfoot

\bottomrule
\endlastfoot

Model input grid
& $256$ & $256$ & $256$ & $256$ & $256$
& $32\times32$ & $64\times64$ & $64\times64$ & $64\times64$ \\

State channels
& 1 & 1 & 1 & 1 & 1 & 2 & 2 & 1 & 1 \\

Patch size (time, space)
& $1\times8$
& $1\times8$
& $1\times8$
& $1\times8$
& $1\times8$
& $1\times4\times4$
& $1\times8\times8$
& $1\times8\times8$
& $1\times8\times8$ \\

Input history (frames)
& 5 & 5 & 5 & 5 & 5
& 5 & 5 & 5 & 5 \\

Transformer depth
& 4 & 4 & 4 & 4 & 4
& 4 & 4 & 4 & 4 \\

Hidden size
& 192 & 192 & 192 & 192 & 192
& 192 & 192 & 192 & 192 \\

MLP ratio
& 2 & 2 & 2 & 2 & 2
& 2 & 2 & 2 & 2 \\

Attention heads
& 6 & 6 & 6 & 6 & 6
& 6 & 6 & 6 & 6 \\

Attention head size
& 32 & 32 & 32 & 32 & 32
& 32 & 32 & 32 & 32 \\

Dropout
& 0 & 0 & 0 & 0 & 0
& 0 & 0 & 0 & 0 \\

Block normalization
& Layer\allowbreak Norm & Layer\allowbreak Norm & Layer\allowbreak Norm & Layer\allowbreak Norm & Layer\allowbreak Norm
& Layer\allowbreak Norm & Layer\allowbreak Norm & Layer\allowbreak Norm & Layer\allowbreak Norm \\

Activation
& GELU & GELU & GELU & GELU & GELU
& GELU & GELU & GELU & GELU \\

Positional embedding
& Learned & Learned & Learned & Learned & Learned
& Learned & Learned & Learned & Learned \\

MAE decoder depth / width
& 2 / 64 & 2 / 64 & 2 / 64 & 2 / 64 & 2 / 64
& 2 / 64 & 2 / 64 & 2 / 64 & 2 / 64 \\

MAE decoder heads
& 4 & 4 & 4 & 4 & 4
& 4 & 4 & 4 & 4 \\

Masking ratio
& 0.75 & 0.75 & 0.75 & 0.75 & 0.75
& 0.9 & 0.9 & 0.9 & 0.9 \\

FNO layers / width
& 4 / 48 & 4 / 48 & 4 / 48 & 4 / 48 & 4 / 48
& 4 / 48 & 4 / 48 & 4 / 48 & 4 / 48 \\

FNO modes per axis
& 24 & 24 & 24 & 24 & 24
& 8 & 12 & 12 & 12 \\

FNO condition embedding
& 64 & 64 & 64 & 64 & 64
& 64 & 64 & 64 & 64 \\

Pretrain / forecast epochs
& 20 / 10
& 20 / 10
& 20 / 10
& 20 / 10
& 20 / 10
& 20 / 10
& 20 / 10
& 20 / 10
& 20 / 10 \\

Pretrain / forecast batch
& 96
& 128
& 128
& 128
& 128
& 12
& 12
& 12
& 12 \\

Pretraining LR
& 0.001 & 0.001 & 0.001 & 0.001 & 0.001
& 0.001 & 0.001 & 0.001 & 0.001 \\

Forecasting LR
& $8\!\times\!10^{-4}$
& $8\!\times\!10^{-4}$
& $8\!\times\!10^{-4}$
& $8\!\times\!10^{-4}$
& $8\!\times\!10^{-4}$
& $8\!\times\!10^{-4}$
& $8\!\times\!10^{-4}$
& $8\!\times\!10^{-4}$
& $8\!\times\!10^{-4}$ \\

Weight decay
& $10^{-4}$ & $10^{-4}$ & $10^{-4}$ & $10^{-4}$ & $10^{-4}$
& $10^{-4}$ & $10^{-4}$ & $10^{-4}$ & $10^{-4}$ \\

Optimizer / schedule
& AdamW / cos.
& AdamW / cos.
& AdamW / cos.
& AdamW / cos.
& AdamW / cos.
& AdamW / cos.
& AdamW / cos.
& AdamW / cos.
& AdamW / cos. \\

Warmup fraction
& 0.1 & 0.1 & 0.1 & 0.1 & 0.1
& 0.1 & 0.1 & 0.1 & 0.1 \\

Gradient clipping
& 1 & 1 & 1 & 1 & 1
& 1 & 1 & 1 & 1 \\

\end{xltabular}

\noindent
{\footnotesize
AdamW uses $(\beta_1,\beta_2)=(0.9,0.98)$ during pretraining and
$(\beta_1,\beta_2)=(0.9,0.999)$ during forecasting.
}

\endgroup

\lead{ENMA.} 
We encoded solution trajectories with a convolutional variational autoencoder using 32 latent channels. A temporal transformer provided context to a spatial transformer that predicted masked latent tokens; each stack contained six layers with a hidden size of 512, eight attention heads, and a feed-forward width of 2048. Each token covered one latent spatial site. Both stacks used RMSNorm, query/key normalization, SwiGLU, sinusoidal positional embeddings, and rotary attention. A conditional flow-matching head with three residual blocks and width 512 generated latent values, which were mapped back to physical fields by the VAE decoder. The configuration is given in Table~\ref{tab:baseline-enma}.

\begingroup

\scriptsize
\setlength{\tabcolsep}{2pt}
\renewcommand{\arraystretch}{1.08}

\setlength{\LTleft}{0pt}
\setlength{\LTright}{0pt}
\setlength{\LTpost}{0pt}

\begin{xltabular}{\textwidth}{
@{}
>{\raggedright\arraybackslash}p{0.20\textwidth}
*{9}{>{\centering\arraybackslash}X}
@{}
}

\caption{\textbf{ENMA implementation hyperparameters.}}
\label{tab:baseline-enma}\\

\toprule
Hyperparameter
& Combined
& Advection
& Burgers
& Heat
& Wave-B
& GS
& Wave2D
& Vorticity
& HeterNS \\
\midrule
\endfirsthead

\multicolumn{10}{c}{
\tablename\ \thetable{} -- continued from previous page
}\\
\toprule
Hyperparameter
& Combined
& Advection
& Burgers
& Heat
& Wave-B
& GS
& Wave2D
& Vorticity
& HeterNS \\
\midrule
\endhead

\midrule
\multicolumn{10}{r}{Continued on next page}\\
\endfoot

\bottomrule
\endlastfoot

Model input grid
& 128 & 128 & 128 & 128 & 128
& $32\times32$ & $64\times64$ & $64\times64$ & $64\times64$ \\

VAE spatial reduction
& 8 & 8 & 8 & 8 & 8
& $4\times4$ & $4\times4$ & $4\times4$ & $4\times4$ \\

Latent spatial grid
& $16$ & $16$ & $16$ & $16$ & $16$
& $64$ & $256$ & $256$ & $256$ \\

Latent channels
& 32 & 32 & 32 & 32 & 32
& 32 & 32 & 32 & 32 \\

Latent spatial patch size
& $1$ & $1$ & $1$ & $1$ & $1$
& $1\times1$ & $1\times1$ & $1\times1$ & $1\times1$ \\

Transformer depth
& 6 & 6 & 6 & 6 & 6
& 6 & 6 & 6 & 6 \\

Hidden size
& 512 & 512 & 512 & 512 & 512
& 512 & 512 & 512 & 512 \\

FFN hidden size
& 2048 & 2048 & 2048 & 2048 & 2048
& 2048 & 2048 & 2048 & 2048 \\

MLP ratio
& 4 & 4 & 4 & 4 & 4
& 4 & 4 & 4 & 4 \\

Attention heads
& 8 & 8 & 8 & 8 & 8
& 8 & 8 & 8 & 8 \\

Dropout
& 0 & 0 & 0 & 0 & 0
& 0 & 0 & 0 & 0 \\

Block normalization
& RMSNorm & RMSNorm & RMSNorm & RMSNorm & RMSNorm
& RMSNorm & RMSNorm & RMSNorm & RMSNorm \\

Activation
& SwiGLU & SwiGLU & SwiGLU & SwiGLU & SwiGLU
& SwiGLU & SwiGLU & SwiGLU & SwiGLU \\

Positional encoding
& Sin. + RoPE & Sin. + RoPE & Sin. + RoPE & Sin. + RoPE & Sin. + RoPE
& Sin. + RoPE & Sin. + RoPE & Sin. + RoPE & Sin. + RoPE \\

Flow head depth / width
& 3 / 512 & 3 / 512 & 3 / 512 & 3 / 512 & 3 / 512
& 3 / 512 & 3 / 512 & 3 / 512 & 3 / 512 \\

Sampling steps
& 10 & 10 & 10 & 10 & 10
& 10 & 10 & 10 & 10 \\

Latent training frames
& 14 & 14 & 14 & 14 & 14
& 10 & 10 & 10 & 10 \\

VAE / dynamics epochs
& 200 / 300
& 200 / 300
& 200 / 300
& 200 / 300
& 200 / 300
& 200 / 300
& 200 / 300
& 500 / 1000
& 200 / 300 \\

VAE / dynamics batch / GPU
& 32 / 32
& 32 / 32
& 32 / 32
& 32 / 32
& 32 / 32
& 32 / 64
& 32 / 64
& 32 / 64
& 32 / 64 \\

VAE / dynamics LR
& 0.001 & 0.001 & 0.001 & 0.001 & 0.001
& 0.001 & 0.001 & 0.001 & 0.001 \\

VAE weight decay
& 0.001 & 0.001 & 0.001 & 0.001 & 0.001
& 0.001 & 0.001 & 0.001 & 0.001 \\

Dynamics weight decay
& $10^{-4}$ & $10^{-4}$ & $10^{-4}$ & $10^{-4}$ & $10^{-4}$
& $10^{-4}$ & $10^{-4}$ & $10^{-4}$ & $10^{-4}$ \\

VAE KL weight
& $5\!\times\!10^{-4}$
& $5\!\times\!10^{-4}$
& $5\!\times\!10^{-4}$
& $5\!\times\!10^{-4}$
& $5\!\times\!10^{-4}$
& $5\!\times\!10^{-4}$
& $5\!\times\!10^{-4}$
& $5\!\times\!10^{-4}$
& $5\!\times\!10^{-4}$ \\

Optimizer / schedule
& AdamW / cos.
& AdamW / cos.
& AdamW / cos.
& AdamW / cos.
& AdamW / cos.
& AdamW / cos.
& AdamW / cos.
& AdamW / cos.
& AdamW / cos. \\

Gradient clipping
& 1 & 1 & 1 & 1 & 1
& 1 & 1 & 1 & 1 \\

\end{xltabular}

\noindent
{\footnotesize
AdamW uses $(\beta_1,\beta_2)=(0.9,0.999)$ for the VAE and
$(\beta_1,\beta_2)=(0.9,0.95)$ for the latent dynamics model.
}

\endgroup

\section{Additional experiments}
\paragraph{Terminology.}
For brevity, throughout the appendix we refer to the JEPA baseline as
\textbf{Vanilla}, the model with Physics-Aligned Latent Geometry (PAG) as
\textbf{Geo}, and the full model with both PAG and the Physics-Structured
Latent Predictor (PSP) as \textbf{Physics}. Thus, Vanilla$\rightarrow$Geo
isolates the effect of PAG, whereas Geo$\rightarrow$Physics isolates the
additional effect of PSP.

\subsection{Representation Analyses}
\label{app:repana}

We compare the representations learned by JEPA with LNS~\citep{li2025latent}, which learns an autoencoder-based latent state, and VideoMAE~\citep{tong2022videomae}, which learns representations through masked pixel reconstruction. We organize our analysis mainly on Vorticity around three questions. \textbf{Q1: What physical information is encoded?} We first examine whether the representations preserve governing parameters and local physical states through parameter and local-state probing in subsection~\ref{app:rep_pram} and~\ref{app:rep_ext}. \textbf{Q2: How is this information organized?} We then characterize the global and spatial organization of the latent space through variance concentration, cross-model similarity, and local spatial structure in subsection~\ref{app:rep_var}. \textbf{Q3: Is the representation easy to evolve?} Finally, we study temporal trajectory geometry and explicitly separate state reconstruction from autoregressive rollout accuracy in subsection~\ref{app:rep_tem} and~\ref{app:rep_direct_reconstruction_rollout}. Together, these analyses distinguish physical informativeness from latent organization and temporal evolvability. 

\subsubsection{Models and evaluation protocol}
Let $z=E(x)$ denote a frozen encoder representation of the model's native observation $x$. The primary VideoMAE baseline is the 10 M-parameter Tiny encoder pretrained for 100 epochs.

We analyze the learned representations from three complementary analyses on Vorticity. First, we evaluate parameter readout using attentive probes over the representations of the first 16 observed frames, using 10\%, 50\%, and 100\% of the labeled training data. Second, we evaluate local physical-state accessibility using resolution- and dimension-controlled linear probes across multiple PDEs. Third, we analyze the latent organization of 2,400 training, 120 external-ID, and 120 far-OOD trajectories across LNS, VideoMAE, and JEPA. For each 30-frame trajectory, we construct 15 representations corresponding to adjacent frame pairs, centered at times $0.5,2.5,\ldots,28.5$. For LNS and JEPA, each representation is obtained by averaging the features of the two adjacent frames. For VideoMAE, we use the corresponding tubelet features extracted from two overlapping 16-frame clips (frames 0--15 and 14--29). Thus, the VideoMAE context changes between the pair centers at 14.5 and 16.5. All representations are extracted from observed trajectories and are used only for representation analysis, rather than causal forecasting.

For the Vorticity latent-organization analysis, we process each model independently. We first standardize each latent channel using statistics computed from the training trajectories and then fit a model-specific 16-dimensional PCA basis (PCA16~\citep{pearson1901liii}) on the standardized training representations. The retained PCA coordinates are further normalized by a single global scale determined by their total retained variance, without whitening individual principal components. We use this PCA16 space to characterize variance concentration and temporal geometry, and additionally perform an auxiliary linear viscosity readout. The linear probe is fitted on 1,800 training trajectories and evaluated on 600 internal holdout trajectories, while the external-ID and far-OOD trajectories are kept as separate test sets. Because PCA is fitted independently for each model, neither the PCA axes nor the original latent dimensions are assumed to be shared across JEPA, LNS, and VideoMAE.

\subsubsection{Parameter Information Analyses}
\label{app:rep_pram}

We mainly discuss in Vorticity. The 10\%, 50\%, and 100\% label budgets are nested subsets of complete viscosity environments, containing 1,200, 6,000, and 12,000 training trajectories. The regression target is train-standardized log viscosity. Probe optimization runs for 100 epochs with a fixed seed, and checkpoint selection uses ID validation data before evaluation on the ID test set.

JEPA achieves normalized test MSEs of 0.042597, 0.012902, and 0.009310 across the three supervision budgets (Table~\ref{tab:ns_primary}). Its advantage is modest at 10\%, but widens with more labeled data: at 100\%, JEPA reduces error by 55.3\% relative to LNS and 50.9\% relative to VideoMAE. Compared with VideoMAE, the reduction increases from 7.9\% to 47.5\% to 50.9\% as supervision grows, indicating that the strongest separation emerges in the higher-label regime rather than under limited supervision.


\begin{table}[h]
\centering
\small
\caption{\textbf{Primary comparison.} Vorticity viscosity readout from frozen JEPA, LNS, and VideoMAE.}
\label{tab:ns_primary}
\begin{adjustbox}{max width=\textwidth}
\begin{tabular}{lrrrrr}
\toprule
Representation & MSE, 10\% $\downarrow$ & MSE, 50\% $\downarrow$ & MSE, 100\% $\downarrow$ & $R^2$, 100\% $\uparrow$& Rel. MAE (\%) $\downarrow$ \\
\midrule
JEPA, raw $z$ & 0.042597 & 0.012902 & 0.009310 & 0.9897 & 4.12 \\
LNS AE & 0.043677 & 0.028622 & 0.020830 & 0.9769 & 6.00 \\
VideoMAE-Tiny & 0.046264 & 0.024560 & 0.018950 & 0.9790 & 5.88 \\
\bottomrule
\end{tabular}
\end{adjustbox}
\par\smallskip
\begin{minipage}{\textwidth}\footnotesize Native-token attentive probes use the first 16 observed frames. MSE uses train-standardized log viscosity. VideoMAE denotes the 10M-parameter Tiny encoder pretrained for 100 epochs.
\end{minipage}
\end{table}

VideoMAE nevertheless provides a strong physical representation: its full-budget $R^2$ is 0.9790 and mean relative viscosity error is 5.88\%, compared with 0.9769 and 6.00\% for LNS and 0.9897 and 4.12\% for JEPA. VideoMAE outperforms LNS at the two larger budgets, while LNS has slightly lower error at 10\%.

\begin{table}[htbp]
\centering
\small
\caption{\textbf{Viscosity information in the aligned pooled representations.}}
\label{tab:pooled_primary}
\begin{adjustbox}{max width=\textwidth}
\begin{tabular}{lrrrrr}
\toprule
Representation & Holdout $R^2$ $\uparrow$& External ID $R^2$ $\uparrow$& kNN-5 $R^2$ $\uparrow$& Far-OOD $R^2$ $\uparrow$& Far-OOD RMSE $\downarrow$\\
\midrule
JEPA, raw $z$  & 0.9555 & 0.9659 & 0.7820 & -11.2557 & 4.0933 \\
LNS & 0.9484 & 0.1422 & 0.6780 & -23.3780 & 5.7730 \\
VideoMAE-Tiny & 0.9182 & 0.9194 & 0.7586 & -15.9398 & 4.8123 \\
\bottomrule
\end{tabular}
\end{adjustbox}
\par\smallskip
\begin{minipage}{\textwidth}\footnotesize Linear probes use PCA16 features at the final two-frame center, 28.5, with 1,800 fitting trajectories, 600 internal holdout trajectories, and separate 120-trajectory ID and far-OOD sets. The kNN-5 column uses all 15 centers in the internal holdout. Far-OOD RMSE is standardized by the training target scale.
\end{minipage}
\end{table}

\begin{figure}[htbp]
\centering
\includegraphics[width=\textwidth,height=0.72\textheight,keepaspectratio]{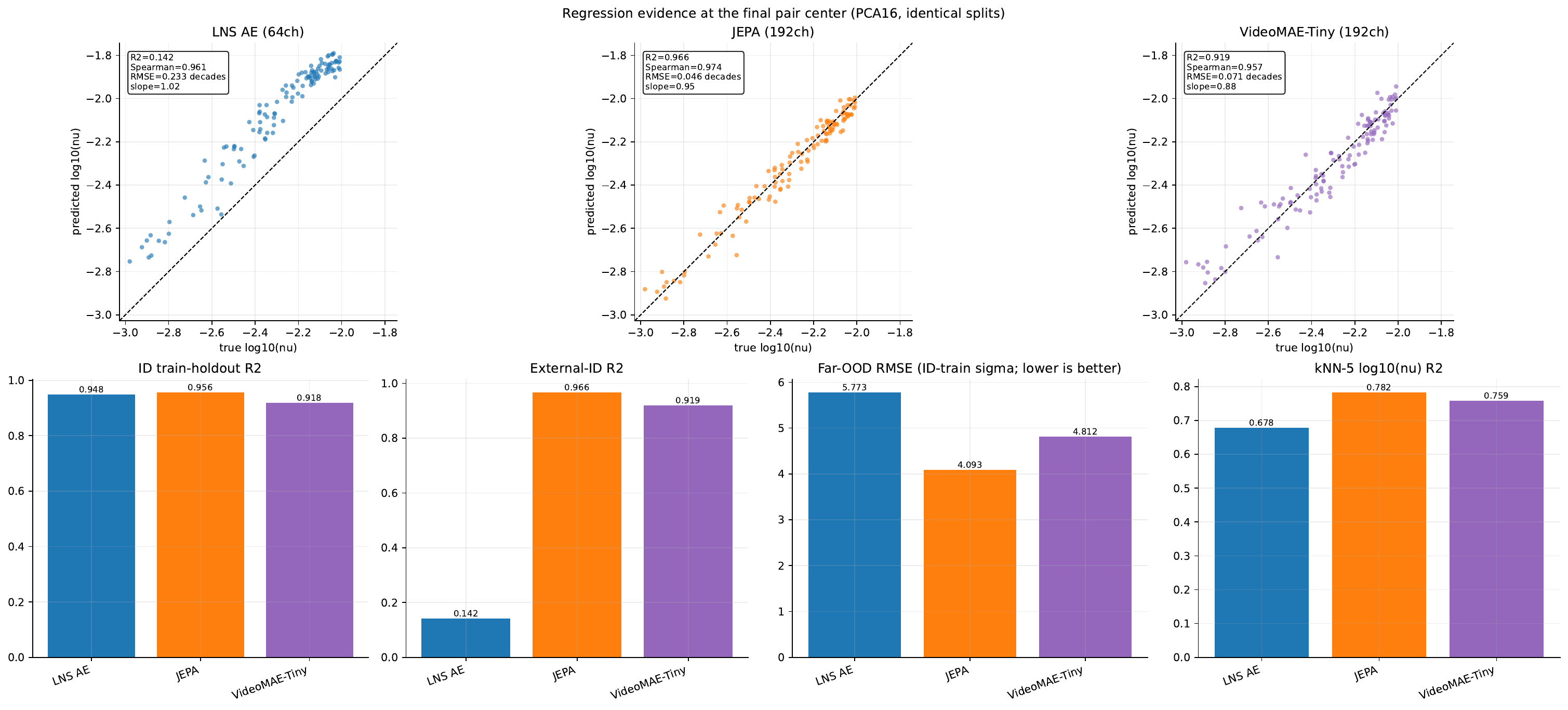}
\caption{\textbf{Viscosity readout across evaluation regimes.} Upper panels compare external-ID viscosity predictions with numerical truth at the final two-frame center. Lower panels report internal holdout readout, external-ID readout, far-OOD standardized RMSE, and five-neighbor readout. LNS retains strong rank ordering despite a calibration shift on external ID.}
\label{fig:pooled_regression}
\end{figure}

The standardized PCA-based comparison reaches a consistent conclusion using a much smaller linear readout (Table~\ref{tab:pooled_primary}; Figure~\ref{fig:pooled_regression}). At the final two-frame center, JEPA has internal-holdout/external-ID $R^2$ of 0.9555/0.9659, compared with 0.9484/0.1422 for LNS and 0.9182/0.9194 for VideoMAE. LNS's external-ID decline coexists with a Spearman correlation of 0.9605, and the prediction scatter shows a systematic offset despite preserved ordering. However, strong ID readout does not directly translate to parameter extrapolation. On far-OOD viscosities, all three linear probes yield negative $R^2$, with standardized RMSEs of 4.0933, 5.7730, and 4.8123 for JEPA, LNS, and VideoMAE, respectively. JEPA therefore retains the smallest extrapolation error, but it does not provide reliable out-of-range prediction.


A five-nearest-neighbor probe (kNN-5) over all aligned pair centers further separates physical information from linear probing. Viscosity $R^2$ is 0.7820 for JEPA, 0.6780 for LNS, and 0.7586 for VideoMAE. VideoMAE therefore retains substantial local parameter organization even though its dominant PCA structure differs visibly from the other two models.

\subsubsection{Variance concentration and latent organization}
\label{app:rep_var}
\begin{figure}[h]
\centering
\includegraphics[width=\textwidth,height=0.72\textheight,keepaspectratio]{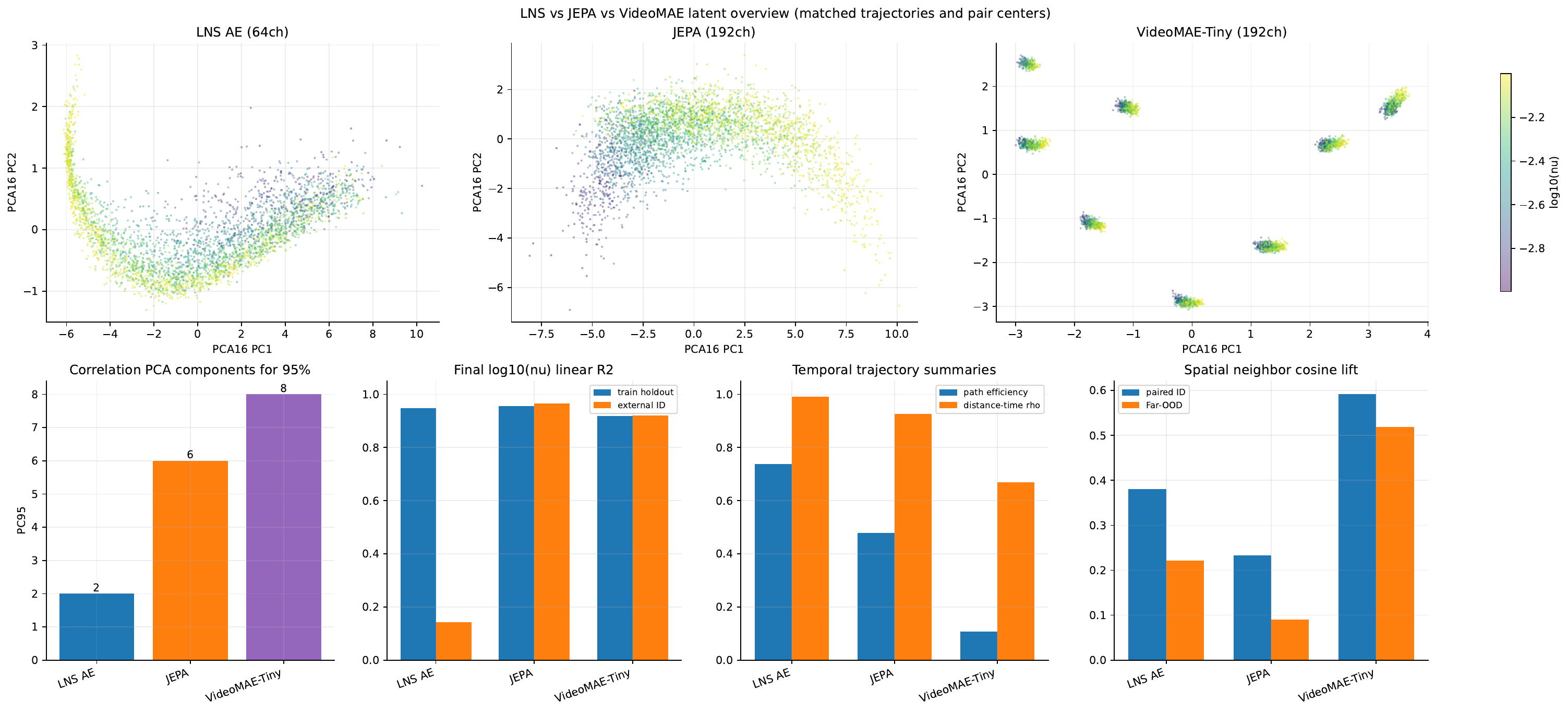}
\caption{\textbf{Primary baseline families on matched Vorticity trajectories and two-frame centers.} The upper panels show model-specific pooled PCA coordinates colored by viscosity. Lower panels summarize variance concentration, held-out linear readout, training-trajectory geometry, and native-grid spatial similarity.}
\label{fig:video_overview}
\end{figure}

The three representations exhibit markedly different variance concentrations (Figure~\ref{fig:video_overview}; Table~\ref{tab:primary_structure}). After standardizing the training channels, LNS is highly concentrated in a few dominant directions: its first two principal components explain 97.11\% of the variance, and only two components are required to reach 95\% explained variance. JEPA is less concentrated, with 86.47\% captured by the first two components and six components required for 95\% variance. VideoMAE exhibits the broadest spectrum, with only 45.02\% captured by the first two components and eight components required to reach 95\%. The corresponding entropy effective ranks are 1.31, 2.77, and 8.92 for LNS, JEPA, and VideoMAE, respectively. These results show that LNS compresses variation into a small number of dominant directions, whereas JEPA and particularly VideoMAE distribute variation across a broader latent subspace.

\begin{table}[htbp]
\centering
\small
\caption{\textbf{Variance concentration and temporal organization of pooled features.}}
\label{tab:primary_structure}
\begin{adjustbox}{max width=\textwidth}
\begin{tabular}{lrrrrr}
\toprule
Representation & PC95 & Eff. rank & Top-2 var. (\%) & ID path efficiency $\uparrow$& ID distance--time $\rho$ $\uparrow$\\
\midrule
JEPA & 6 & 2.77 & 86.47 & 0.4901 & 0.9451 \\
LNS & 2 & 1.31 & 97.11 & 0.7581 & 0.9925 \\
VideoMAE-Tiny & 8 & 8.92 & 45.02 & 0.1070 & 0.6695 \\
\bottomrule
\end{tabular}
\end{adjustbox}
\par\smallskip
\begin{minipage}{\textwidth}\footnotesize Spectral statistics use training-channel-standardized pooled features. Temporal statistics use external ID trajectories in each model's PCA16 space. Higher path efficiency means a straighter path; neither it nor effective rank is a physical-accuracy score. These are pooled-channel spectra, not token PCA spectra.
\end{minipage}
\end{table}

Neither a compact spectrum nor a higher effective rank directly ranks physical usefulness. LNS retains strong viscosity ordering in a concentrated representation, JEPA combines a moderately distributed spectrum with stronger readout, and VideoMAE's larger effective rank partly reflects temporal-position structure. The pairwise linear centered kernel alignment (CKA)~\citep{kornblith2019similarity} matrix further characterizes the differences between their representations (Table~\ref{tab:primary_cka}). On 10,000 aligned pooled samples, linear CKA is 0.8018 between JEPA and LNS, compared with 0.1361 between JEPA and VideoMAE and 0.1964 between LNS and VideoMAE. Despite this low cross-model similarity, VideoMAE achieves an external-ID viscosity $R^2$ of 0.9194 (Table~\ref{tab:pooled_primary}). Thus, different representation structures can support useful viscosity inference.

\begin{table}[h]
\centering
\small
\caption{Pairwise linear CKA on 10,000 aligned pooled samples.}
\label{tab:primary_cka}
\begin{adjustbox}{max width=\textwidth}
\begin{tabular}{lrrr}
\toprule
Model & JEPA & LNS & VideoMAE \\
\midrule
JEPA & 1.0000 & 0.8018 & 0.1361 \\
LNS & 0.8018 & 1.0000 & 0.1964 \\
VideoMAE & 0.1361 & 0.1964 & 1.0000 \\
\bottomrule
\end{tabular}
\end{adjustbox}
\par\smallskip
\end{table}

\subsubsection{Temporal and spatial organization}
\label{app:rep_tem}
\begin{figure}[h]
\centering
\includegraphics[width=\textwidth,height=0.3\textheight,keepaspectratio]{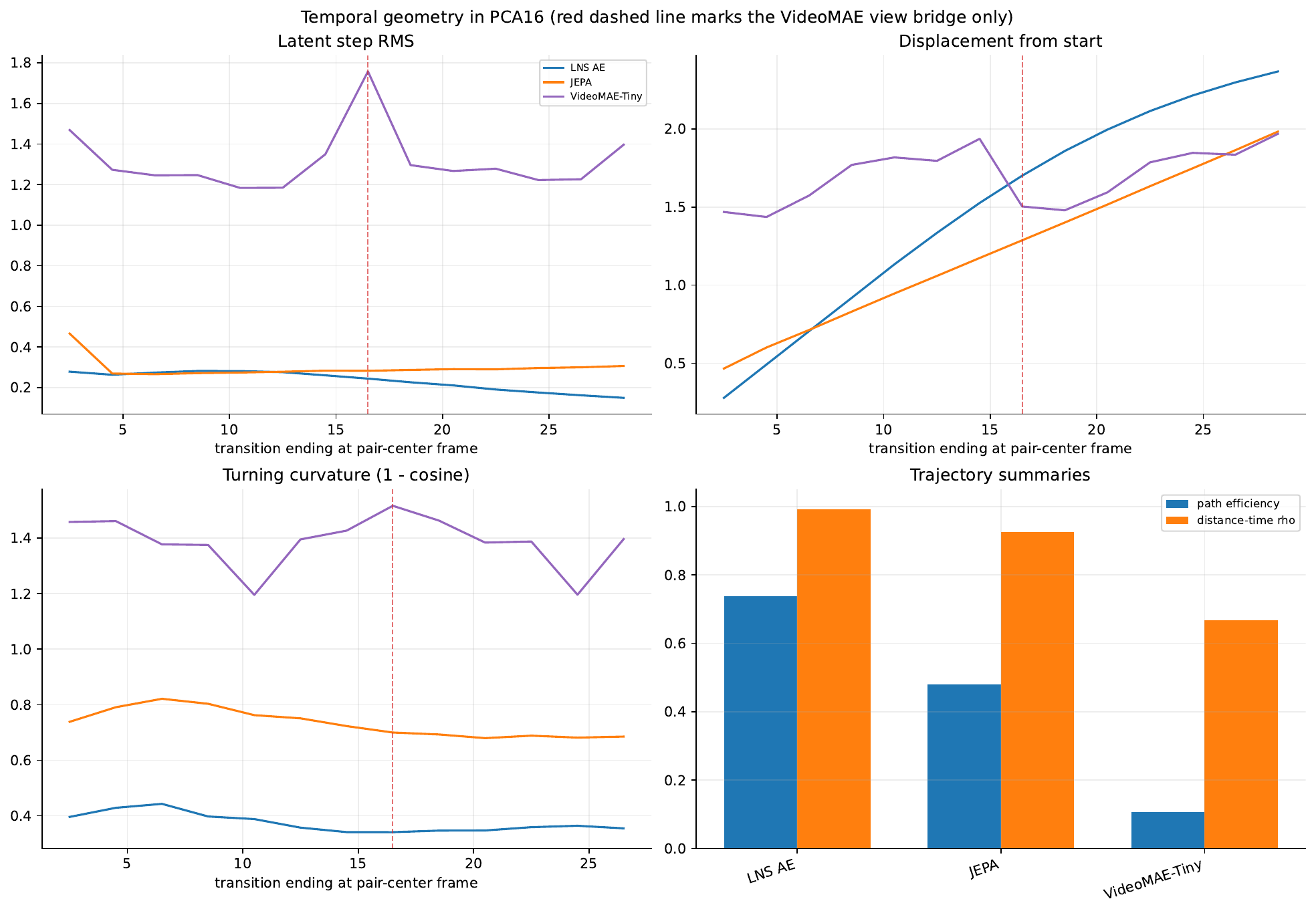}
\caption{\textbf{Temporal organization in model-specific pooled PCA16 coordinates.} \textit{UpperLeft:} Latent step magnitude measures the size of consecutive temporal transitions; LNS contracts, JEPA stays stable, and VideoMAE fluctuates sharply. \textit{UpperRight:} Distance from the initial state measures global temporal progression; LNS separates most steadily; JEPA remains stable; VideoMAE is disrupted. \textit{LowerLeft:} Temporal turning, quantified by one minus the cosine similarity between consecutive latent increments, measures changes in trajectory direction; LNS turns least, JEPA moderately, and VideoMAE most. \textit{LowerRight:} Path efficiency and distance--time correlation summary; LNS is most smooth; JEPA remains ordered; VideoMAE is most context dependent. }
\label{fig:pooled_temporal}
\end{figure}

We next examine whether physically informative representations also induce latent trajectories that are easy to evolve. Path efficiency, defined as the endpoint displacement divided by the accumulated step length, measures how directly a trajectory progresses through latent space. On external ID data, LNS achieves a path efficiency of 0.7581, compared with 0.4901 for JEPA and 0.1070 for VideoMAE (Table~\ref{tab:primary_structure}). The corresponding correlations between elapsed time and distance from the initial state are 0.9925, 0.9451, and 0.6695, respectively. Thus, LNS follows a straighter and more monotonically separating trajectory than JEPA, despite JEPA exhibiting stronger physical readout in the preceding analyses. This distinction suggests that physical structure and evolvability are complementary properties: a representation may encode governing factors well without automatically yielding the simplest geometry for rollout. The same trend is observed on the training trajectories (Figure~\ref{fig:pooled_temporal}), motivating the explicit geometry adaptation introduced later.

\begin{figure}[h]
\centering
\includegraphics[width=\textwidth,height=0.42\textheight,keepaspectratio]{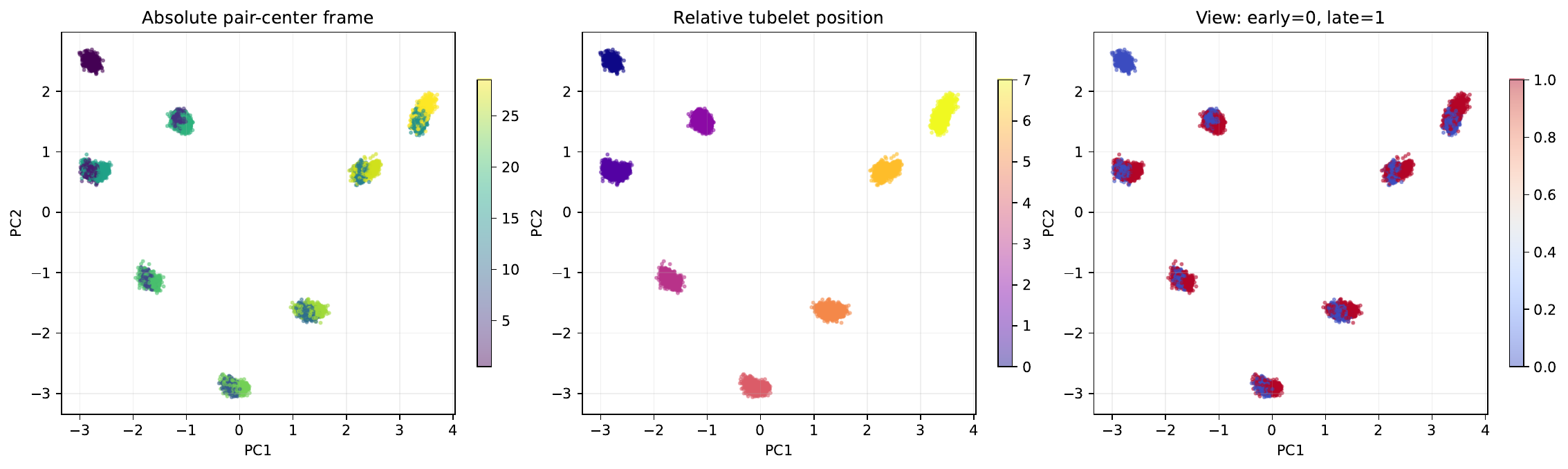}
\caption{\textbf{VideoMAE pooled features colored by absolute time, relative tubelet position, and clip identity.} \textit{Left:} Absolute-time coloring tests whether the clusters reflect physical time. \textit{Middle:} Relative-position coloring shows that the clusters are primarily organized by tubelet position within the input clip. \textit{Right:} Early/late view coloring shows substantial overlap within clusters, indicating that clip identity alone does not explain the separation.}
\label{fig:videomae_position}
\end{figure}

VideoMAE requires an additional caveat because its pooled geometry is strongly influenced by clip-relative position. Relative tubelet position is nearly perfectly linearly readable ($R^2=0.9995$) and explains 94.5\% of the pooled PCA16 variance (Figure~\ref{fig:videomae_position}). Although its stitched trajectory is sensitive to the clip-context boundary, temporal progression remains well organized within each individual view ($\rho\approx0.95$). We therefore treat its pooled temporal geometry as context dependent rather than as a context-invariant instantaneous state.

\begin{figure}[htbp]
\centering
\includegraphics[width=\textwidth,height=0.72\textheight,keepaspectratio]{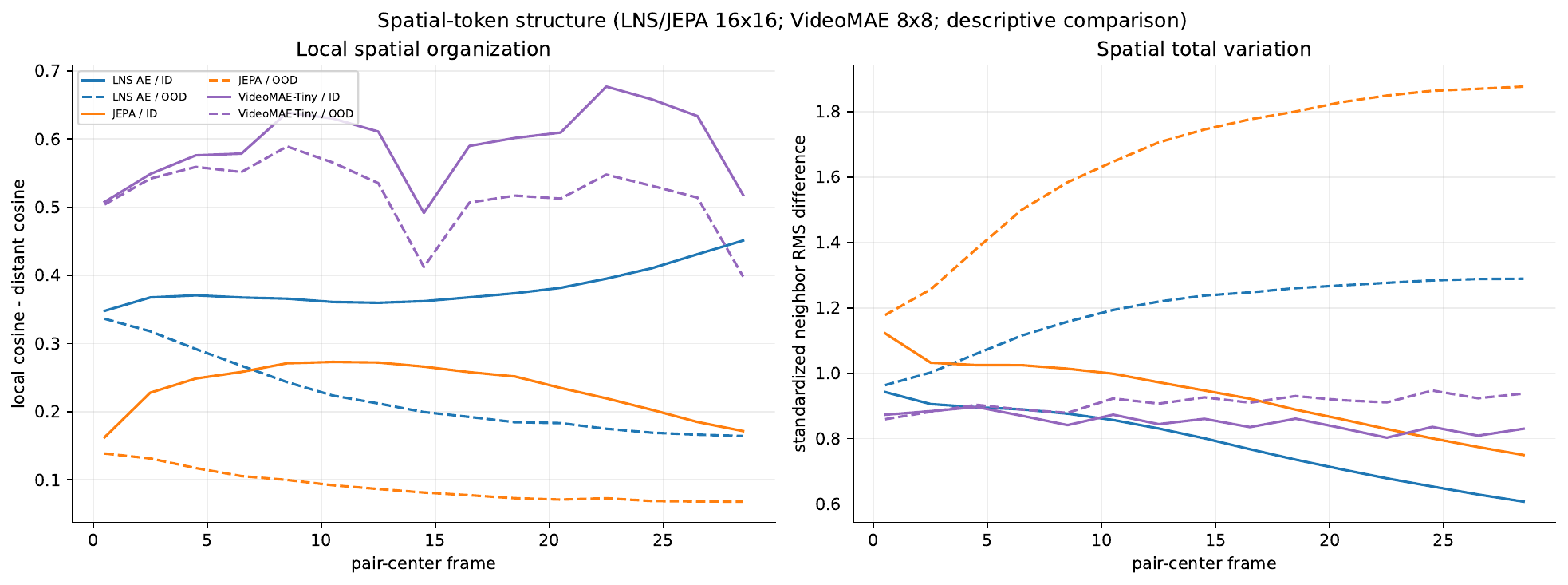}
\caption{\textbf{Native-token spatial organization on paired ID and far-OOD trajectories.} \textit{Left:} Local spatial organization. Local lift measures spatial locality; LNS exceeds JEPA, while VideoMAE is highest descriptively.
\textit{Right:} Spatial total variation. Spatial variation grows most under JEPA OOD, while ID becomes smoother.}
\label{fig:pooled_spatial}
\end{figure}

All three representations also preserve local spatial organization (Table~\ref{tab:primary_spatial}; Figure~\ref{fig:pooled_spatial}). On their shared $16\times16$ token grids, LNS and JEPA obtain ID neighbor--distant cosine lifts of 0.3809 and 0.2335, respectively, with corresponding far-OOD values of 0.2218 and 0.0902. VideoMAE also exhibits persistent local structure, but its $8\times8$ token grid has a different physical spacing. Its spatial statistics are therefore included as a descriptive reference rather than a resolution-controlled comparison. Overall, these results further separate physical informativeness from geometric simplicity: JEPA provides a physically informative representation, while additional geometric adaptation is needed to make its latent evolution more favorable for rollout.

\begin{table}[h]
\centering
\small
\caption{\textbf{Spatial organization of the primary representation families.}}
\label{tab:primary_spatial}
\begin{adjustbox}{max width=\textwidth}
\begin{tabular}{lrrrrr}
\toprule
Representation & Native grid & ID neighbor cos. $\uparrow$ & ID distant cos. $\uparrow$& ID lift $\uparrow$& Far-OOD lift $\uparrow$\\
\midrule
JEPA & $16\times16$ & 0.4792 & 0.2457 & 0.2335 & 0.0902 \\
LNS & $16\times16$ & 0.4501 & 0.0692 & 0.3809 & 0.2218 \\
VideoMAE-Tiny & $8\times8$ & 0.6210 & 0.0296 & 0.5914 & 0.5192 \\
\bottomrule
\end{tabular}
\end{adjustbox}
\par\smallskip
\begin{minipage}{\textwidth}\footnotesize
Lift denotes neighboring-token minus distant-token cosine similarity, averaged across the 15 two-frame centers. A larger lift indicates stronger local spatial organization.
\end{minipage}
\end{table}

\subsubsection{Extension to other PDEs}
\label{app:rep_ext}
\begin{figure}[h]
\centering
\includegraphics[width=\textwidth,height=0.3\textheight,keepaspectratio]{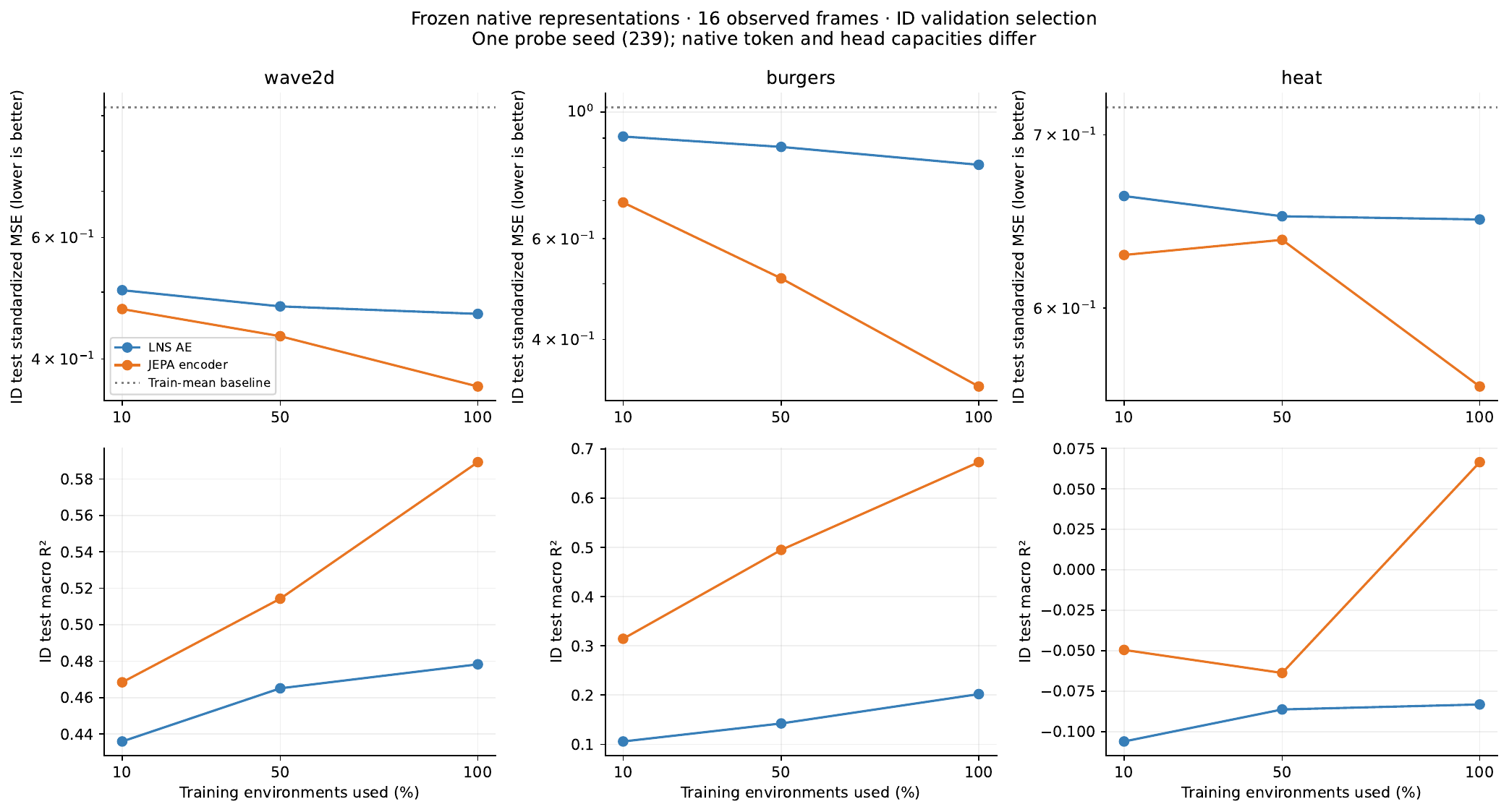}
\caption{\textbf{Parameter readout versus the number of labeled training environments.} \textit{Top row:} gives test standardized MSE; \textit{Bottom row:} gives R-squared.}
\label{fig:label_efficiency}
\end{figure}

The attentive comparison further extends JEPA and LNS to Wave2D, Burgers, and Heat (Table~\ref{tab:cross_probe}; Figure~\ref{fig:label_efficiency}), where JEPA achieves lower test MSE across all label budgets. At full supervision, the error reductions are 21.5\%, 59.1\%, and 13.8\%, respectively. The largest gains appear in Wave2D wave-speed readout ($R^2$: 0.4783$\rightarrow$0.5893) and Burgers log-diffusivity (0.2020$\rightarrow$0.6734), while Heat remains challenging for both models.

\begin{table}[htbp]
\centering\small

\caption{\textbf{Parameter readout across label budgets and PDEs.}}
\label{tab:cross_probe}
\begin{adjustbox}{max width=\textwidth}
\begin{tabular}{llrrrrr}
\toprule
Dataset & Budget & LNS MSE $\downarrow$ & JEPA MSE $\downarrow$& Decrease (\%) $\uparrow$& LNS $R^2$ $\uparrow$& JEPA $R^2$ $\uparrow$\\
\midrule

Wave2D & 10\% & 0.503139 & 0.472475 & 6.1 & 0.4359 & 0.4684 \\
Wave2D & 50\% & 0.476702 & 0.431590 & 9.5 & 0.4651 & 0.5143 \\
Wave2D & 100\% & 0.464928 & 0.364986 & 21.5 & 0.4783 & 0.5893 \\

Burgers & 10\% & 0.905448 & 0.694140 & 23.3 & 0.1057 & 0.3144 \\
Burgers & 50\% & 0.868372 & 0.511263 & 41.1 & 0.1423 & 0.4950 \\
Burgers & 100\% & 0.807950 & 0.330690 & 59.1 & 0.2020 & 0.6734 \\

Heat & 10\% & 0.662985 & 0.629060 & 5.1 & -0.1061 & -0.0495 \\
Heat & 50\% & 0.651126 & 0.637604 & 2.1 & -0.0863 & -0.0638 \\
Heat & 100\% & 0.649273 & 0.559498 & 13.8 & -0.0832 & 0.0665 \\
\bottomrule
\end{tabular}
\end{adjustbox}
\end{table}

To assess whether the full-budget improvements are consistent across specific test trajectories, we bootstrap the JEPA--LNS MSE difference, where negative values favor JEPA. The 95\% intervals remain below zero on Wave2D (\([-0.1294,-0.0718]\)) and Burgers (\([-1.0273,-0.0847]\)), indicating consistent improvements across resampled environments. For Heat, the interval (\([-0.1867,0.0102]\)) crosses zero, so the improvement is less conclusive.

For local-state analysis, we use MAE-PDE~\citep{zhou2024masked} as a alternative to VideoMAE. Using a shared $8\times8$ grid and 16 PCA channels, JEPA achieves the strongest ID/OOD linear readout on CFD (0.9921/0.9492) (CFD refers to Vorticity in this subsection) and Wave2D (0.8773/0.8633), compared with 0.9578/0.8818 and 0.8258/0.7254 for LNS (Table~\ref{tab:local_2d}). MAE-PDE shows weaker controlled local-state readout on CFD and Wave2D ($R^2\approx0.10$ and $0.37$). Overall, local physical-state information and governing-parameter information remain distinct representation properties.

\begin{table}[h]
\centering
\small
\caption{\textbf{Dimension-controlled local physical-state readout.}}
\label{tab:local_2d}
\begin{adjustbox}{max width=\textwidth}
\begin{tabular}{lrrrr}
\toprule
Representation & CFD ID $R^2$ $\uparrow$& CFD OOD $R^2$ $\uparrow$& Wave2D ID $R^2$ $\uparrow$& Wave2D OOD $R^2$ $\uparrow$\\
\midrule
LNS & 0.9578 & 0.8818 & 0.8258 & 0.7254 \\
JEPA &0.9921 & 0.9492 & 0.8773 & 0.8633 \\
MAE-PDE, single & 0.0962 & 0.0698 & 0.3658 & 0.3063 \\
MAE-PDE, history-5 & 0.1034 & 0.0758 & 0.3696 & 0.2933 \\
\bottomrule
\end{tabular}
\end{adjustbox}
\par\smallskip
\end{table}

\begin{figure}[htbp]
\centering
\includegraphics[width=\textwidth,height=0.72\textheight,keepaspectratio]{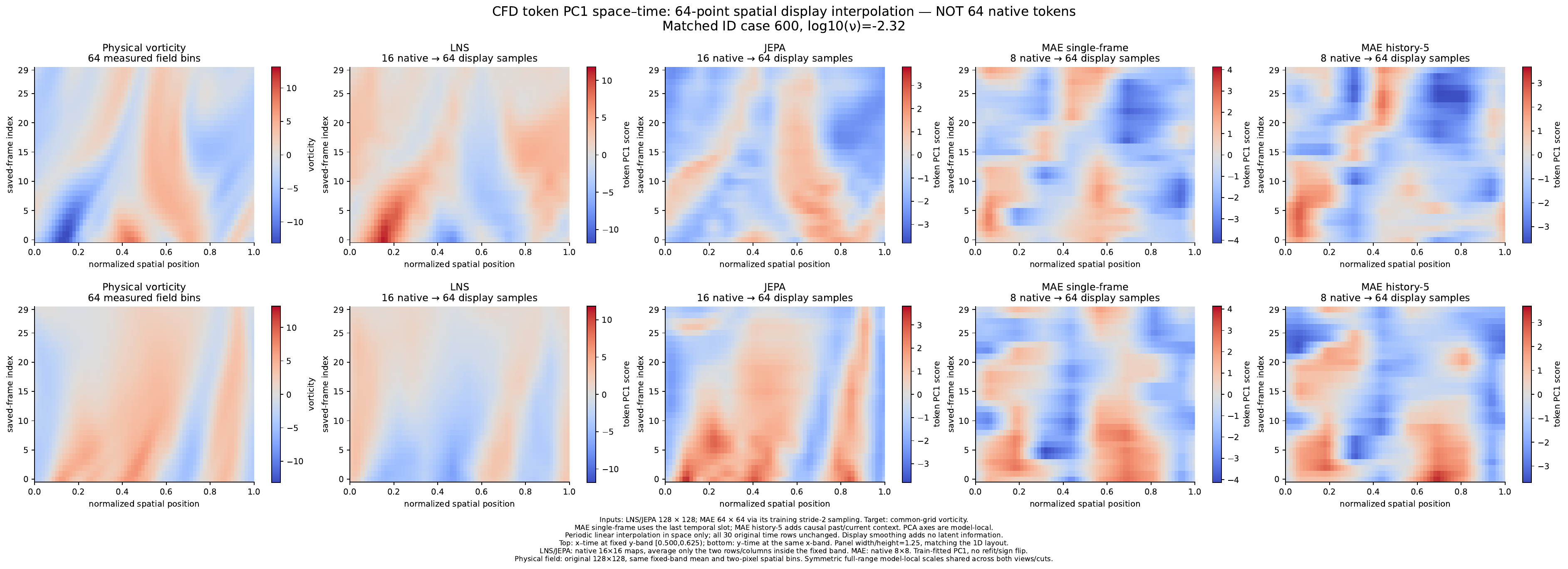}
\caption{\textbf{CFD physical-field and token-PC1 space--time maps.}Matched CFD space--time maps along a fixed spatial line comparing physical vorticity with each model's token-PC1 representation.}
\label{fig:spacetime_cfd}
\end{figure}
\begin{figure}[htbp]
\centering
\includegraphics[width=\textwidth,height=0.72\textheight,keepaspectratio]{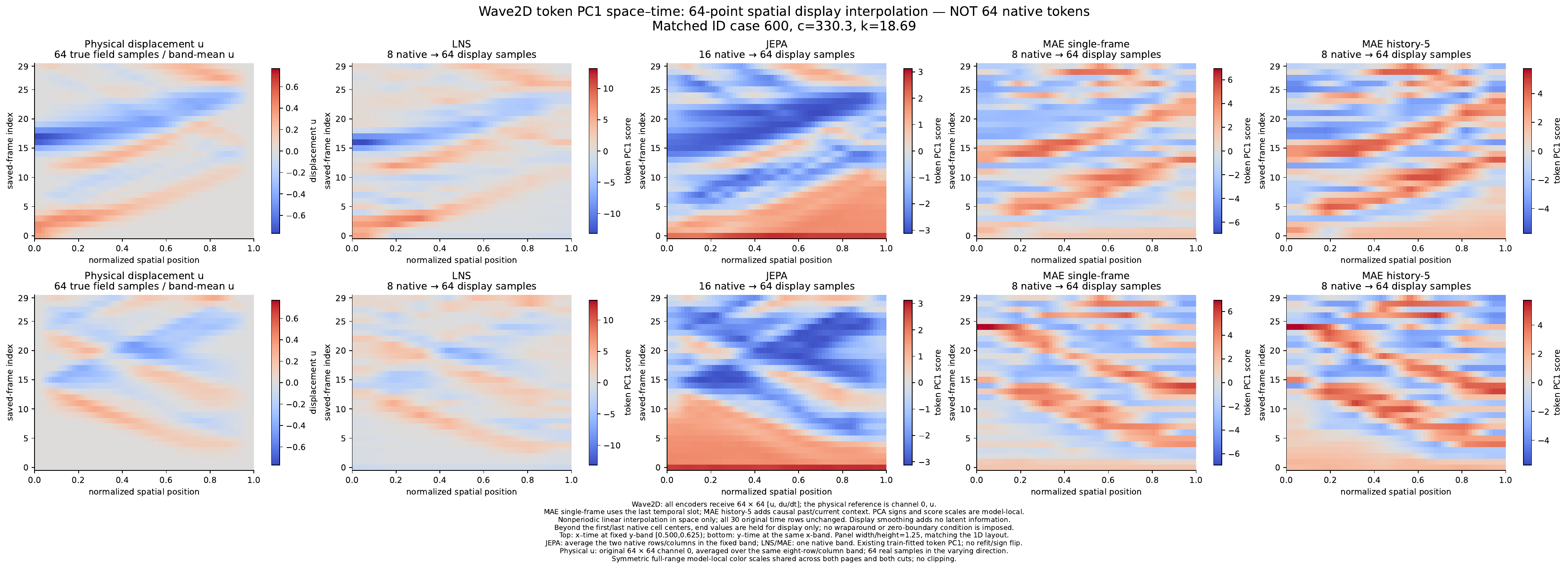}
\caption{\textbf{Matched Wave2D displacement and token-PC1 space--time maps.}}
\label{fig:wave_spacetime}
\end{figure}
\begin{figure}[htbp]
\centering
\includegraphics[width=\textwidth,height=0.72\textheight,keepaspectratio]{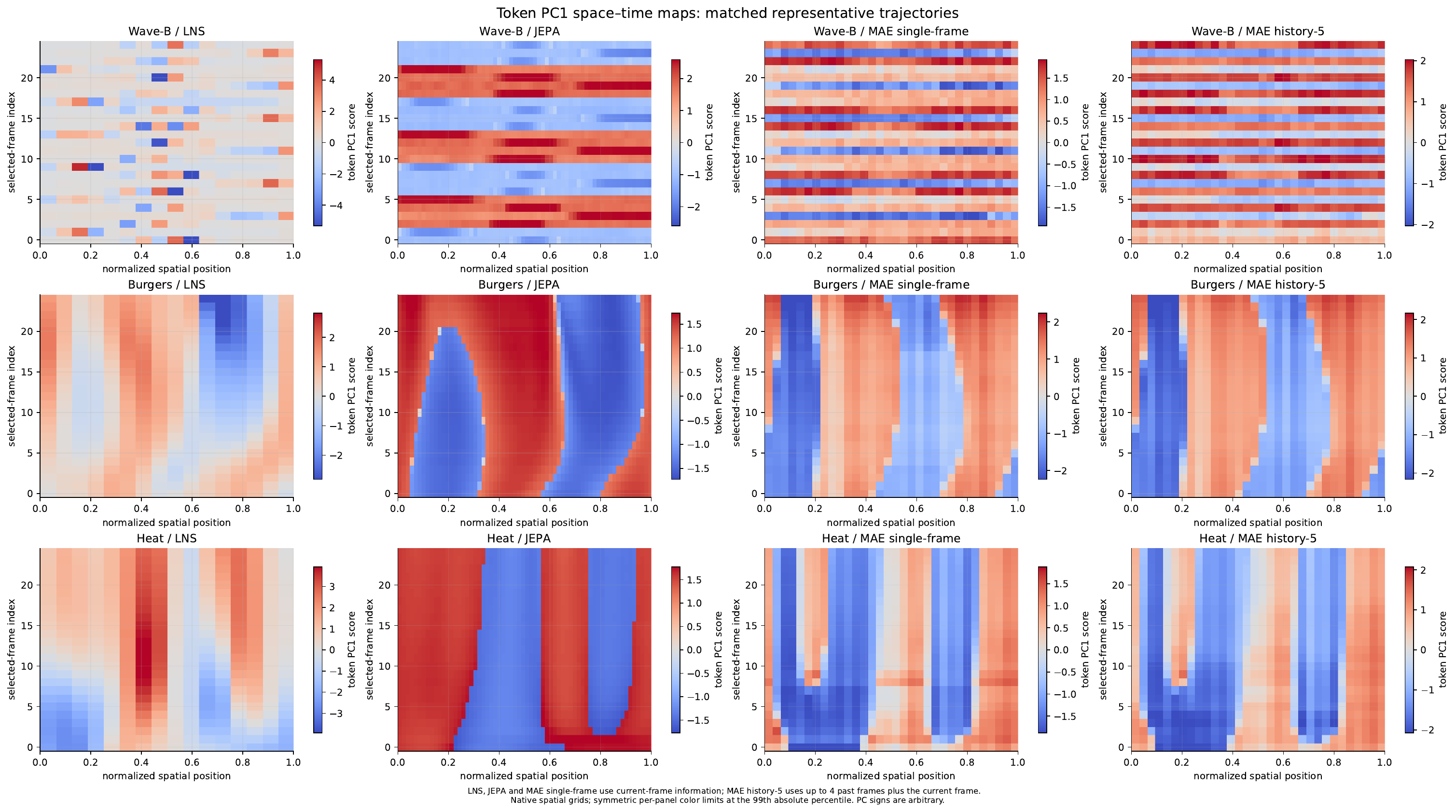}
\caption{\textbf{Matched Wave-B, Burgers, and Heat token-PC1 space--time maps.}}
\label{fig:spacetime_1d}
\end{figure}

Matched CFD and Wave2D space--time cuts compare the physical field with each model's train-fitted PC1 along the same trajectories (Figures~\ref{fig:spacetime_cfd} and~\ref{fig:wave_spacetime}). On the 1D PDEs, the corresponding PC1 maps visualize their space--time organization (Figure~\ref{fig:spacetime_1d}). The space--time maps reveal qualitatively different organizations across the
representations. On Wave2D and CFD, JEPA's leading component forms coherent
spatiotemporal structures that track the propagation and deformation of the
underlying physical fields, whereas LNS appears smoother and MAE-PDE exhibits
more blockwise or stripe-like organization. Similar equation-specific patterns
emerge on the 1D PDEs, particularly for Burgers and Heat. Overall, JEPA exhibits more coherent and physically aligned spatiotemporal organization across PDEs.

\subsubsection{Direct Reconstruction versus Autoregressive Rollout} 
\label{app:rep_direct_reconstruction_rollout}

Accurate reconstruction measures how well a representation preserves a
physical state; forecasting additionally requires predicting its evolution.
We examine these properties separately. Direct
reconstruction decodes the ground-truth representation at each time,
$\widetilde u_t=D_{\mathrm{rec}}(z_t)$.
Autoregressive rollout instead decodes predicted latent states,
$\widehat u_t=D_{\mathrm{roll}}(\widehat z_t)$, generated from the observed
prefix without access to future states. For the evaluated frame set
$\mathcal T$, we average the physical relative $L^2$ error over trajectories
and report
\begin{equation}
 e_a=\frac{1}{N}\sum_{i=1}^{N}
 \frac{\|u^{a}_{i,\mathcal T}-u_{i,\mathcal T}\|_2}
      {\|u_{i,\mathcal T}\|_2},
 \qquad R=\frac{e_{\mathrm{roll}}}{e_{\mathrm{rec}}},
 \label{eq:direct_rollout_metric}
\end{equation}
where $a\in\{\mathrm{rec},\mathrm{roll}\}$ and the norm covers time, space,
and physical channels. Thus, $R$ is a ratio of mean errors.

\begin{table}[htbp]
\centering
\small
\caption{\textbf{Direct reconstruction and autoregressive rollout errors on the ID
validation sets for JEPA and Zebra.} We report
$R=e_{\mathrm{roll}}/e_{\mathrm{rec}}$, which measures the increase from
reconstruction to rollout error; larger values indicate a larger
reconstruction--prediction gap. Parentheses denote the decoder input.
Zebra results are unavailable for Gray--Scott.}
\label{tab:direct_rollout_main}
\begin{tabular}{lrrr|rrr}
\toprule
Dataset (input) 
& $e_{\mathrm{rec}}$ 
& $e_{\mathrm{roll}}$ 
& JEPA-$R$ 
& $e_{\mathrm{rec}}$ 
& $e_{\mathrm{roll}}$ 
& Zebra-$R$ \\ 
\midrule
Advection ($z$) & 0.0029 & 0.0235 & 8.12 & 0.0003 & 0.0079 & 26.33 \\
Burgers ($z$)   & 0.0151 & 0.135 & 8.99 & 0.0016 & 0.1540 & 96.25 \\
Heat ($z$)      & 0.0085 & 0.0808 & 9.51 & 0.0019 & 0.1150 & 60.53 \\
Wave-B ($z$)    & 0.0125 & 0.1115 & 8.92 & 0.0011 & 0.2450 & 222.73 \\
Wave-2D ($z$)   & 0.0415 & 0.363 & 8.74 & 0.0010 & 0.2070 & 207.00 \\
Combined ($z$)  & 0.0027 & 0.0332 & 12.31& 0.0022 & 0.0079 & 3.59 \\
Vorticity ($z$) & 0.0104 & 0.086 & 8.26 & 0.0170 & 0.1190 & 7.00 \\
GS ($z$)        & 0.0079 & 0.0807 & 10.22 & - & - & - \\
\bottomrule
\end{tabular}
\end{table}

Table~\ref{tab:direct_rollout_main} shows a clear separation between
reconstruction fidelity and temporal predictability. For JEPA, the
rollout-to-reconstruction ratio remains relatively consistent across datasets,
ranging from $8.12$ to $12.31$. In contrast, Zebra exhibits much larger and
more variable gaps on several benchmarks, including Burgers ($96.25\times$),
Wave-B ($222.73\times$), and Wave2D ($207.00\times$), despite its substantially
smaller reconstruction errors. The ratio alone, however, does not rank rollout
accuracy: for example, Zebra has a much larger ratio on Wave2D but a lower
absolute rollout error than JEPA, whereas JEPA achieves lower rollout error on
Burgers, Heat, Wave-B, and Vorticity. These results therefore show that
reconstruction quality and temporal predictability are distinct properties,
and that a representation with lower reconstruction error is not necessarily
easier to evolve accurately over time.

\subsection{Mechanism Analyses}
\label{app:mechanisms}

We investigate how geometry alignment and physics-structured prediction improve latent dynamics and long-horizon forecasting. We organize our mechanism analysis around four questions. 
\textbf{Q1: Does geometry projection align latent evolution with the underlying physical trajectory?} 
We examine full-dimensional latent trajectory geometry in subsection~\ref{app:mech_lat} and controlled 1D experiments in subsection~\ref{app:mech_ext} to determine whether the projected representation better reflects physical evolution. 
\textbf{Q2: Does the physics-structured predictor capture parameter-dependent dynamics more faithfully?} 
We compare predicted and numerical parameter responses at matched initial conditions in both latent and physical spaces in subsection~\ref{app:mech_param}. 
\textbf{Q3: Do these improvements translate into better forecasting under parameter shift?} 
We evaluate field- and latent-space rollout accuracy across ID and OOD regimes in subsection~\ref{app:mech_field}. 
\textbf{Q4: Why does physics supervision improve long-horizon extrapolation?} 
We separate one-step prediction defects from subsequent error amplification to identify the source of the long-horizon improvement in subsection~\ref{app:mech_physics_propagation}. 
Together, the analyses support all four questions: geometry projection improves physical alignment, the structured predictor better captures parameter-dependent dynamics, these gains improve field forecasting under parameter shift, and reduced error amplification provides a mechanism for the long-horizon extrapolation benefit.

\subsubsection{Evaluation protocol and dynamics diagnostics}
\label{app:Mech_eval}

\paragraph{Experimental setup.}
We analyze three JEPA-based systems on Vorticity: \emph{Vanilla}, which evolves standardized encoder features; \emph{Geo}, which evolves features transformed by the frozen geometry projector; and \emph{Physics}, which further introduces the structured parameter-dependent predictor. Wave2D provides compatible Geo and Physics systems. Within each dataset, Geo and Physics share the same frozen encoder, projector, decoder, numerical precision, and integration scheme, and all free rollouts start from the observed initial state. For Vorticity, we include a 120 Bridge data ranging viscosity between $[10^{-4},10^{-3}]$ for further analysis.

\paragraph{Parameter response.} For parameter-response analysis, each of the 120 OOD trajectories is paired with a training-reference trajectory sharing the same initial condition but a different physical parameter, so the two rollouts start from the same latent state and differ only in the supplied parameter. We evaluate the resulting finite parameter response over 29 future steps, while the 1,200 ID-validation trajectories are used separately for standard rollout evaluation. Because the OOD split reuses training initial conditions, this protocol tests parameter generalization at fixed initial conditions. A separate perturbation analysis uses depth-eight Vorticity predictors, denoted \emph{Free-D} and \emph{Physics-D}, where Physics-D adds supervision on the viscosity-dependent branch.The 1D controls likewise follow their own dataset-specific protocols.

For two different parameters $\xi_a$ and $\xi_b$, we compare the predicted response
$\Delta\widehat{x}_t=\widehat{x}_t(\xi_b)-\widehat{x}_t(\xi_a)$
with the numerical response
$\Delta x_t=x_t(\xi_b)-x_t(\xi_a)$, where $x$ denotes either the decoded field or latent state. We measure response direction (Cos.), magnitude (Amp.), and relative error as
\[
C_x(t)=
\frac{\langle\Delta\widehat{x}_t,\Delta x_t\rangle}
{\|\Delta\widehat{x}_t\|_2\|\Delta x_t\|_2},
\qquad
R_x(t)=
\frac{\|\Delta\widehat{x}_t\|_2}{\|\Delta x_t\|_2},
\qquad
E_x(t)=
\frac{\|\Delta\widehat{x}_t-\Delta x_t\|_2}
{\|\Delta x_t\|_2}.
\]
The targets are $C_x=R_x=1$ and $E_x=0$. Metrics are averaged over time and then across the 120 matched initial conditions, with paired confidence intervals obtained by resampling complete initial conditions 1,000 times. Geo and Physics are compared against the same frozen target representation $q_t=G(E(u_t))$, so their latent-response comparison is independent of the physical decoder.

\subsubsection{Parameter-dependent dynamics}
\label{app:mech_param}
\begin{figure}[h]
\centering
\includegraphics[width=\textwidth,height=0.24\textheight,keepaspectratio]{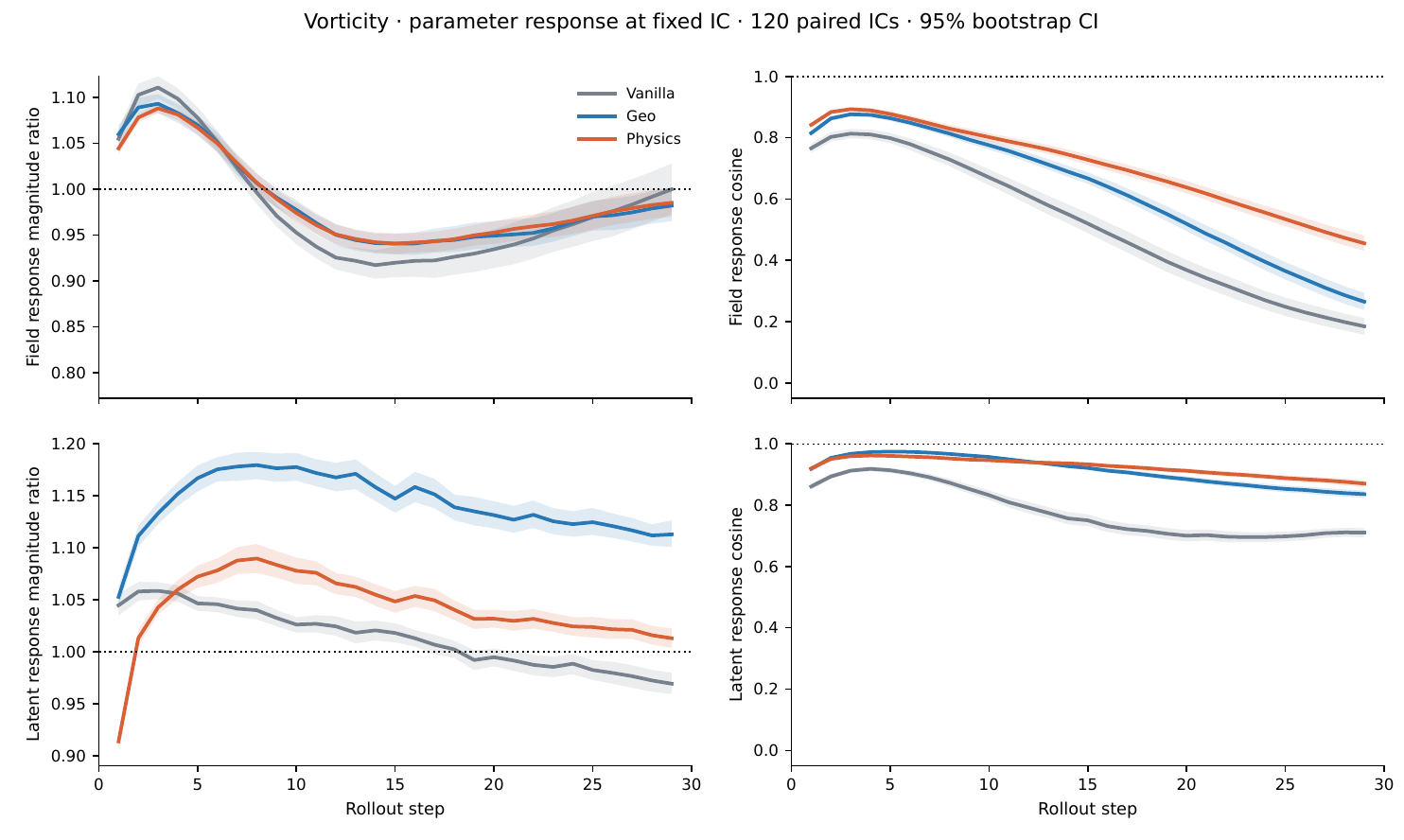}
\caption{\textbf{Vorticity parameter-response magnitude and direction at fixed initial conditions.} Curves average all 120 matched pairs; shaded bands are pointwise 95\% initial-condition bootstrap intervals. \textit{Upper row} measures decoded fields. \textit{Lower row} measures latent states. Both magnitude and cosine have target 1.}
\label{fig:mech_ns2d_response}
\end{figure}

On Vorticity, the main improvement comes from a more accurate direction of the
parameter-induced dynamics while preserving the response magnitude
(Table~\ref{tab:mech_response}; Figure~\ref{fig:mech_ns2d_response}).
The mean field-response cosine increases from 0.5161 for Vanilla to 0.6259 for
Geo and 0.7079 for Physics, whereas only physics maintain field-response magnitude
ratios close to one. In the shared projected
latent space, Physics further increases the response cosine from 0.9128 to
0.9240 and reduces Geo's magnitude overestimation from 1.1423 to 1.0428.
Thus, geometry alignment makes the parameter response substantially more
directionally consistent, while the structured predictor further improves its
direction and latent amplitude calibration.

\begin{table}[h]
\centering\small
\caption{\textbf{Parameter-response accuracy at fixed initial conditions.}}
\label{tab:mech_response}
\begin{adjustbox}{max width=\textwidth}\begin{tabular}{llrrrr}
\toprule
Dataset & Model & $C_u$ $\to 1$ & $R_u$ $\to 1$ & $C_q$ $\to 1$ & $R_q$ $\to 1$  \\
\midrule
Vorticity & Vanilla & 0.5161 & 0.7923 & 0.7802 & 1.0138 \\
Vorticity & Geo & 0.6259 & 0.8462 & 0.9128 & 1.1423 \\
Vorticity & Physics & 0.7079 & 0.9870 & 0.9240 & 1.0428 \\
Wave2D & Geo & 0.9582 & 0.9786 & 0.9389 & 0.9903 \\
Wave2D & Physics & 0.9861 & 0.9959 & 0.9465 & 1.0023 \\
\bottomrule
\end{tabular}\end{adjustbox}
\par\smallskip
\begin{minipage}{\textwidth}\footnotesize Means over 29 steps and 120 matched initial conditions. Cosine has target 1; the magnitude ratio also has target 1. Geo and Physics use the same unquantized projected latent target. Vanilla uses its own standardized encoder coordinates.
\end{minipage}
\end{table}

\begin{figure}[h]
\centering
\includegraphics[width=\textwidth,height=0.24\textheight,keepaspectratio]{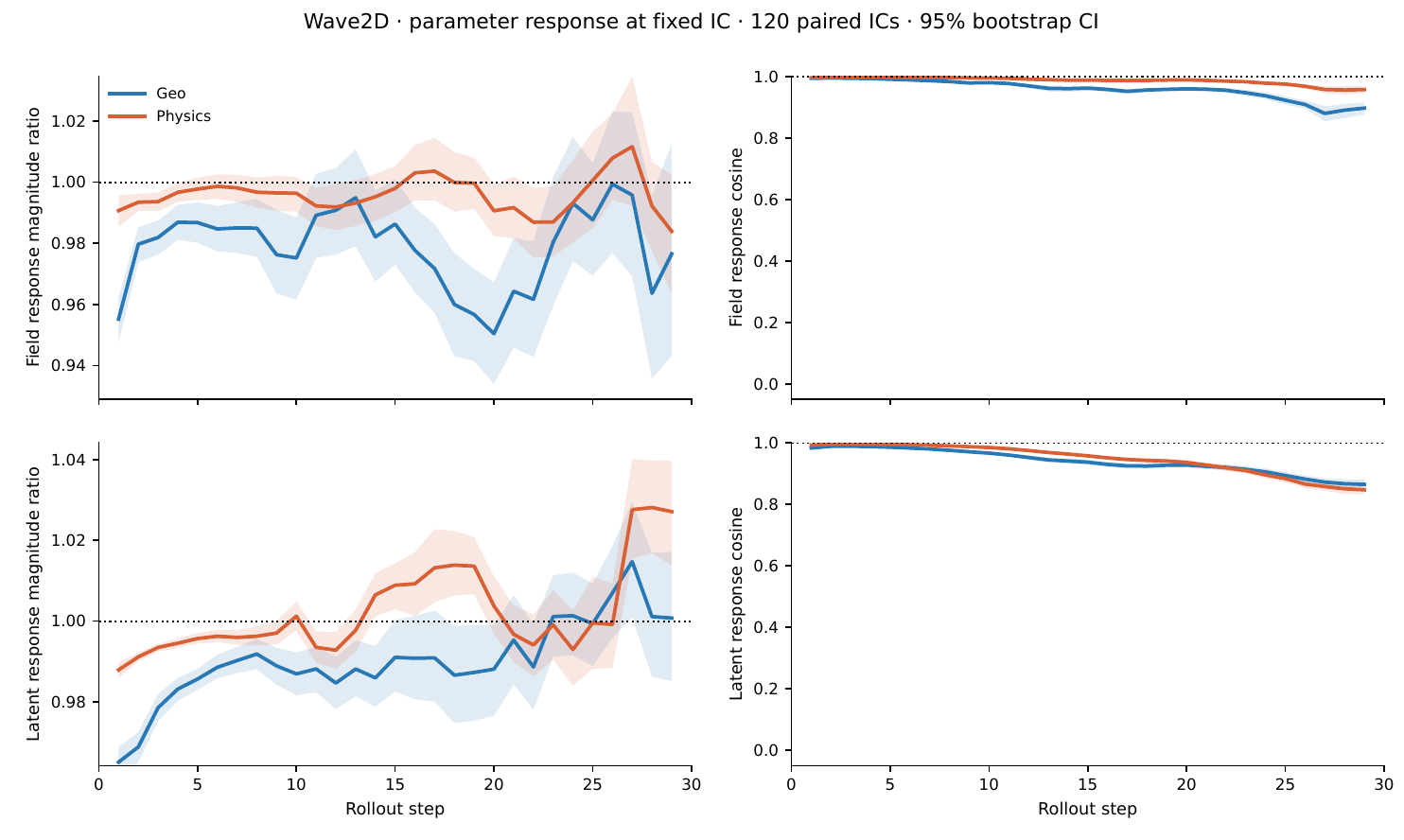}
\caption{\textbf{Wave2D parameter-response magnitude and direction at fixed initial conditions.} Curves average all 120 matched pairs; shaded bands are pointwise 95\% initial-condition bootstrap intervals. \textit{Upper row} measures decoded fields. \textit{Lower row} measures latent states. Both magnitude and cosine have target 1.}
\label{fig:mech_wave2d_response}
\end{figure}

Wave2D shows an even cleaner improvement in both response direction and
magnitude (Figure~\ref{fig:mech_wave2d_response}). Physics increases the
field-response cosine from 0.9582 to 0.9861 and moves the magnitude ratio from
0.9786 to 0.9959, closely matching the numerical parameter response. In latent
space, the cosine also improves from 0.9389 to 0.9465, while the magnitude
ratio moves from 0.9903 to 1.0023. These results show that the structured
predictor more faithfully captures the combined effect of changing $(c,k)$ in
both latent and physical space, with the largest gain appearing in the decoded
field response.

\begin{figure}[H]
\centering
\includegraphics[width=\textwidth,height=0.14\textheight,keepaspectratio]{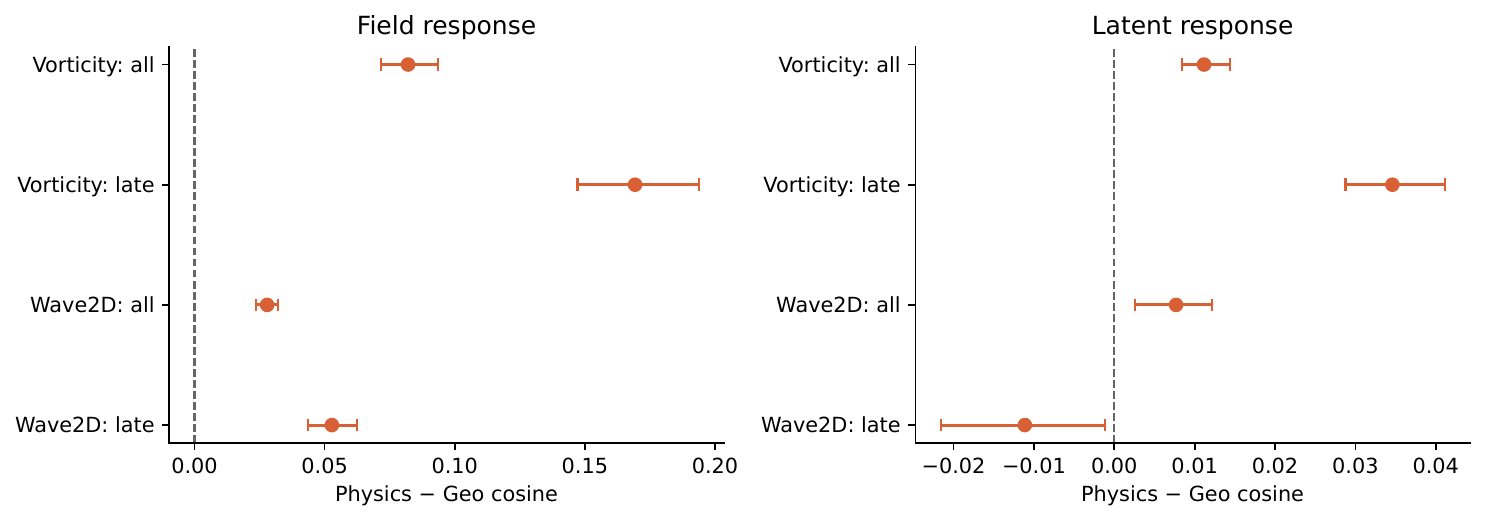}
\caption{\textbf{Paired changes in parameter-response direction}. With 95\% initial-condition bootstrap intervals. All denotes steps 1--29 and late denotes steps 22--29. The Wave2D late latent interval lies below zero, while its late field interval lies above zero.}
\label{fig:mech_response_intervals}
\end{figure}

The temporal decomposition exposes a meaningful difference between the equations (Table~\ref{tab:mech_response_delta}; Figure~\ref{fig:mech_response_intervals}). Over steps 22--29, Vorticiy retains positive field and latent cosine gains of 0.1693 and 0.0346. Wave2D retains a field cosine gain of 0.0528, but its latent cosine decreases from 0.8900 to 0.8789. The paired change is -0.0111 [-0.0216, -0.0011], and latent relative-response error increases by 0.0234 [0.0050, 0.0431]. Average improvement therefore coexists with a measurable late-horizon latent limitation. A single whole-rollout average would conceal this reversal.

\begin{table}[H]
\centering\small
\caption{\textbf{Paired changes in response accuracy: Physics minus Geo.}}
\label{tab:mech_response_delta}
\begin{adjustbox}{max width=\textwidth}\begin{tabular}{llrrr}
\toprule
Dataset & Steps & $\Delta C_u$ [95\% CI] & $\Delta C_q$ [95\% CI] & $\Delta E_q$ [95\% CI] \\
\midrule
NS & 1--29 & +0.0820 [0.0716, 0.0935] & +0.0112 [0.0084, 0.0144] & -0.0595 [-0.0695, -0.0509] \\
NS & 22--29 & +0.1693 [0.1472, 0.1937] & +0.0346 [0.0288, 0.0411] & -0.1067 [-0.1215, -0.0932] \\
Wave2D & 1--29 & +0.0279 [0.0236, 0.0321] & +0.0077 [0.0025, 0.0122] & -0.0368 [-0.0468, -0.0258] \\
Wave2D & 22--29 & +0.0528 [0.0435, 0.0625] & -0.0111 [-0.0216, -0.0011] & +0.0234 [0.0050, 0.0431] \\
\bottomrule
\end{tabular}\end{adjustbox}
\par\smallskip
\begin{minipage}{\textwidth}\footnotesize Positive cosine changes and negative relative-error changes favor Physics. Intervals use 1,000 paired bootstrap resamples of complete initial conditions; steps within an initial condition remain together.
\end{minipage}
\end{table}

\subsubsection{Latent trajectory geometry}
\label{app:mech_lat}
\begin{figure}[htbp]
\centering
\includegraphics[width=\textwidth,height=0.24\textheight,keepaspectratio]{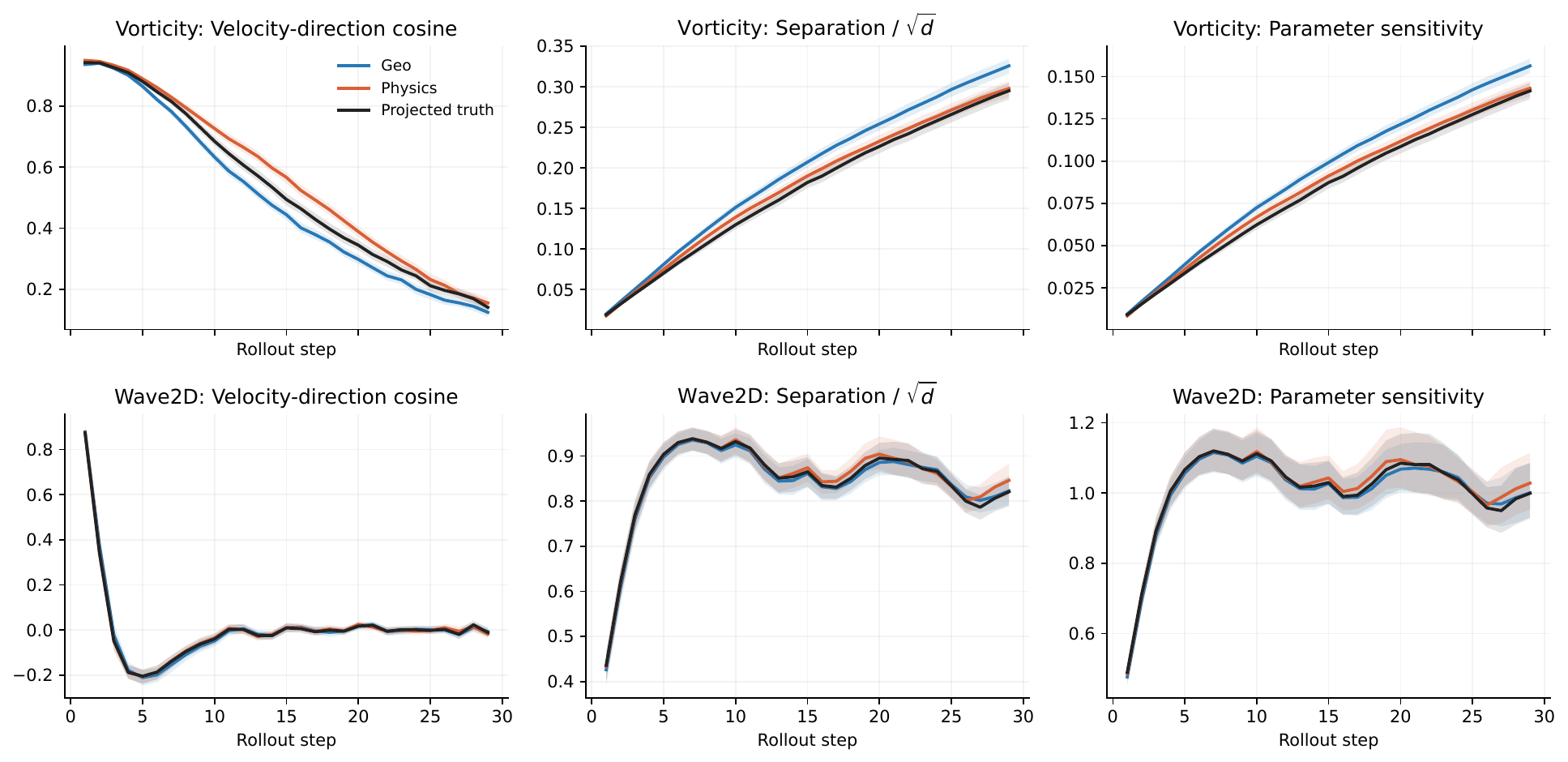}
\caption{\textbf{Full-dimensional geometry of the two trajectories generated from each matched initial condition. }Velocity-direction cosine compares the two parameter settings within a model, whereas response cosine compares predicted and true parameter-induced differences. Lower cross-parameter direction similarity can therefore be correct, as the Wave2D projected-truth curve illustrates. Shading uses 1,000 initial-condition bootstrap resamples.}
\label{fig:mech_matched_geometry}
\end{figure}

\begin{figure}[htbp]
\centering
\includegraphics[width=\textwidth,height=0.24\textheight,keepaspectratio]{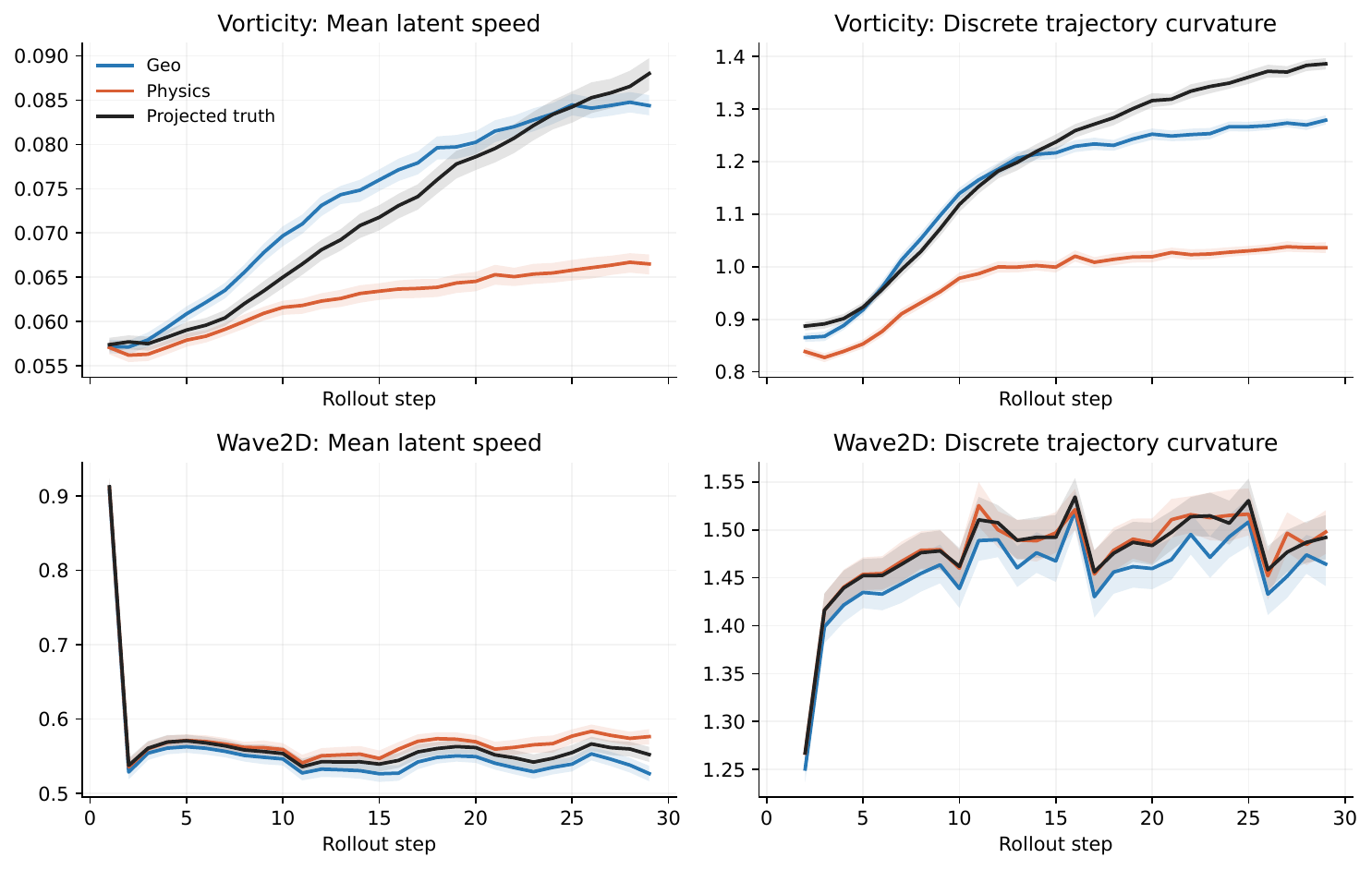}
\caption{\textbf{Speed and direction-change curvature in the shared projected state space.} Shading is a trajectory bootstrap interval. The first curvature value is undefined and omitted. Physics is smoother than projected truth on Vorticity; the Wave2D speed and curvature are slightly larger than Geo. Smoothness alone does not rank predictive accuracy.}
\label{fig:mech_geometry_time}
\end{figure}

This section examines how parameter changes reshape latent trajectories and whether the predicted trajectory geometry matches the projected numerical dynamics. We consider two complementary diagnostics. First, for matched initial conditions under two parameter settings, we measure cross-parameter trajectory separation and velocity-direction similarity (Figure~\ref{fig:mech_matched_geometry}). Second, for each individual trajectory, we measure speed, curvature, and path efficiency in the full projected latent space (Table~\ref{tab:mech_geometry}; Figure~\ref{fig:mech_geometry_time}). The projected numerical trajectory, $q_t=G(E(u_t))$, serves as the reference, while PCA is used only for qualitative visualization.

\paragraph{Cross-parameter trajectory geometry.}
On Vorticity, Physics more closely reproduces the late geometry of the projected numerical trajectories (Figure~\ref{fig:mech_matched_geometry}). Its cross-parameter separation is 0.2743 versus 0.2694 for the projected truth, compared with 0.2995 for Geo, while its velocity-direction similarity is 0.2296 versus 0.2124 for the projected truth and 0.1804 for Geo. On Wave2D, the numerical trajectories under the two parameter settings are themselves nearly orthogonal at late times, with a velocity-direction similarity of $-0.0010$; Physics closely reproduces this behavior at $-0.0014$, compared with $-0.0026$ for Geo. Thus, cross-parameter direction similarity should be interpreted relative to the numerical dynamics rather than as a standalone measure of organization.

\begin{figure}[htbp]
\centering
\includegraphics[width=\textwidth,height=0.74\textheight,keepaspectratio]{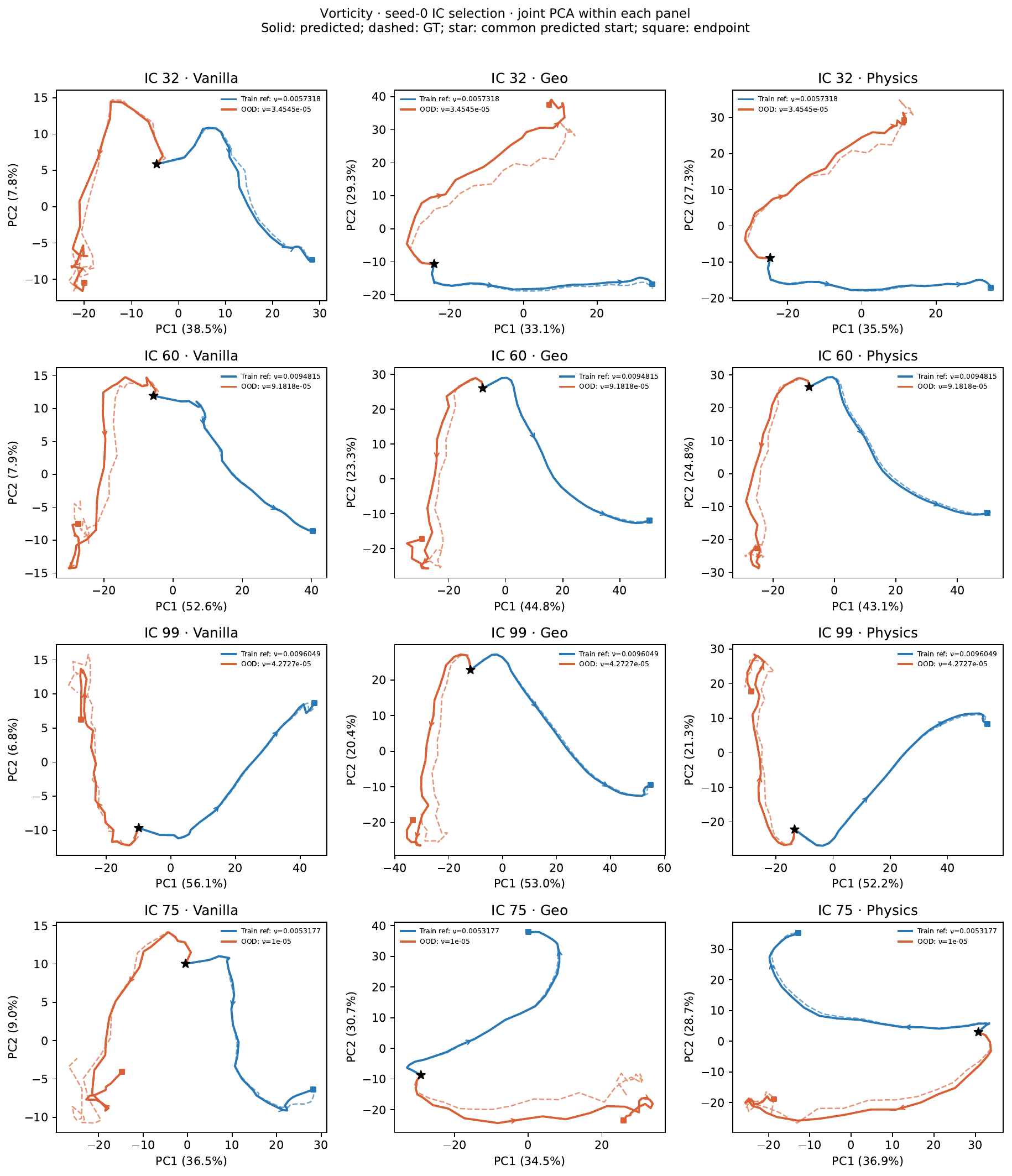}
\caption{\textbf{Vorticity predicted and true latent trajectories for all four initial conditions selected. Each panel fits one joint PCA to the two predicted and two true trajectories.} Solid curves are predictions; dashed curves are truth. Stars mark the common predicted initial state.}
\label{fig:mech_ns2d_family}
\end{figure}

\begin{figure}[htbp]
\centering
\includegraphics[width=\textwidth,height=0.74\textheight,keepaspectratio]{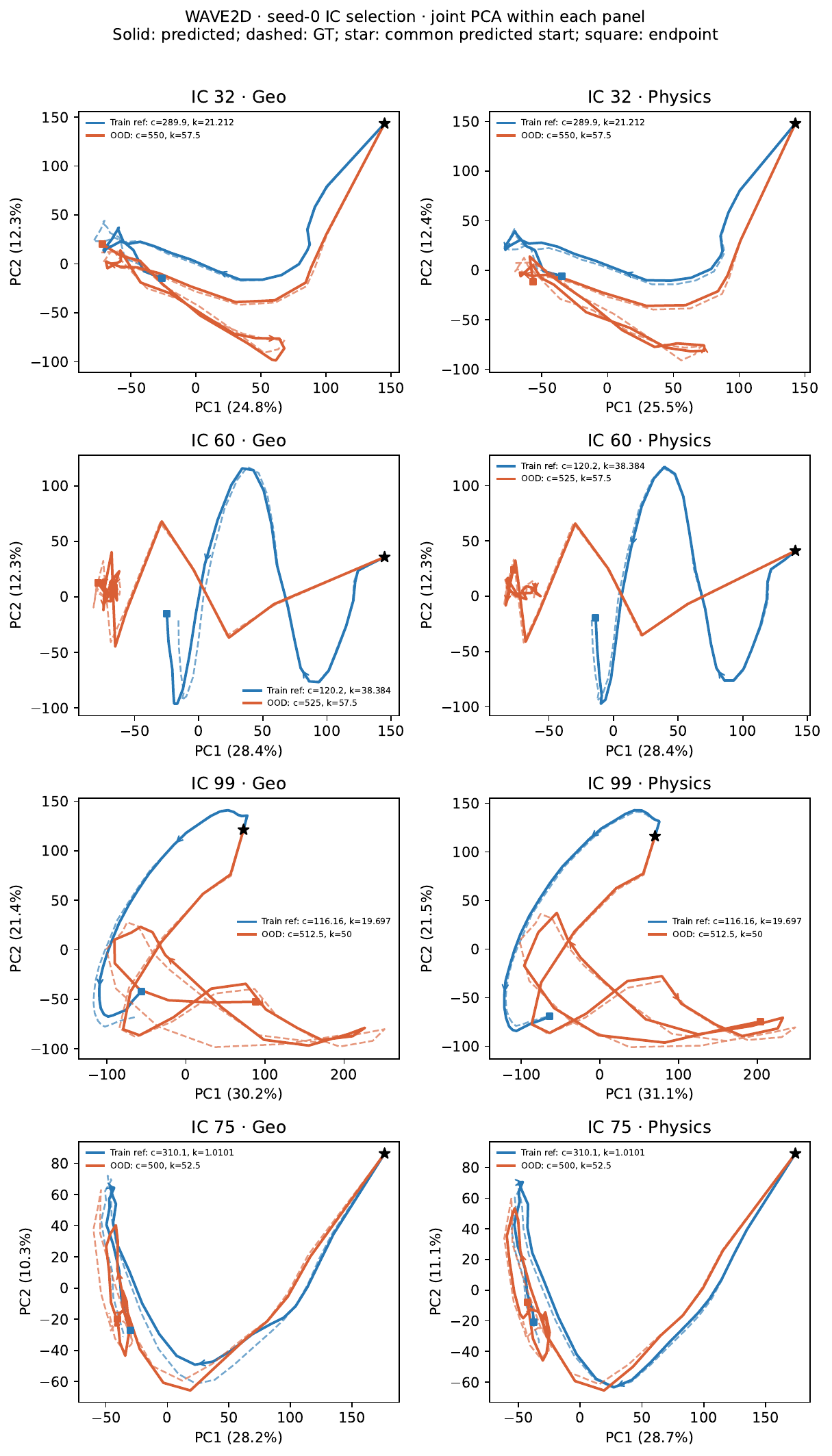}
\caption{\textbf{Wave2D predicted and true latent trajectories for all four initial conditions selected.} Each panel fits one joint PCA to the two predicted and two true trajectories. Solid curves are predictions; dashed curves are truth. Stars mark the common predicted initial state.}
\label{fig:mech_wave2d_family}
\end{figure}

The corresponding PCA trajectory families provide qualitative examples of how the same initial state branches under different parameter settings (Figures~\ref{fig:mech_ns2d_family} and~\ref{fig:mech_wave2d_family}). These visualizations are illustrative only; all quantitative geometry metrics are computed in the full latent space.

\paragraph{Single-trajectory temporal geometry.}

The aggregate trajectory statistics reveal different behaviors across equations (Table~\ref{tab:mech_geometry}; Figure~\ref{fig:mech_geometry_time}). On Wave2D, Physics more closely matches the projected numerical speed and curvature than Geo: its mean speed is 0.5759 versus 0.5664 for the projected truth, while its curvature is 1.4768 versus 1.4755. On Vorticity, however, Physics produces a slower and smoother trajectory than the projected numerical reference, with speed 0.0624 versus 0.0718 and curvature 0.9768 versus 1.1934. Despite this departure, Physics achieves better parameter-response direction and lower field rollout error. These results show that smoother latent dynamics are not necessarily more physical; the relevant criterion is agreement with the projected numerical trajectory rather than smoothness alone.

\begin{table}[h]
\centering\small
\caption{\textbf{Trajectory geometry in the shared projected coordinates on the original OOD split.}}
\label{tab:mech_geometry}
\begin{adjustbox}{max width=\textwidth}\begin{tabular}{llrrr}
\toprule
Dataset & Source & Mean speed & Mean curvature & Path efficiency \\
\midrule
Vorticity & Projected truth & 0.0718 & 1.1934 & 0.1449 \\
Vorticity & Geo & 0.0737 & 1.1557 & 0.1452 \\
Vorticity & Physics & 0.0624 & 0.9768 & 0.1673 \\
Wave2D & Projected truth & 0.5664 & 1.4755 & 0.0775 \\
Wave2D & Geo & 0.5548 & 1.4539 & 0.0784 \\
Wave2D & Physics & 0.5759 & 1.4768 & 0.0759 \\
\bottomrule
\end{tabular}\end{adjustbox}
\par\smallskip
\begin{minipage}{\textwidth}\footnotesize Each row averages 120 trajectories. Speed is the step displacement norm divided by the square root of latent dimension; curvature is the change in normalized step direction. Projected truth is unquantized.
\end{minipage}
\end{table}

\subsubsection{Field and latent rollout accuracy}
\label{app:mech_field}
\begin{figure}[H]
\centering
\includegraphics[width=\textwidth,height=0.24\textheight,keepaspectratio]{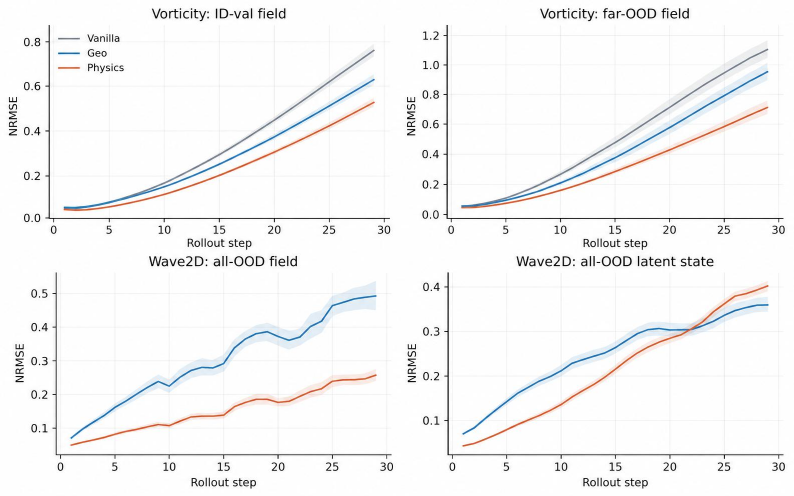}
\caption{\textbf{Original-protocol rollout diagnostics.} Vorticity panels show ID-val and the farthest available OOD viscosity; Wave2D panels show all original OOD trajectories in field and latent space.}
\label{fig:mech_rollout}
\end{figure}

To connect parameter-response accuracy with forecasting, we report the mean per-frame NRMSE over steps 1--29 (Table~\ref{tab:mech_rollout}). On Vorticity, Physics reduces field error from 0.0429 to 0.0362 on ID validation and from 0.3917 to 0.2966 on OOD trajectories, corresponding to reductions of 15.6\% and 24.3\%. Wave2D shows a larger improvement in physical space: OOD field error decreases from 0.3039 to 0.1521 and far-OOD error from 0.2652 to 0.1405, while the corresponding latent-space gains are smaller and reverse under the far-OOD condition (0.2212$\rightarrow$0.2366). The horizon curves show that these differences persist across rollout (Figure~\ref{fig:mech_rollout}). Overall, the structured predictor consistently improves physical-field forecasting, although lower field error does not require uniformly lower latent error.

\begin{table}[htbp]
\centering\small

\caption{\textbf{Average rollout error on the original evaluation splits.}}
\label{tab:mech_rollout}
\begin{adjustbox}{max width=\textwidth}
\begin{tabular}{llrrrr}
\toprule
Dataset & Space & Split & Geo AUEC $\downarrow$ & Physics AUEC $\downarrow$ & Reduction (\%) $\uparrow$\\
\midrule

Vorticity & Field  & ID-val  & 0.0429 & 0.0362 & 15.61 \\
Vorticity & Field  & OOD     & 0.3917 & 0.2966 & 24.27 \\
Wave2D    & Field  & ID-val  & 0.1548 & 0.1083 & 30.01 \\
Wave2D    & Field  & OOD     & 0.3039 & 0.1521 & 49.94 \\
Wave2D    & Field  & Far-OOD & 0.2652 & 0.1405 & 47.03 \\
Wave2D    & Latent & ID-val  & 0.1447 & 0.1357 & 6.18 \\
Wave2D    & Latent & OOD     & 0.2479 & 0.2168 & 12.54 \\
Wave2D    & Latent & Far-OOD & 0.2212 & 0.2366 & -6.97 \\

\bottomrule
\end{tabular}
\end{adjustbox}
\par\smallskip
\begin{minipage}{\textwidth}\footnotesize
AUEC averages per-frame NRMSE over steps 1--29. OOD uses all 120 original OOD trajectories; the Wave2D far-OOD group contains ten trajectories at $(c,k)=(512.5,60)$.
\end{minipage}

\end{table}

\begin{figure}[H]
\centering
\includegraphics[width=\textwidth,height=0.24\textheight,keepaspectratio]{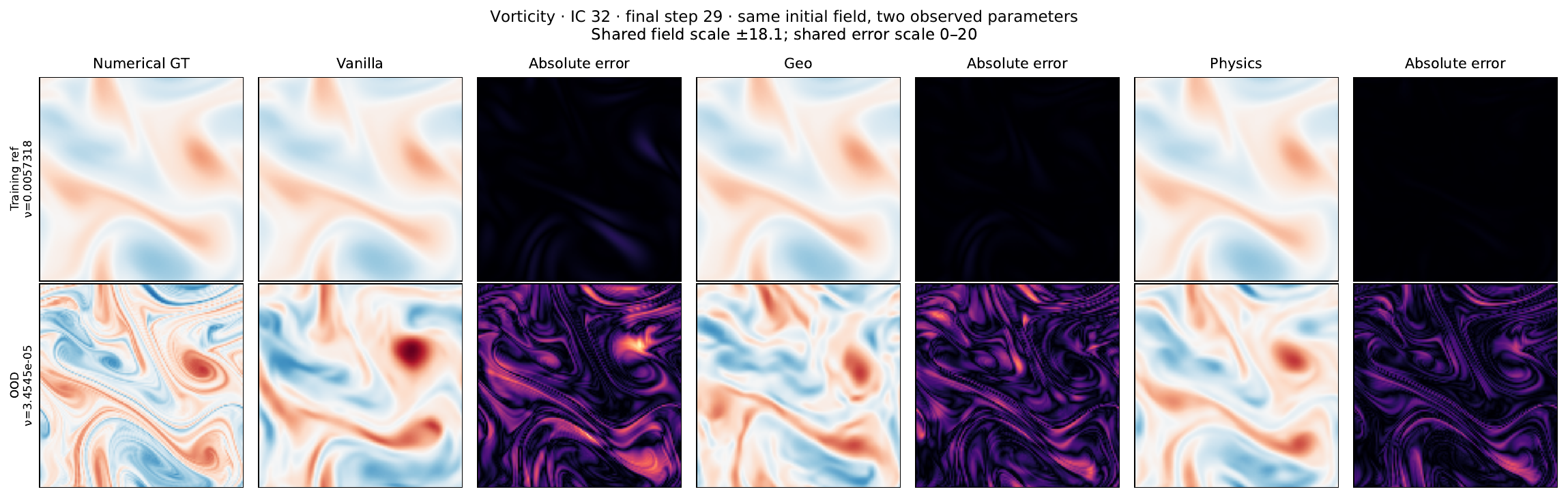}
\caption{\textbf{Vorticity final-step fields for IC 32}, with the training-reference parameter above and the OOD parameter below. The two parameter settings start from exactly the same vorticity field.}
\label{fig:mech_ns2d_field_32}
\end{figure}

\begin{figure}[H]
\centering
\includegraphics[width=\textwidth,height=0.24\textheight,keepaspectratio]{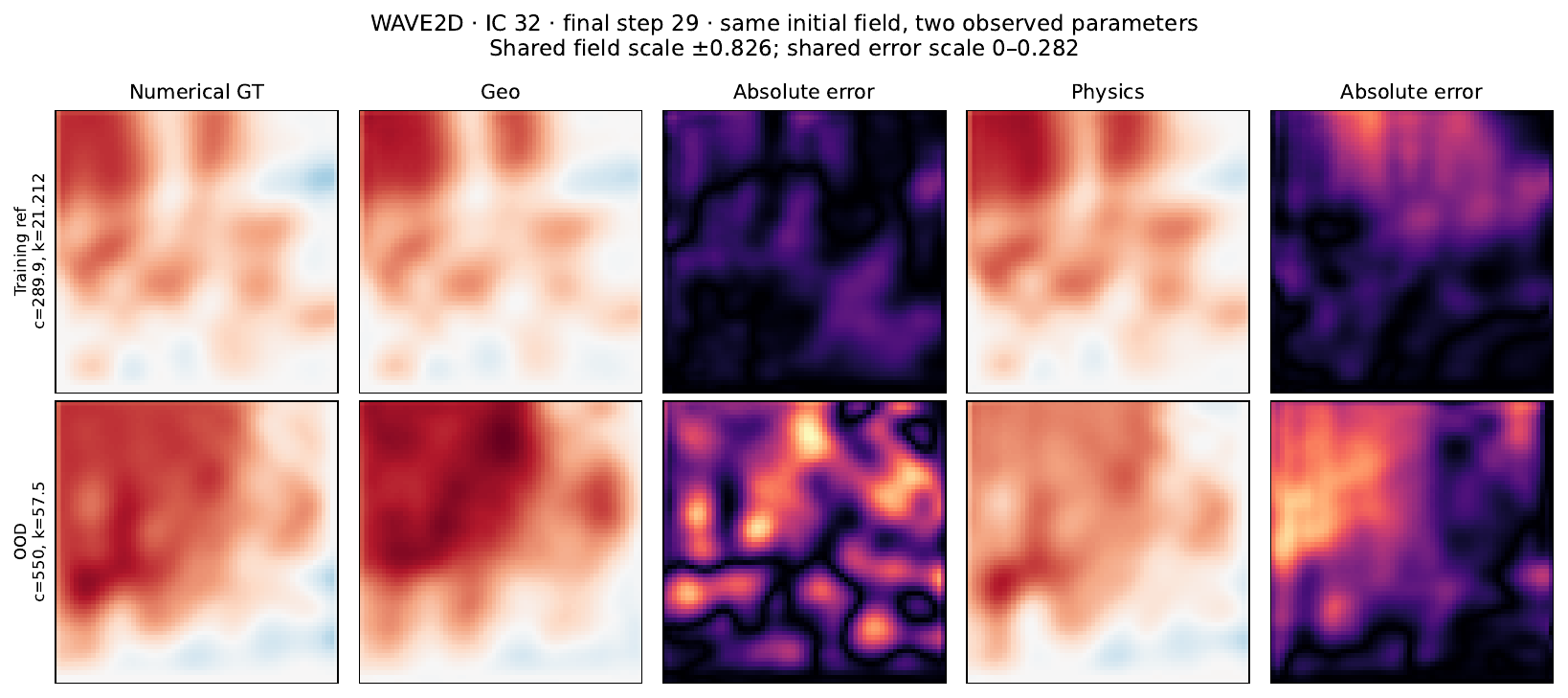}
\caption{\textbf{Wave2D final-step fields for IC 32}, with the training-reference parameter above and the OOD parameter below. }
\label{fig:mech_wave2d_field_32}
\end{figure}

The final-step field visualizations put these averages alongside individual parameter changes (Figure~\ref{fig:mech_ns2d_field_32}; Figure~\ref{fig:mech_wave2d_field_32}). The Wave2D example shows smaller OOD field errors for Physics even though the preceding latent metrics identify a late-horizon limitation.

\subsubsection{Physics Supervision Reduces Error Propagation}
\label{app:mech_physics_propagation}
To understand why physics supervision improves long-horizon extrapolation, we
separate two sources of rollout error: the error introduced at each prediction
step and the subsequent propagation of existing errors. Let
$q_t=G(E(u_t))$ denote the projected numerical trajectory,
$\widehat q_t$ the free rollout, and $\Psi_\theta$ one prediction step.
Defining the rollout error $e_t=\widehat q_t-q_t$ and the one-step defect
$r_t=\Psi_\theta(q_t)-q_{t+1}$ gives
\[
e_{t+1}
=
r_t+
\bigl[\Psi_\theta(q_t+e_t)-\Psi_\theta(q_t)\bigr]
\approx
r_t+A_t e_t,
\qquad
A_t=J\Psi_\theta(q_t).
\]
Thus, long-horizon error depends not only on one-step prediction quality, but
also on how strongly existing errors are amplified during rollout.

We test this mechanism using the Free-D and Physics-D Vorticity predictors
described in Section~\ref{app:Mech_eval}. In addition to ID and far-OOD
evaluation, we use an intermediate Bridge regime with
$\nu\in[10^{-4},10^{-3}]$, lying between the training range
($\sim[10^{-3},10^{-2}]$) and the far-OOD range
($[10^{-5},10^{-4}]$). To quantify error propagation, we measure the
finite-time perturbation gain
\begin{equation}
A_K(q,\delta q)
=
\frac{
\|F_\theta^{(K)}(q+\delta q)-F_\theta^{(K)}(q)\|_2
}{
\|\delta q\|_2
},
\label{eq:mech_perturbation_gain}
\end{equation}
where $F_\theta^{(K)}$ denotes the $K$-step prediction map. Larger values
indicate stronger amplification of an initial latent perturbation.

\begin{figure}[htbp]
\centering
\includegraphics[
    width=\textwidth,
    height=0.65\textheight,
    keepaspectratio
]{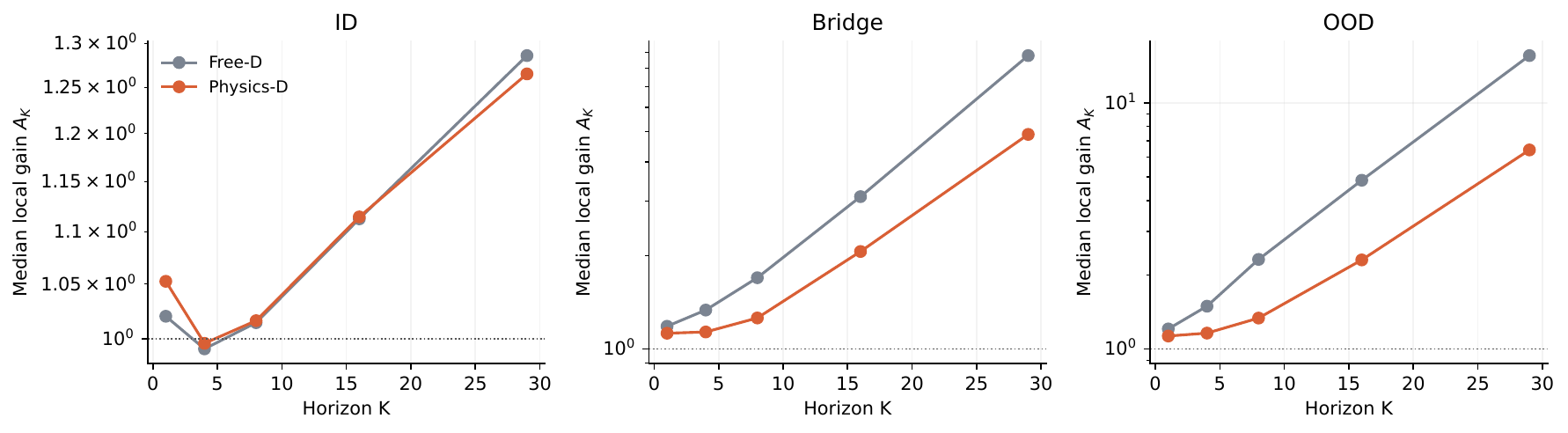}
\caption{
\textbf{Finite-time perturbation amplification on Vorticity.}
Physics-D leaves ID amplification nearly unchanged but substantially suppresses
long-horizon amplification on Bridge and far-OOD trajectories.
}
\label{fig:mech_gain_horizon}
\end{figure}

Physics-D substantially reduces long-horizon amplification under distribution
shift (Figure~\ref{fig:mech_gain_horizon};
Table~\ref{tab:mech_defect_propagation}). At $K=29$, the median gain changes
only slightly on ID data, from $1.2860$ to $1.2653$, but decreases from
$8.8056$ to $4.9087$ on Bridge and from $15.5361$ to $6.4286$ on far-OOD.
On the far-OOD trajectories, Physics-D has lower gain in 98.3\% of cases.
The horizon curves further show that this difference grows with rollout
length, indicating that physics supervision mainly changes the stability of
long-horizon evolution rather than the local ID dynamics.

Importantly, the lower amplification is not explained by better one-step
fitting. As shown in Table~\ref{tab:mech_defect_propagation}, Physics-D has
slightly larger teacher-forced error on all three splits:
$0.002188$ versus $0.002086$ on ID,
$0.020687$ versus $0.019979$ on Bridge, and
$0.027629$ versus $0.026543$ on far-OOD.
Nevertheless, its long-horizon errors improve under distribution shift.
On far-OOD trajectories, the final-step latent error decreases from
$0.169647$ to $0.159624$, while the whole-rollout error decreases from
$0.096307$ to $0.090602$. Bridge shows the same trend with a smaller margin,
whereas ID rollout error slightly increases.

\begin{table}[htbp]
\centering
\small
\caption{
\textbf{One-step prediction error, perturbation amplification, and latent
rollout error on Vorticity.}
}
\label{tab:mech_defect_propagation}
\begin{adjustbox}{max width=\textwidth}
\begin{tabular}{llrrrrr}
\toprule
Split & Model & $N$ & Teacher error & $A_{29}$
& Final latent error & Whole latent error \\
\midrule
ID & Free-D
& 120 & 0.002086 & 1.2860 & 0.036409 & 0.022268 \\
ID & Physics-D
& 120 & 0.002188 & 1.2653 & 0.040790 & 0.025092 \\
Bridge & Free-D
& 240 & 0.019979 & 8.8056 & 0.135650 & 0.077281 \\
Bridge & Physics-D
& 240 & 0.020687 & 4.9087 & 0.133460 & 0.075893 \\
OOD & Free-D
& 120 & 0.026543 & 15.5361 & 0.169647 & 0.096307 \\
OOD & Physics-D
& 120 & 0.027629 & 6.4286 & 0.159624 & 0.090602 \\
\bottomrule
\end{tabular}
\end{adjustbox}
\end{table}

Perturbation amplification is also associated with long-horizon prediction
error: across the 480 evaluated initial states, the Spearman correlation
between log gain and final-step latent error is approximately $0.81$ for both
predictors. This association weakens after controlling for viscosity,
indicating that amplification explains only part of the rollout behavior.

Overall, these results separate one-step prediction error from error
propagation. Physics-D does not improve the former and slightly worsens it,
but substantially reduces the latter under Bridge and far-OOD conditions.
The resulting reduction in long-horizon rollout error therefore provides
direct evidence that suppressed error amplification is an important
contributor to the extrapolation benefit of physics supervision.

\subsubsection{Extension to 1D PDEs}
\label{app:mech_ext}

The one-dimensional experiments provide two complementary controls on the role of geometry. First, Burgers, Heat, and Wave-B compare the same held-out trajectories before and after the frozen geometry projection. Each test set contains 120 trajectories which is different from Table~\ref{tab:geo_alignment} with 25 observed frames, and all metrics use the full native state vectors. Projection reduces the turning-angle MAE from 46.0253 to 23.3122 degrees on Burgers, from 78.1210 to 23.0318 on Heat, and from 13.3181 to 5.7452 on Wave-B; the weighted multi-lag geometry error also decreases on all three datasets (Table~\ref{tab:mech_one_d_geometry}).

\begin{table}[htbp]
\centering
\small
\caption{\textbf{Geometry projection on held-out one-dimensional trajectories.}}
\label{tab:mech_one_d_geometry}
\begin{adjustbox}{max width=\textwidth}
\begin{tabular}{llrrrr}
\toprule
Dataset & $N$ & Angle MAE: $z$ & Angle MAE: $q$ & Multi-lag: $z$ & Multi-lag: $q$ \\
\midrule
Burgers & 120 & 46.0253 & 23.3122 & 0.2387 & 0.0530 \\
Heat    & 120 & 78.1210 & 23.0318 & 0.6688 & 0.0150 \\
Wave-B  & 120 & 13.3181 & 5.7452  & 0.0746 & 0.0184 \\
\bottomrule
\end{tabular}
\end{adjustbox}
\par\smallskip
\begin{minipage}{\textwidth}
\footnotesize
All metrics use the full 120-trajectory ID-test split and 25 observed frames. Angle errors are in degrees; multi-lag error compares latent and physical increment-direction cosines at lags 1, 2, and 4. Smaller is better.
\end{minipage}
\end{table}

Wave-B further shows that the objective is alignment rather than simply reducing curvature. Its mean physical turning angle is 118.7228 degrees, compared with 111.8439 for the encoder state and 118.4202 after projection. Thus, projection can either increase or decrease latent turning as needed to better match the physical trajectory geometry. These results isolate a coordinate-level effect of Geo, but do not imply that lower geometry error necessarily guarantees lower forecast error.

\subsection{Extended Experiments and Ablation Studies}  

\subsubsection{Full OOD results}
\label{app:full_ood}
Our method achieves the best OOD rollout performance across all evaluated PDE benchmarks~\ref{tab:ood_main_full}. The largest gains are observed on Combined and Wave2D, where the error is reduced from the previous best 0.038 to 0.008 and from 0.610 to 0.157, respectively. On Vorticity, our method further improves over Zebra from 0.320 to 0.288, while substantial gains are also obtained on HeterNS and Gray--Scott. These results demonstrate consistently stronger parameter extrapolation across heterogeneous PDE dynamics.

\begin{table}[H]
\centering
\captionsetup{width=\linewidth}
\caption{\textbf{Full OOD rollout performance across PDE benchmarks.}
Relative $L^2$ error is reported, lower is better.
Best results are \textbf{bold} and second-best results are \underline{underlined}. - means the method diverges.}
\label{tab:ood_main_full}

\footnotesize
\setlength{\tabcolsep}{2.2pt}
\renewcommand{\arraystretch}{1.08}

\begin{tabular}{lccccc}
\toprule
Method
& Combined
& Wave-2D
& Vorticity
& \makecell{HeterNS\\Visc./Force}
& GS \\
\midrule

FNO
& -
& -
& 0.447
& -/-
& 0.192 \\

GEPS
& 0.570
& 1.005
& 0.421
& 0.162/-
& 0.184 \\

UniSolver
& \underline{0.038}
& 1.003
& 0.923
& \underline{0.037}/\underline{0.105}
& 0.1636 \\

MPP
& -
& -
& 0.845
& -/-
& - \\

DPOT
& -
& 1.197
& 0.483
& 2.43/1.47
& 0.597 \\

Poseidon-T
& 0.146
& 1.511
& 0.665
& 0.560/0.821
& \underline{0.083} \\

LE-PDE
& 0.169
& 0.924
& 0.715
& 0.334/0.923
& 0.239 \\

LNS
& 0.1667
& \underline{0.610}
& 0.481
& 0.610/0.932
& 0.146 \\

MAE-PDE
& 0.254
& 1.070
& 0.503
& 0.045/0.248
& 0.196 \\

ENMA
& 0.243
& 1.151
& 0.467
& 1.501/2.271
& 0.134 \\


\makecell[l]{Zebra}
& --
& 0.680
& \underline{0.320}
& --
& -- \\

\midrule
\textbf{Ours}
& \textbf{0.008}
& \textbf{0.157}
& \textbf{0.288}
& \textbf{0.011}/\textbf{0.103}
& \textbf{0.033} \\

\textit{Rel. Impr.}
& 77.9\%
& 74.2\%
& 9.7\%
& 69.5\%/1.5\%
& 59.7\% \\

\bottomrule
\end{tabular}
\end{table}

\subsubsection{Geometry Alignment Module}

\begin{figure}[h]
    \centering
    \includegraphics[width=\linewidth]{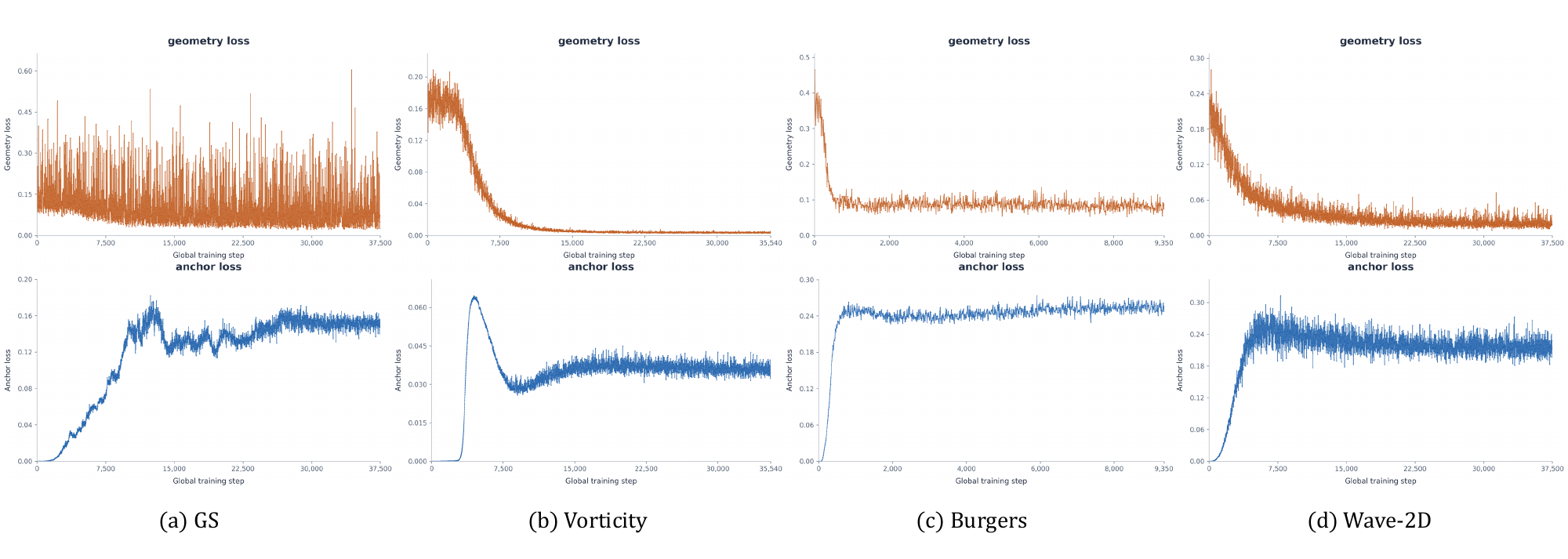}
    \caption{\textbf{Anchor and Geometry training loss for four benchmarks.}}
    \label{fig:anchor_loss_curve}
\end{figure}

\paragraph{Optimization behavior of geometry alignment.}
We first examine the optimization behavior of the geometry projector.
Figure~\ref{fig:anchor_loss_curve} reports both the geometry loss and
the anchor loss during training on four representative PDE systems.
The two objectives exhibit a consistent complementary pattern.
For Vorticity, Burgers, and Wave-2D, the geometry loss decreases rapidly
during the early stage of training and subsequently approaches a stable
regime, indicating that the projector progressively aligns the latent
trajectory with the geometry of the physical evolution. Gray--Scott
shows substantially noisier geometry optimization, but the loss remains
bounded throughout training.

In parallel, the anchor loss increases from its initially small value
and gradually stabilizes at a dataset-dependent level. Burgers reaches
this regime almost immediately, while Vorticity exhibits a transient
overshoot before settling to a lower plateau. Wave-2D stabilizes after
the initial rise with a mild downward drift, whereas Gray--Scott
undergoes a longer adjustment before reaching a relatively steady
regime. Together, these curves suggest that geometry alignment does not
proceed by unconstrained deformation of the pretrained representation:
as the geometry discrepancy is reduced, the anchor term becomes active
and limits excessive deviation from the original latent structure.
The absence of persistent growth in either objective further indicates
that the two constraints can be jointly optimized across PDE systems
with substantially different dynamics. This complementary behavior motivates examining whether the anchor
constraint is necessary when lower geometric discrepancy can otherwise
be achieved through more aggressive deformation of the latent space.

Although the anchor loss stabilizes during training, its role is not
to simply minimize the geometry discrepancy. We therefore remove the
anchor while keeping the remaining alignment objective unchanged.
As shown in Table~\ref{tab:anchor_ablation}, removing the anchor
further reduces the turning-angle MAE from $6.3^\circ$ to $4.8^\circ$
on Vorticity and from $10.1^\circ$ to $7.6^\circ$ on Wave-2D.
However, the resulting representations become substantially less
informative: local-state probing deteriorates across all evaluated
datasets, accompanied by considerably higher parameter-probing error.
The ID/OOD rollout errors also increase from $0.040/0.397$ to
$0.047/0.423$ on Vorticity and from $0.143/0.321$ to $0.166/0.354$
on Wave-2D. Thus, lower geometric error alone does not imply a better
state space for dynamics prediction. The anchor constraint prevents
over-deformation of the pretrained representation, balancing trajectory
alignment with preservation of dynamics-relevant information.

\begin{table*}[h]
\centering
\scriptsize
\setlength{\tabcolsep}{7pt}
\renewcommand{\arraystretch}{1.10}
\caption{
\textbf{Effect of the anchor constraint in geometry alignment.}
Removing the anchor further decreases the turning-angle error, but
substantially degrades the information retained in the representation
and leads to worse ID and OOD rollout accuracy.
}
\label{tab:anchor_ablation}
\begin{tabular}{llccccc}
\toprule
Dataset & Variant
& Angle MAE ($^\circ$) $\downarrow$
& Local-state $R^2$ $\uparrow$
& Param. MSE $\downarrow$
& ID Rollout $\downarrow$
& OOD Rollout $\downarrow$ \\
\midrule

Wave-2D
& + Anchor
& 10.1 & \textbf{0.87} & \textbf{8.63}
& \textbf{0.143} & \textbf{0.321} \\
& w/o Anchor
& \textbf{7.6} & 0.74 & 31.7
& 0.166 & 0.354 \\

\addlinespace[2pt]
Vorticity
& + Anchor
& 6.3 & \textbf{0.94} & \textbf{0.009}
& \textbf{0.040} & \textbf{0.397} \\
& w/o Anchor
& \textbf{4.8} & 0.81 & 0.024
& 0.047 & 0.423 \\

\bottomrule
\end{tabular}
\end{table*}

\paragraph{Geo modules behavior on AE.} We further examine whether geometry alignment provides a representation-agnostic
benefit or specifically complements predictive representations. For LNS, we freeze
the pretrained encoder and apply the same geometry projector as in PDE-JEPA. We then
train the dynamics predictors from scratch on the original and geometry-aligned
latent spaces using identical architectures, capacities, and optimization budgets.
As shown in Table~\ref{tab:lns_geo}, the original LNS representation already exhibits
a substantially smaller trajectory-angle discrepancy than JEPA
($9.37^\circ$ vs.\ $29.1^\circ$) and a lower ID rollout error
($0.059$ vs.\ $0.086$). Applying geometry alignment to JEPA reduces the angle
discrepancy to $6.30^\circ$ and simultaneously improves rollout error to $0.040$.
In contrast, the same alignment only slightly reduces the LNS angle discrepancy
from $9.37^\circ$ to $8.21^\circ$, while degrading rollout accuracy from $0.059$
to $0.062$. More aggressive deformation without the anchor further reduces the
angle error to $5.94^\circ$, but increases rollout error to $0.071$.
These results show that improved physical-geometry agreement does not universally
translate into better latent dynamics. Instead, geometry alignment specifically
addresses the substantial geometric mismatch of JEPA representations, whereas
deforming the reconstruction-oriented LNS latent space can disturb its learned
dynamical organization.

\begin{table}[h]
\centering
\caption{\textbf{Effect of geometry alignment on JEPA and LNS representations.}
Angle MAE measures the discrepancy between latent and physical trajectory geometry.
Anchor deviation is computed as $\|q-z\|_2^2 / (\|z\|_2^2+\epsilon)$.
Lower is better for all metrics.}
\label{tab:lns_geo}

\small
\setlength{\tabcolsep}{5pt}
\renewcommand{\arraystretch}{1.08}

\begin{tabular}{llccc}
\toprule
Representation
& Variant
& \makecell{Angle MAE\\($^\circ$) $\downarrow$}
& \makecell{Anchor\\Deviation $\downarrow$}
& \makecell{ID Rollout\\Rel. $L^2$ $\downarrow$} \\
\midrule

JEPA
& Original
& 29.1
& 0
& 0.086 \\

JEPA
& + Geo
& \textbf{6.30}
& 0.08
& \textbf{0.040} \\

\midrule

LNS
& Original
& 9.37
& 0
& \textbf{0.059} \\

LNS
& + Geo
& 8.21
& \textbf{0.03}
& 0.062 \\

LNS
& + Geo w/o Anchor
& \textbf{5.94}
& 0.21
& 0.071 \\

\bottomrule
\end{tabular}
\end{table}

\subsubsection{Predictor Module}

\paragraph{Structured prediction and capacity on Gray--Scott.}
Gray--Scott supplies an additional matched comparison between a direct
conditional transition and a structured continuous predictor,
\[
\dot{\mathbf q}
=
\mathbf C_{\theta}(\mathbf q)
+
\frac{F}{s_F}\mathbf D^F_{\theta}(\mathbf q)
+
\frac{k}{s_k^{\rm GS}}\mathbf D^k_{\theta}(\mathbf q)
\]
where $F$ and $k$ are the reaction parameters and the scales are fixed
from training. They share the frozen latent cache, sampled
transitions, global batch of 64, optimizer schedule, 50 training epochs,
and ID-validation selection rule. The continuous predictor uses RK4
with four substeps. The comparison tests the complete prediction rule,
including its parameterization and numerical integration.

\begin{figure}[H]
\centering
\includegraphics[width=0.88\textwidth]{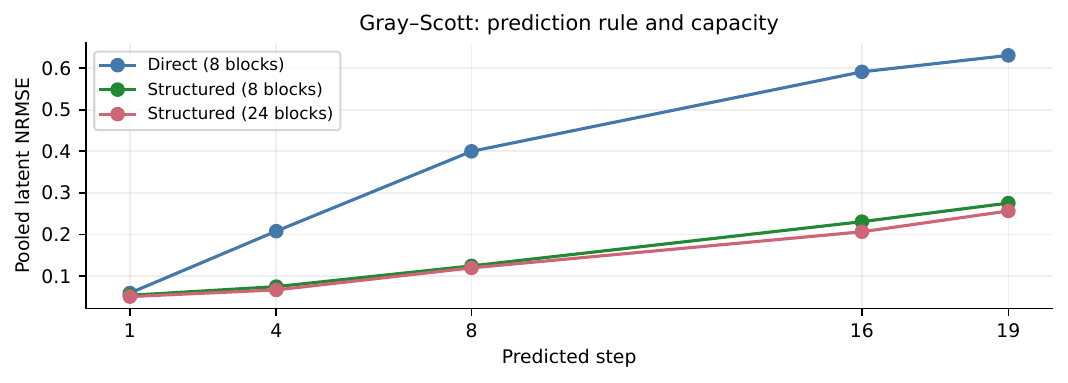}
\caption{\textbf{Gray--Scott latent prediction on the complete ID-validation
set.} The direct and structured eight-block predictors have equal
parameter counts and training budgets. The twenty-four-block model
tests additional capacity within the structured family.}
\label{fig:ext_gray_scott_horizon}
\end{figure}

The structured predictor substantially improves multi-step accuracy
despite nearly identical teacher errors
(Table~\ref{tab:ext_gray_scott}; Figure~\ref{fig:ext_gray_scott_horizon}).
Relative to the direct predictor, the 8-block structured model reduces
the final-step pooled latent error from 0.630593 to 0.275269 and the
whole-rollout error from 0.457266 to 0.171373, while maintaining
essentially the same teacher error (0.041156 vs.\ 0.041170).
The improvement is already visible at intermediate horizons, with
$h_8$ decreasing from 0.399747 to 0.124119. These results show that
the structured prediction rule provides a clear ID-validation benefit
for Gray--Scott, complementing the coarse Advection result and
indicating that the effect of predictor structure depends on the
equation and its sampled dynamics.

Increasing the structured predictor from eight to twenty-four blocks
provides a further, though more modest, improvement under the same
optimizer-step budget. The final-step error decreases from 0.275269
to 0.253280, while the whole-rollout error decreases from 0.171373
to 0.162613; the earlier-horizon errors also improve from 0.053405
to 0.051337 at $h_1$ and from 0.124119 to 0.119414 at $h_8$.
Thus, additional capacity consistently benefits the structured
Gray--Scott predictor, in line with the trend observed on Vorticity.

\begin{table}[htbp]
\centering\small
\setlength{\tabcolsep}{4pt}
\caption{\textbf{Gray--Scott prediction rule and capacity controls on ID validation.}}
\label{tab:ext_gray_scott}
\begin{adjustbox}{max width=\textwidth}
\begin{tabular}{lrrrrrr}
\toprule
Predictor & Params (M) & Teacher $\downarrow$ & $h_1$ $\downarrow$& $h_8$ $\downarrow$& $h_{19}$ $\downarrow$& Whole $\downarrow$\\
\midrule
Direct (8 blocks) & 14.42 & 0.041156 & 0.058798 & 0.399747 & 0.630593 & 0.457266 \\
Structured (8 blocks) & 14.42 & 0.041170 & 0.053405 & 0.124119 & 0.275269 & 0.171373 \\
Structured (24 blocks) & 42.81 & 0.039671 & 0.051337 & 0.119414 & 0.25328 & 0.162613 \\
\bottomrule
\end{tabular}
\end{adjustbox}
\end{table}

\paragraph{Physical-tangent supervision}
\label{sec:predabl_temporal_sampling}
We next examine explicit factorization and physical-tangent supervision. Given a physical state $u$ with instantaneous PDE tangent
$\dot u$, we obtain its corresponding latent tangent through the
Jacobian-vector product (JVP)
\[
\dot q_{\rm phys}
=
J_{E}(u)\dot u
=
D_uE(u)\,\dot u,
\]
where $E$ is the frozen encoder. This computes the directional
derivative of the representation along the physical evolution without
forming the full encoder Jacobian. We compare a conditional generator
M0, a strictly factorized generator M1, and the same factorized
generator with JVP supervision M2,
\[
\mathrm{M0}:\quad
\dot q = F_\theta(q,\beta/s_\beta),
\qquad
\mathrm{M1,M2}:\quad
\dot q = (\beta/s_\beta)A_\theta(q).
\]
M2 additionally penalizes the discrepancy between the predicted latent
tangent and $\dot q_{\rm phys}$ with weight $0.1$. All models share the
frozen representation and decoder, use RK4 with four substeps, and are
trained for 12,200 optimizer steps; M1 and M2 additionally share the
same initialization.

As shown in Table~\ref{tab:predabl_advection_raw_interval}, explicit
factorization substantially improves prediction: from M0 to M1, the
raw whole-rollout latent error decreases from $0.1100$ to $0.0848$,
the selected latent error from $0.1091$ to $0.0821$, and the field
RelL2 from $0.1009$ to $0.0769$, with corresponding reductions in
phase and speed error. Adding JVP supervision further improves the
aggregate trajectory metrics, yielding the lowest teacher error
($0.0034$), raw whole-rollout latent error ($0.0200$), selected latent
error ($0.0798$), and field RelL2 ($0.0737$). M1 nevertheless retains
lower phase and speed errors than M2, suggesting that the tangent
constraint primarily improves aggregate rollout accuracy rather than
every dynamical diagnostic. Overall, the matched comparison supports
both explicit factorization and physical-tangent supervision at a
temporal resolution that better resolves the local dynamics.

\begin{table}[htbp]
\centering\small
\setlength{\tabcolsep}{4pt}
\caption{\textbf{Ablation of factorization and physical-tangent supervision on raw-interval Advection.}
All models are trained and evaluated under the same raw-interval protocol; M1 isolates the effect of explicit factorization over the conditional baseline M0, while M2 further adds physical-tangent JVP supervision.}
\label{tab:predabl_advection_raw_interval}
\begin{adjustbox}{max width=\textwidth}
\begin{tabular}{lrrrrrr}
\toprule
Model & Teacher $\downarrow$ & Raw latent whole $\downarrow$& Selected latent whole $\downarrow$& Field RelL2 $\downarrow$& Phase & Speed \\
\midrule
M0: conditional & 0.0039 & 0.1100 & 0.1091 & 0.1009 & 2.4208 & 0.2128 \\
M1: factorized & 0.0035 & 0.0848 & 0.0821 & 0.0769 & 1.1478 & 0.1018 \\
M2: factorized + JVP & 0.0034 & 0.0200 & 0.0798 & 0.0737 & 1.6361 & 0.1189 \\
\bottomrule
\end{tabular}
\end{adjustbox}
\end{table}

\paragraph{Integration Strategies.} We further examine whether the performance of the structured predictor
depends strongly on the numerical integrator. As shown in
Table~\ref{tab:predabl_precision_solver}, low-order integration noticeably
degrades prediction accuracy. Using Euler increases the final-step and
whole-rollout errors to $0.1914$ and $0.0936$, respectively, whereas
RK2 reduces them to $0.1648$ and $0.0796$. Moving from RK2 to RK4
provides a further improvement, reaching $0.159669$ final-step error
and $0.077184$ whole-rollout error.

\begin{table}[htbp]
\centering\small
\setlength{\tabcolsep}{4pt}
\caption{\textbf{Effect of numerical integration accuracy on Vorticity OOD prediction.}
Higher-order integration substantially improves over Euler, while
forecasting accuracy largely saturates by RK4.}
\label{tab:predabl_precision_solver}
\begin{adjustbox}{max width=\textwidth}
\begin{tabular}{lrrrrr}
\toprule
Predictor & Integral & Late teacher & Final & Whole & Round trip \\
\midrule
Physics-D & Euler        & 0.0518   & 0.1914   & 0.0936   & 1.1243 \\
Physics-D & RK2          & 0.0462   & 0.1648   & 0.0796   & 0.8465 \\
Physics-D & RK4          & 0.045295 & 0.159669 & 0.077184 & 0.781600 \\
Physics-D & RK8          & 0.045275 & 0.159705 & 0.077226 & 0.670196 \\
\bottomrule
\end{tabular}
\end{adjustbox}
\end{table}

Beyond RK4, however, the forward-prediction metrics essentially
saturate. RK8 yields nearly identical final and whole-rollout errors
($0.159705$ and $0.077226$), despite further reducing the round-trip
error from $0.781600$ to $0.670196$. This distinction suggests that
higher-order integration continues to improve numerical reversibility,
while RK4 already provides sufficient precision for forward forecasting.
Therefore, the predictive advantage of the structured model is unlikely
to arise simply from using an increasingly accurate numerical solver;
rather, the learned continuous dynamics account for most of the
forecasting gain once integration error is sufficiently controlled.

\subsection{Trajectory Visualizations}
\label{app:tra_vis}

\subsubsection{Advection}

\begin{figure}[h]
    \centering

    \begin{subfigure}{\textwidth}
        \centering
        \includegraphics[width=\linewidth]{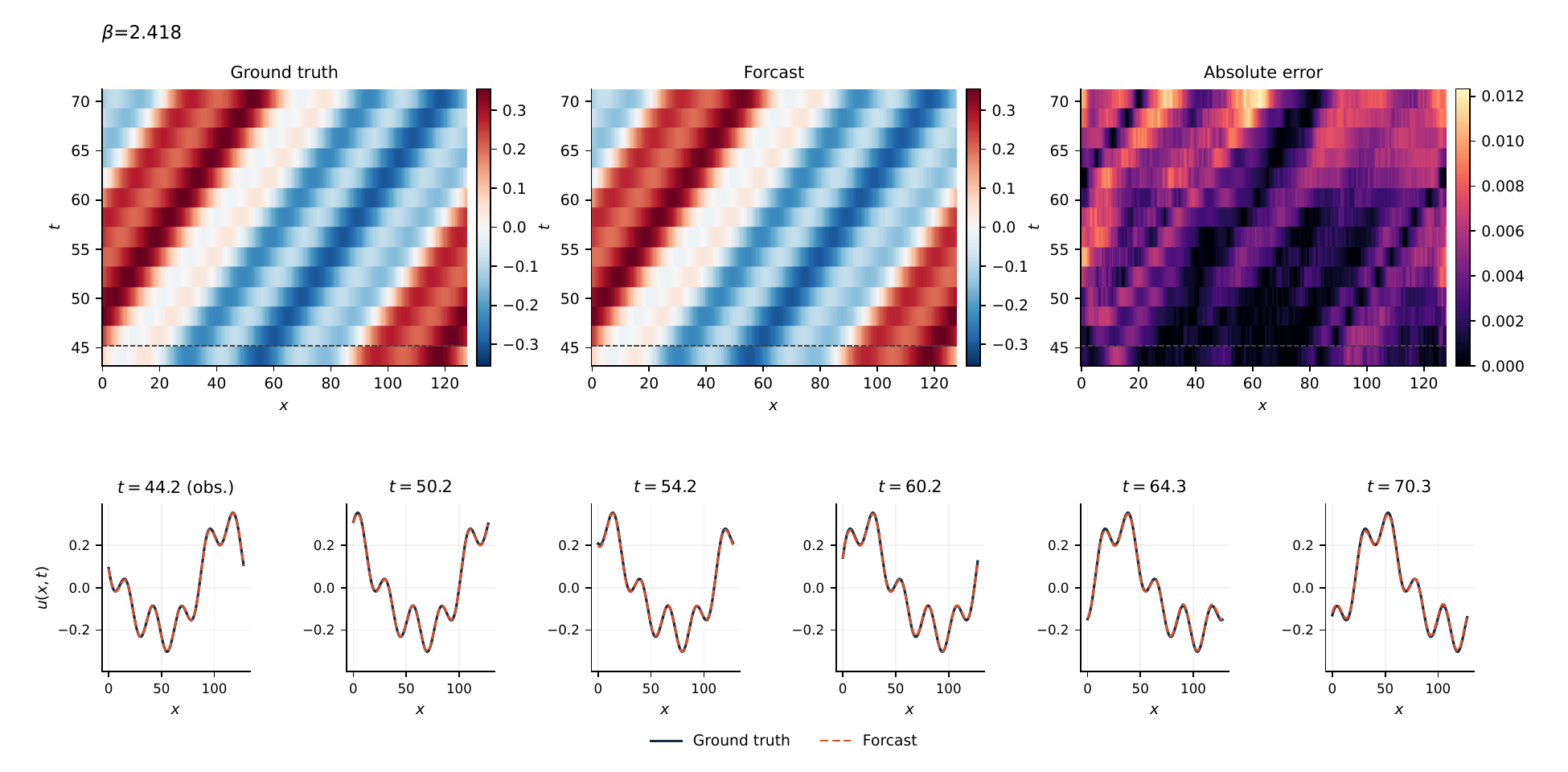}
        \caption{$\beta=2.418$}
        \label{fig:ad_vis_00000}
    \end{subfigure}

    \begin{subfigure}{\textwidth}
        \centering
        \includegraphics[width=\linewidth]{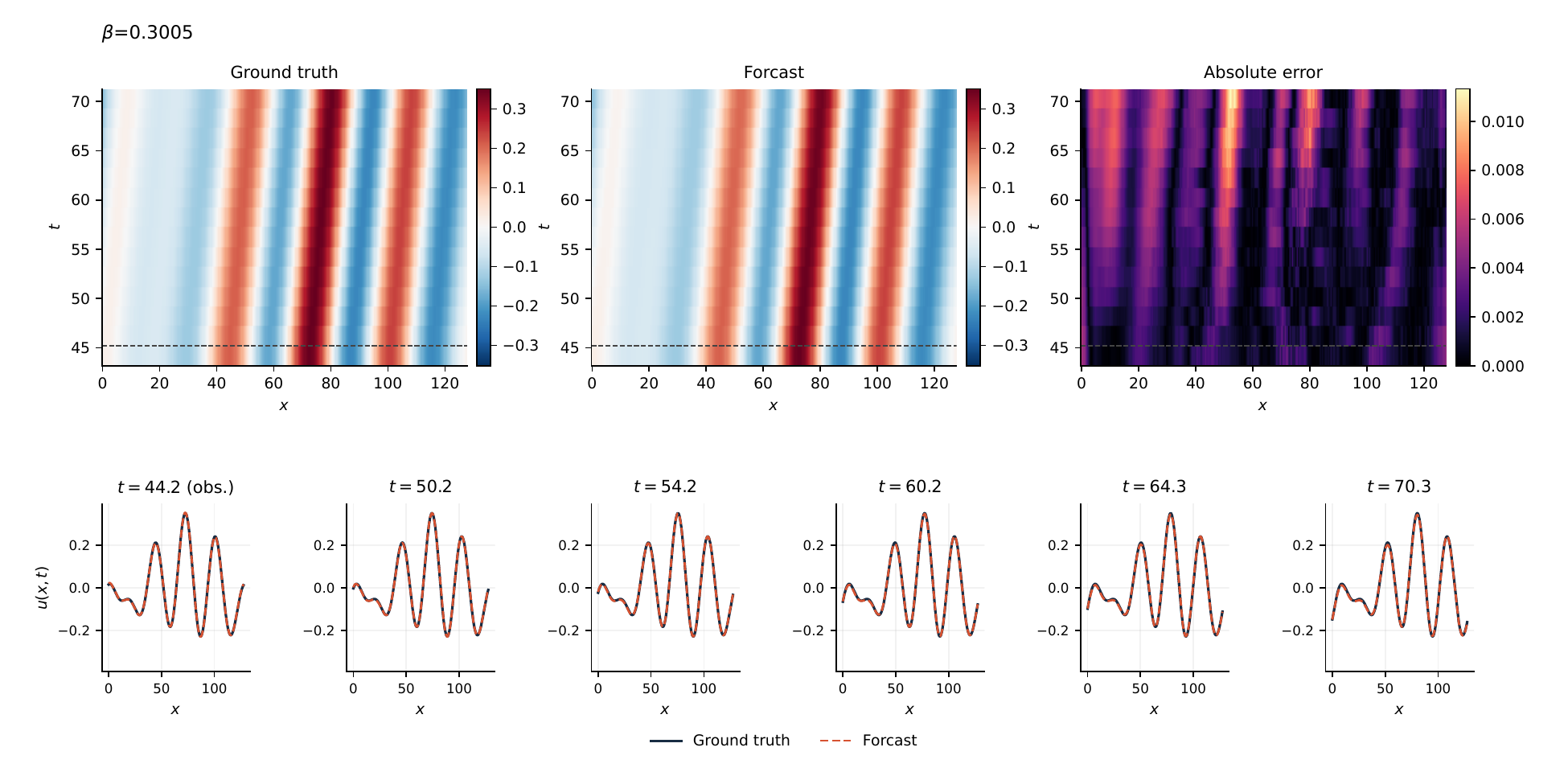}
        \caption{$\beta=0.3005$}
        \label{fig:ad_vis_00030}
    \end{subfigure}

    \caption{\textbf{Qualitative Advection forecasts under different governing parameters.}}
    \label{fig:advection_visualization}
\end{figure}

Figure~\ref{fig:advection_visualization} visualizes Advection trajectories for two different values of the governing parameter $\beta$, showing the ground truth, forecast, absolute error, and representative spatial profiles over time.

\subsubsection{Burgers}

\begin{figure}[h]
    \centering

    \begin{subfigure}{\textwidth}
        \centering
        \includegraphics[width=\linewidth]{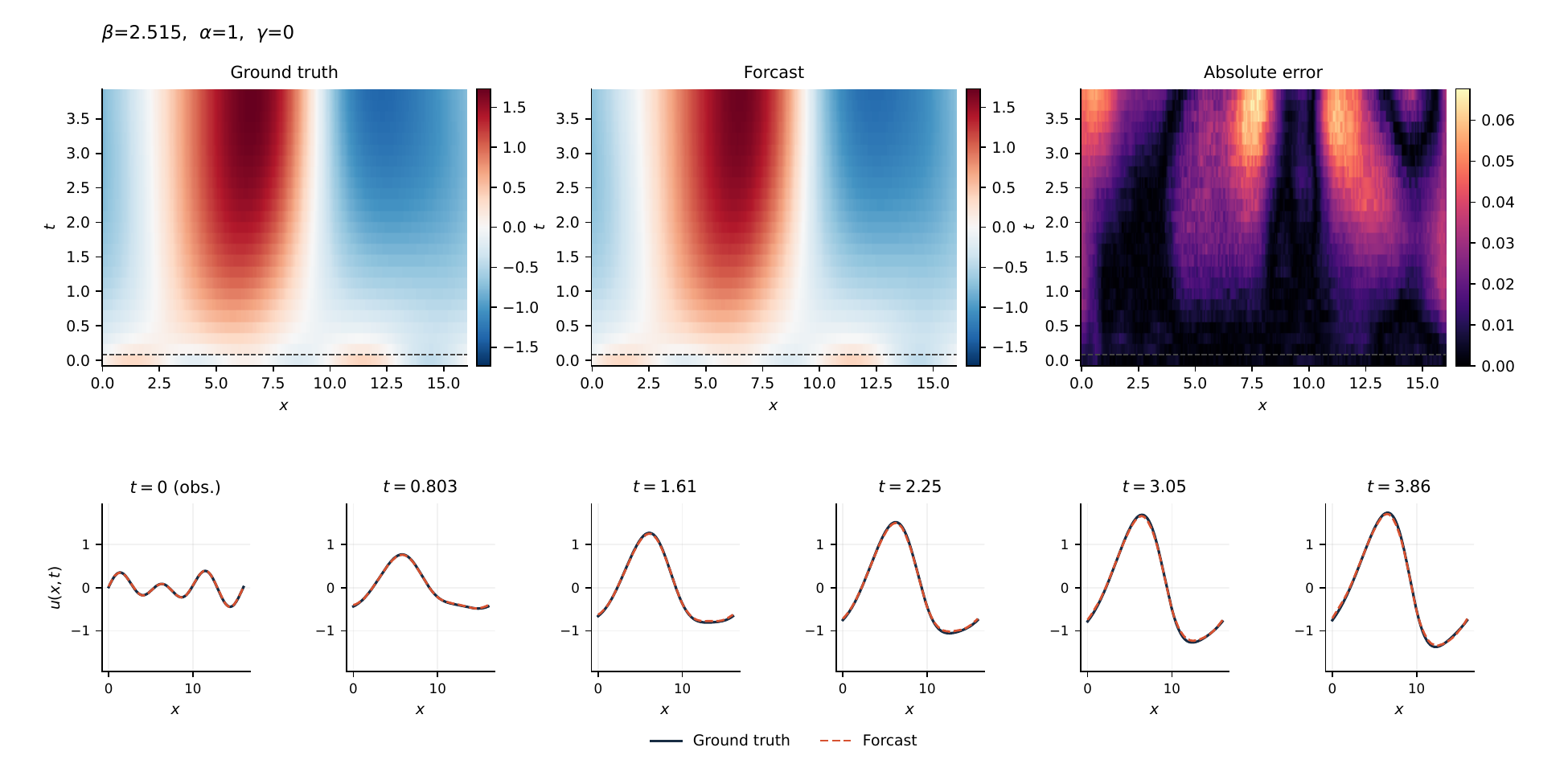}
        \caption{$\beta=2.515,\ \alpha=1,\ \gamma=0$}
        \label{fig:bg_vis_00000}
    \end{subfigure}

    \begin{subfigure}{\textwidth}
        \centering
        \includegraphics[width=\linewidth]{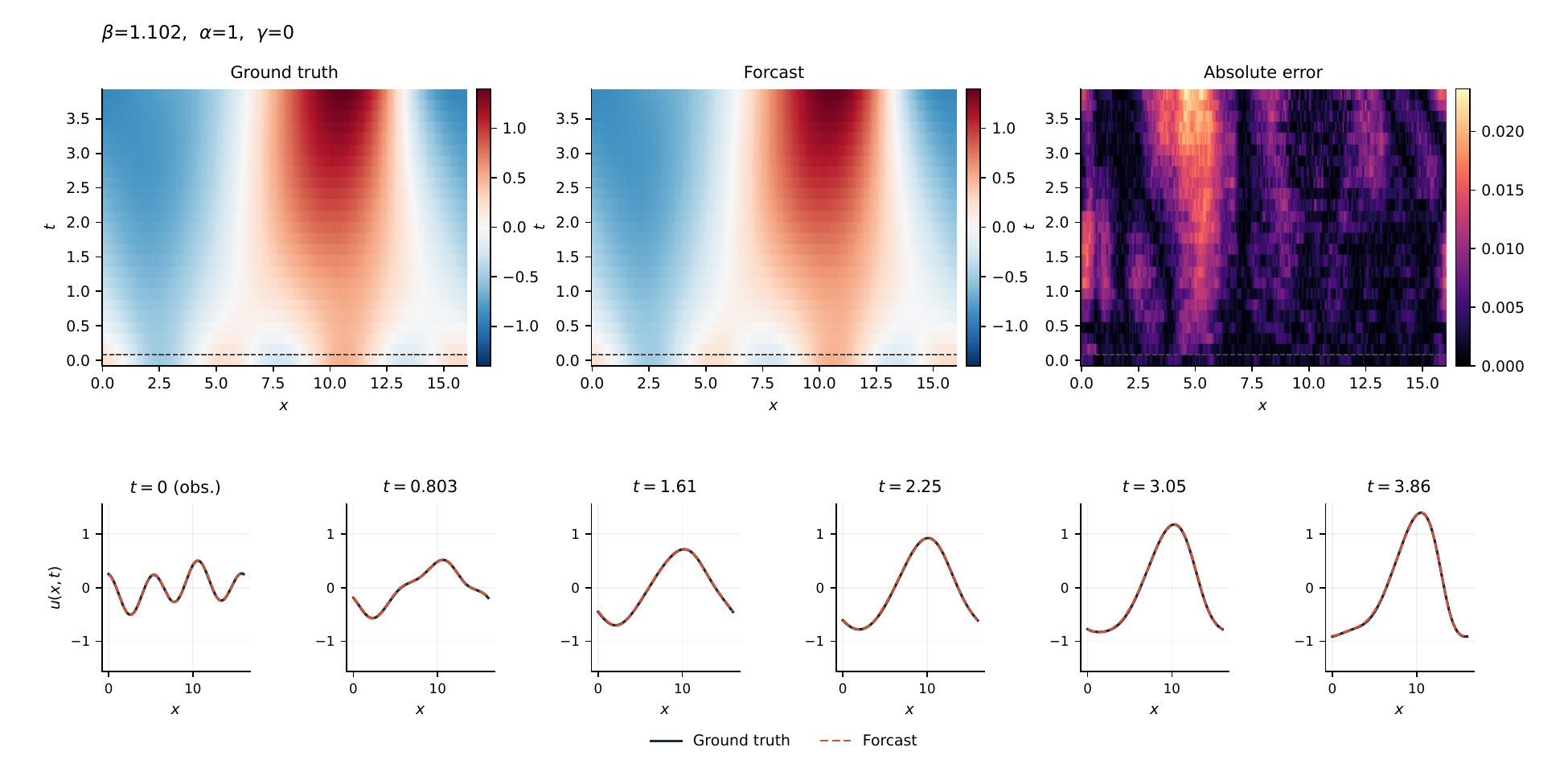}
        \caption{$\beta=1.102,\ \alpha=1,\ \gamma=0$}
        \label{fig:bg_vis_00060}
    \end{subfigure}

    \caption{\textbf{Qualitative Burgers forecasts under different governing parameters.}}
    \label{fig:burgers_visualization}
\end{figure}

Figure~\ref{fig:burgers_visualization} visualizes Burgers trajectories for two different values of the governing parameter $\beta$ with fixed $\alpha=1$ and $\gamma=0$, showing the ground truth, forecast, absolute error, and representative spatial profiles over time.

\subsubsection{Heat}

\begin{figure}[h]
    \centering

    \begin{subfigure}{\textwidth}
        \centering
        \includegraphics[width=\linewidth]{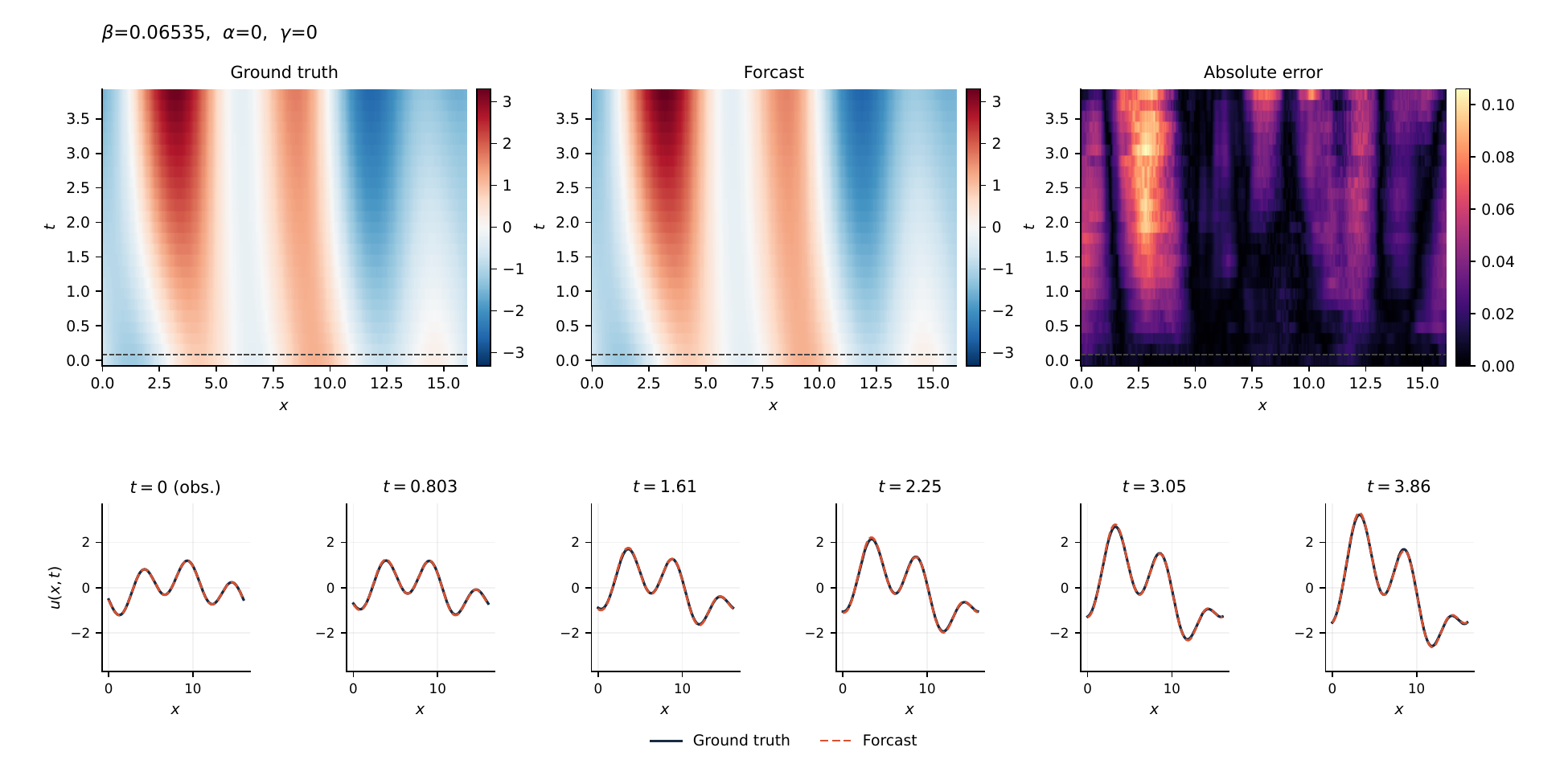}
        \caption{$\beta=0.06535,\ \alpha=0,\ \gamma=0$}
        \label{fig:heat_vis_00060}
    \end{subfigure}

    \begin{subfigure}{\textwidth}
        \centering
        \includegraphics[width=\linewidth]{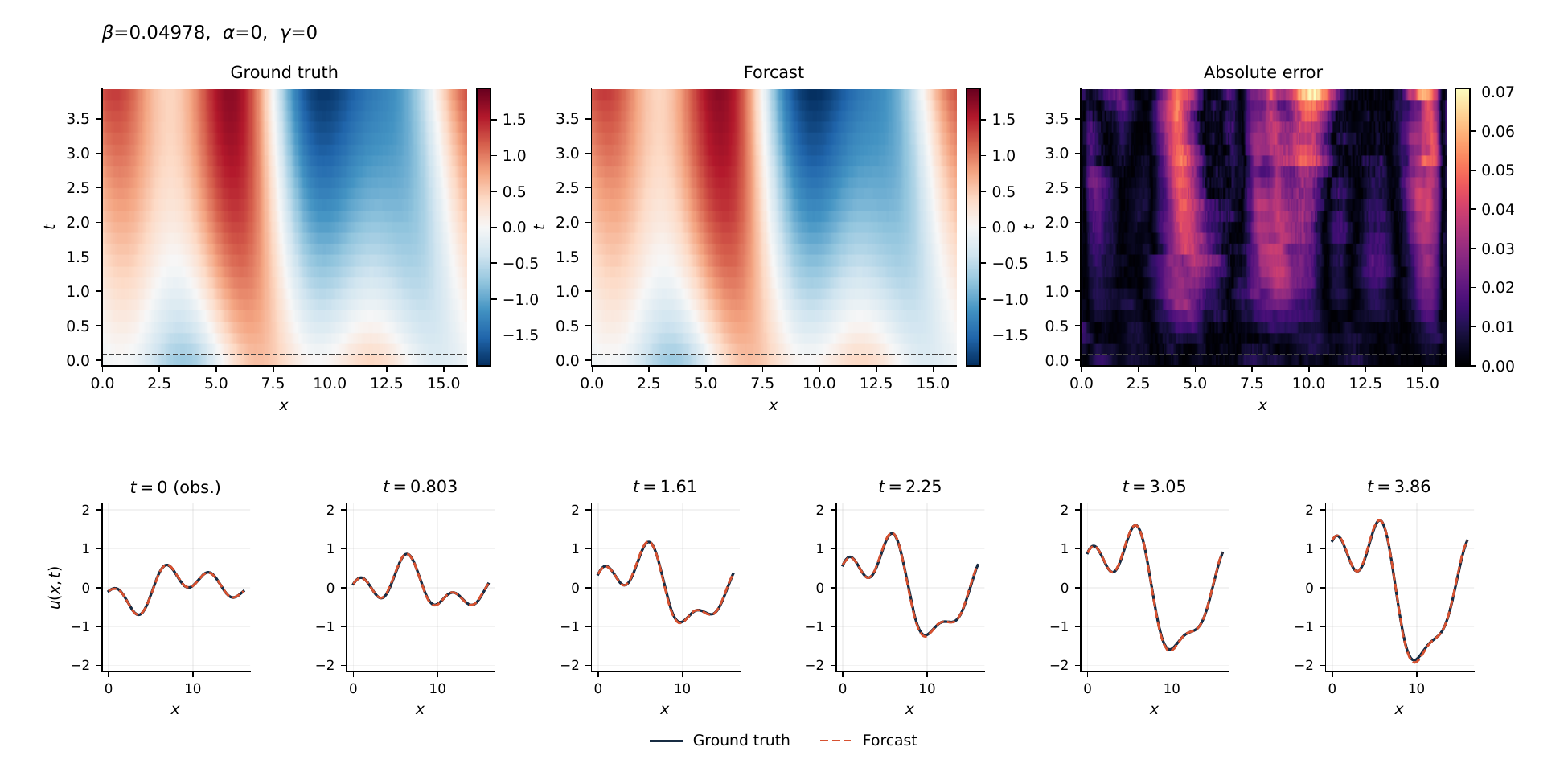}
        \caption{$\beta=0.04978,\ \alpha=0,\ \gamma=0$}
        \label{fig:heat_vis_00010}
    \end{subfigure}

    \caption{\textbf{Qualitative Heat forecasts under different governing parameters.}}
    \label{fig:heat_visualization}
\end{figure}

Figure~\ref{fig:heat_visualization} visualizes Heat trajectories for two different values of the governing parameter $\beta$ with fixed $\alpha=0$ and $\gamma=0$, showing the ground truth, forecast, absolute error, and representative spatial profiles over time.

\subsubsection{Wave-B}

\begin{figure}[h]
    \centering

    \begin{subfigure}{\textwidth}
        \centering
        \includegraphics[width=\linewidth]{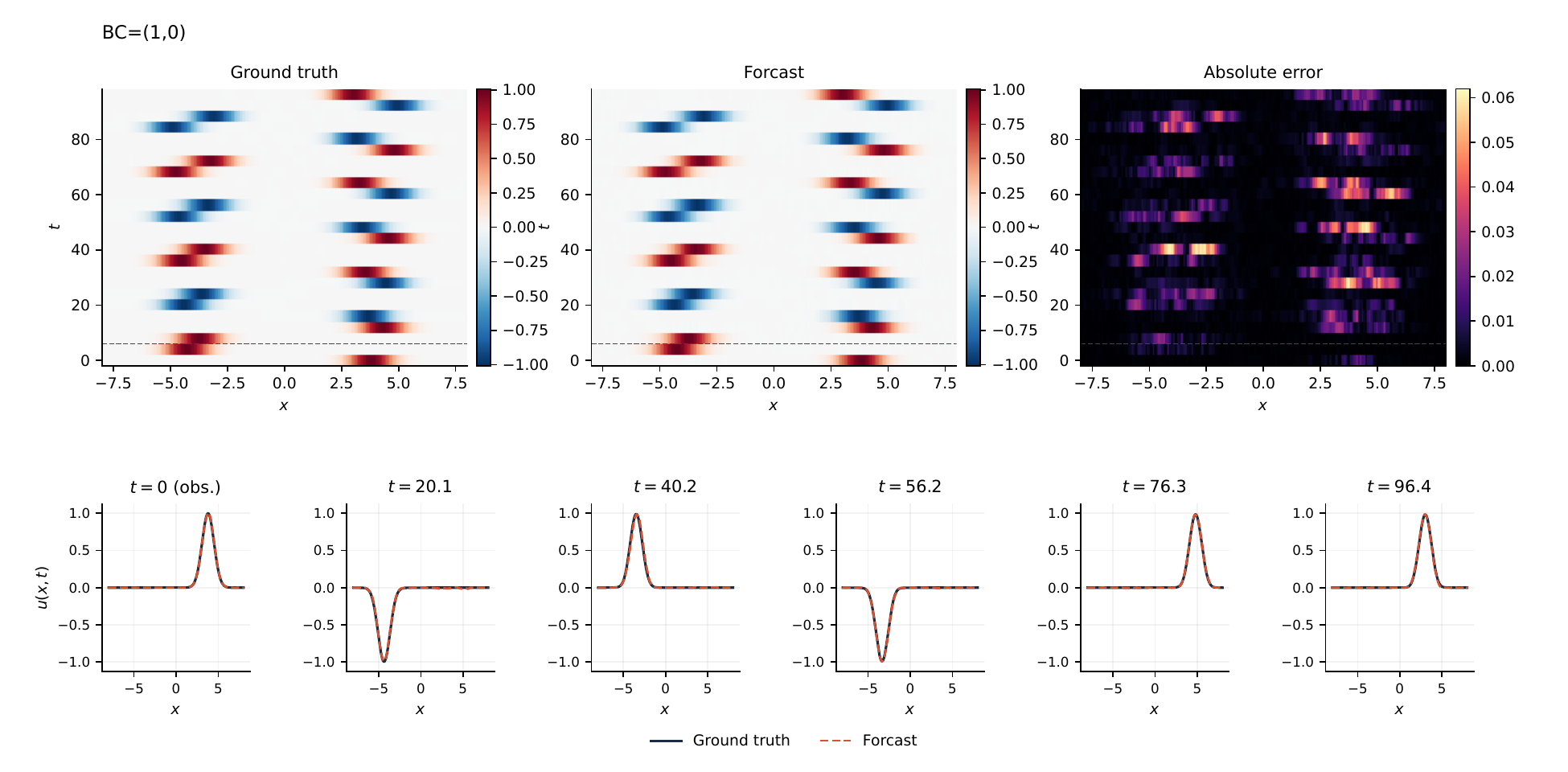}
        \caption{$\mathrm{BC}=(1,0)$}
        \label{fig:waveb_vis_00075}
    \end{subfigure}

    \begin{subfigure}{\textwidth}
        \centering
        \includegraphics[width=\linewidth]{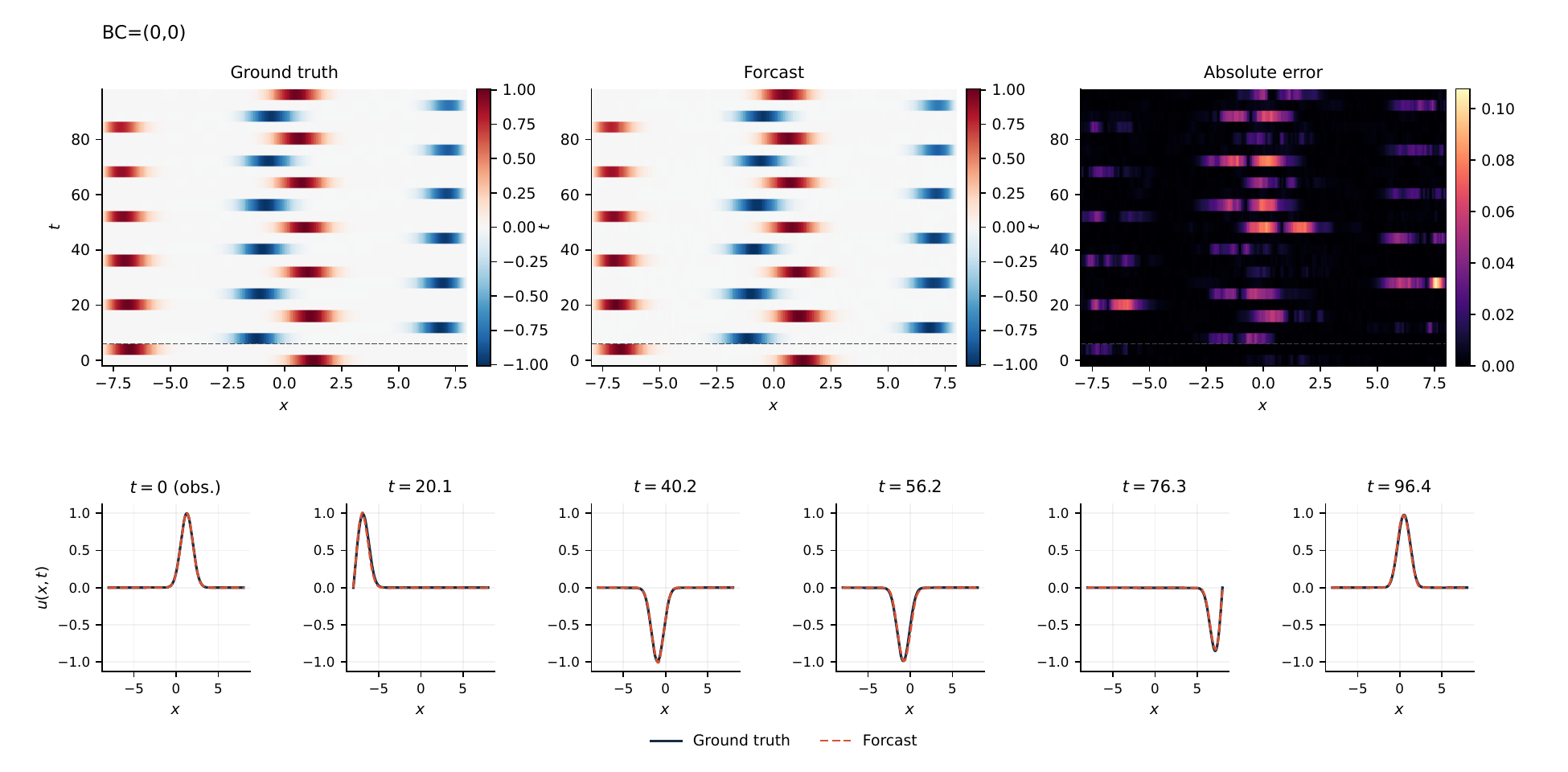}
        \caption{$\mathrm{BC}=(0,0)$}
        \label{fig:waveb_vis_00000}
    \end{subfigure}

    \caption{\textbf{Qualitative Wave-B forecasts under different boundary conditions.}}
    \label{fig:waveb_visualization}
\end{figure}

Figure~\ref{fig:waveb_visualization} visualizes Wave-B trajectories under two different boundary conditions, showing the ground truth, forecast, absolute error, and representative spatial profiles over time.

\subsubsection{Combined Equation}

\begin{figure}[h]
    \centering

    \begin{subfigure}{\textwidth}
        \centering
        \includegraphics[width=\linewidth]{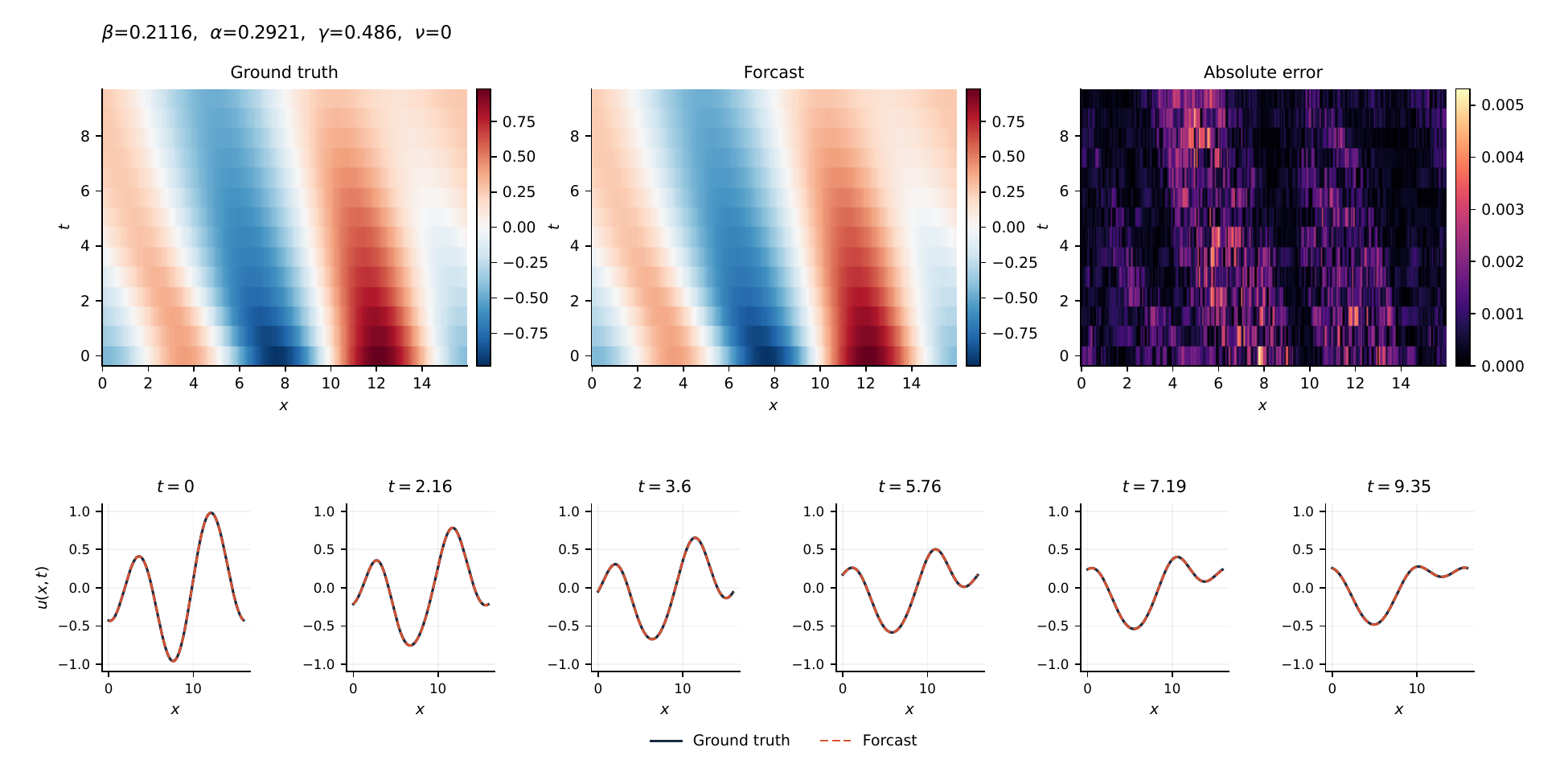}
        \caption{ID: $\beta=0.2116,\ \alpha=0.2921,\ \gamma=0.486,\ \nu=0$}
        \label{fig:combine_id_vis_00080}
    \end{subfigure}

    \begin{subfigure}{\textwidth}
        \centering
        \includegraphics[width=\linewidth]{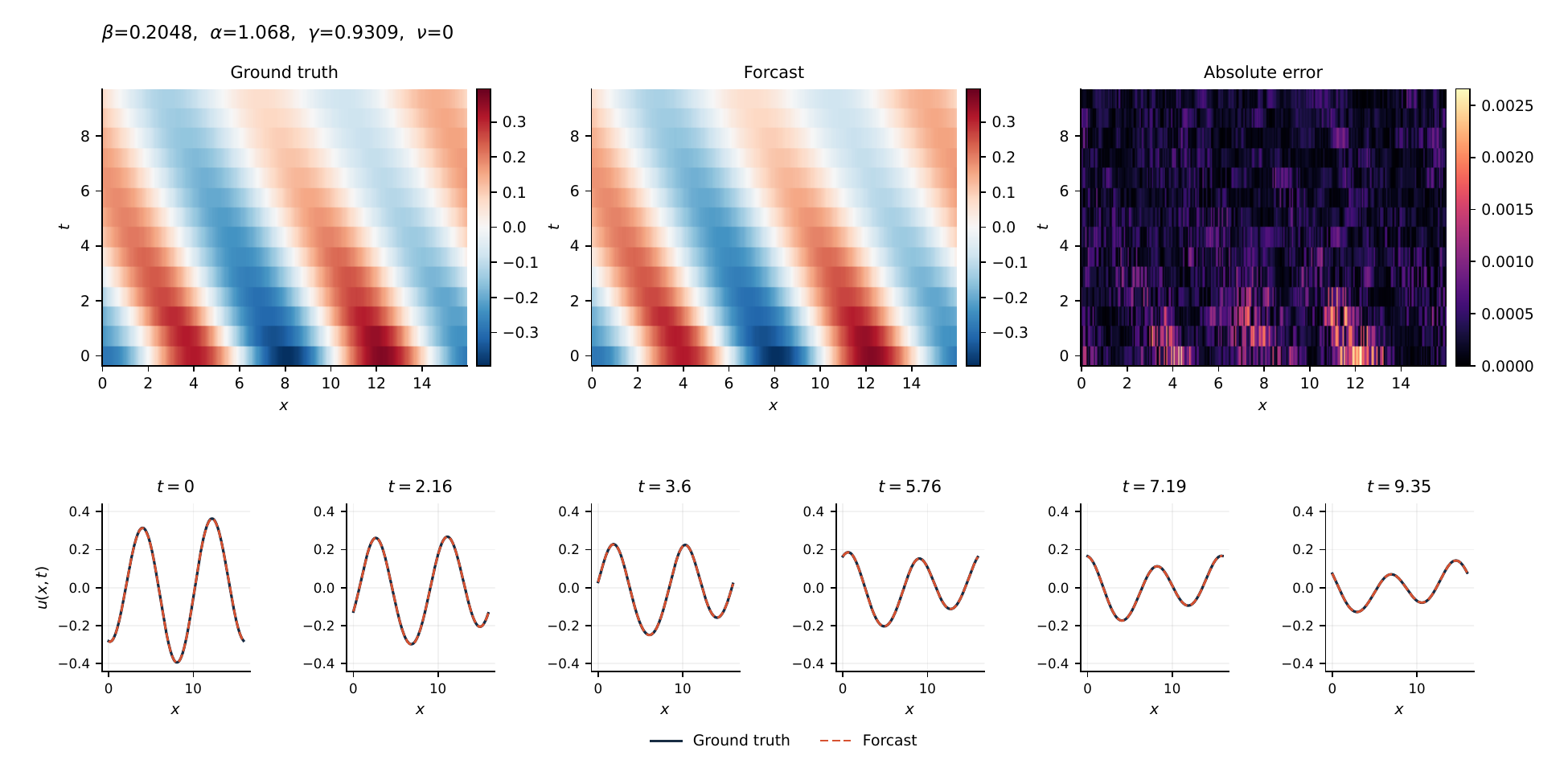}
        \caption{OOD: $\beta=0.2048,\ \alpha=1.068,\ \gamma=0.9309,\ \nu=0$}
        \label{fig:combine_ood_vis_00080}
    \end{subfigure}

    \caption{\textbf{Qualitative Combined Equation forecasts under ID and OOD governing parameters.}}
    \label{fig:combine_visualization}
\end{figure}

Figure~\ref{fig:combine_visualization} visualizes Combined Equation trajectories under representative ID and OOD governing parameters, showing the ground truth, forecast, absolute error, and representative spatial profiles over time.

\subsubsection{Wave-2D}

\begin{figure}[h]
    \centering

    \begin{subfigure}{\textwidth}
        \centering
        \includegraphics[width=\linewidth]{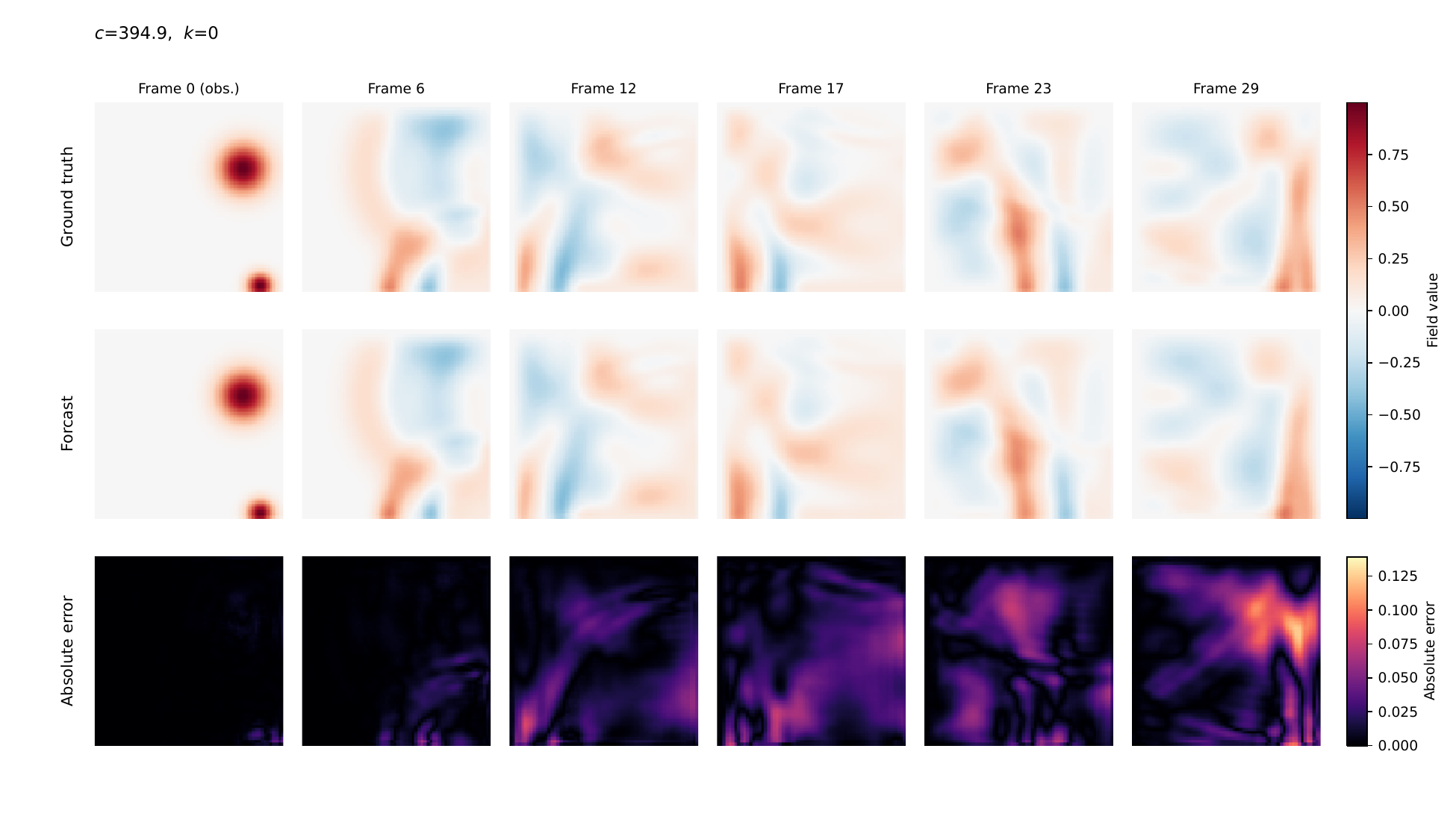}
        \caption{ID: $c=394.9,\ k=0$}
        \label{fig:wave2d_id_vis_00050}
    \end{subfigure}

    \begin{subfigure}{\textwidth}
        \centering
        \includegraphics[width=\linewidth]{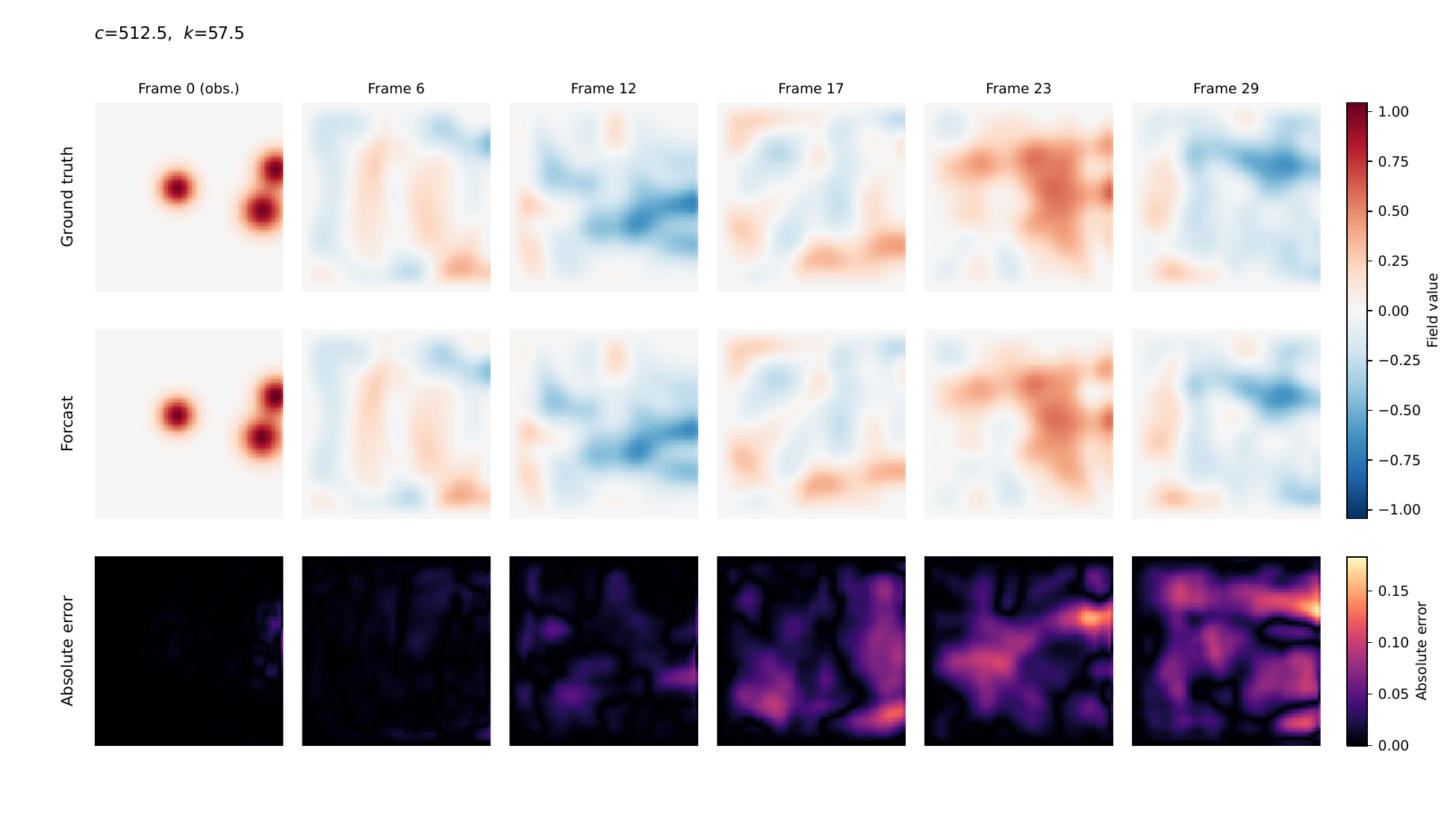}
        \caption{OOD: $c=512.5,\ k=57.5$}
        \label{fig:wave2d_ood_vis_00000}
    \end{subfigure}

    \caption{\textbf{Qualitative Wave-2D forecasts under ID and OOD governing parameters.}}
    \label{fig:wave2d_visualization}
\end{figure}

Figure~\ref{fig:wave2d_visualization} visualizes Wave-2D trajectories under representative ID and OOD governing parameters $(c,k)$, showing the ground truth, forecast, and absolute error across representative rollout frames.

\subsubsection{Vorticity}

\begin{figure}[h]
    \centering

    \begin{subfigure}{\textwidth}
        \centering
        \includegraphics[width=\linewidth]{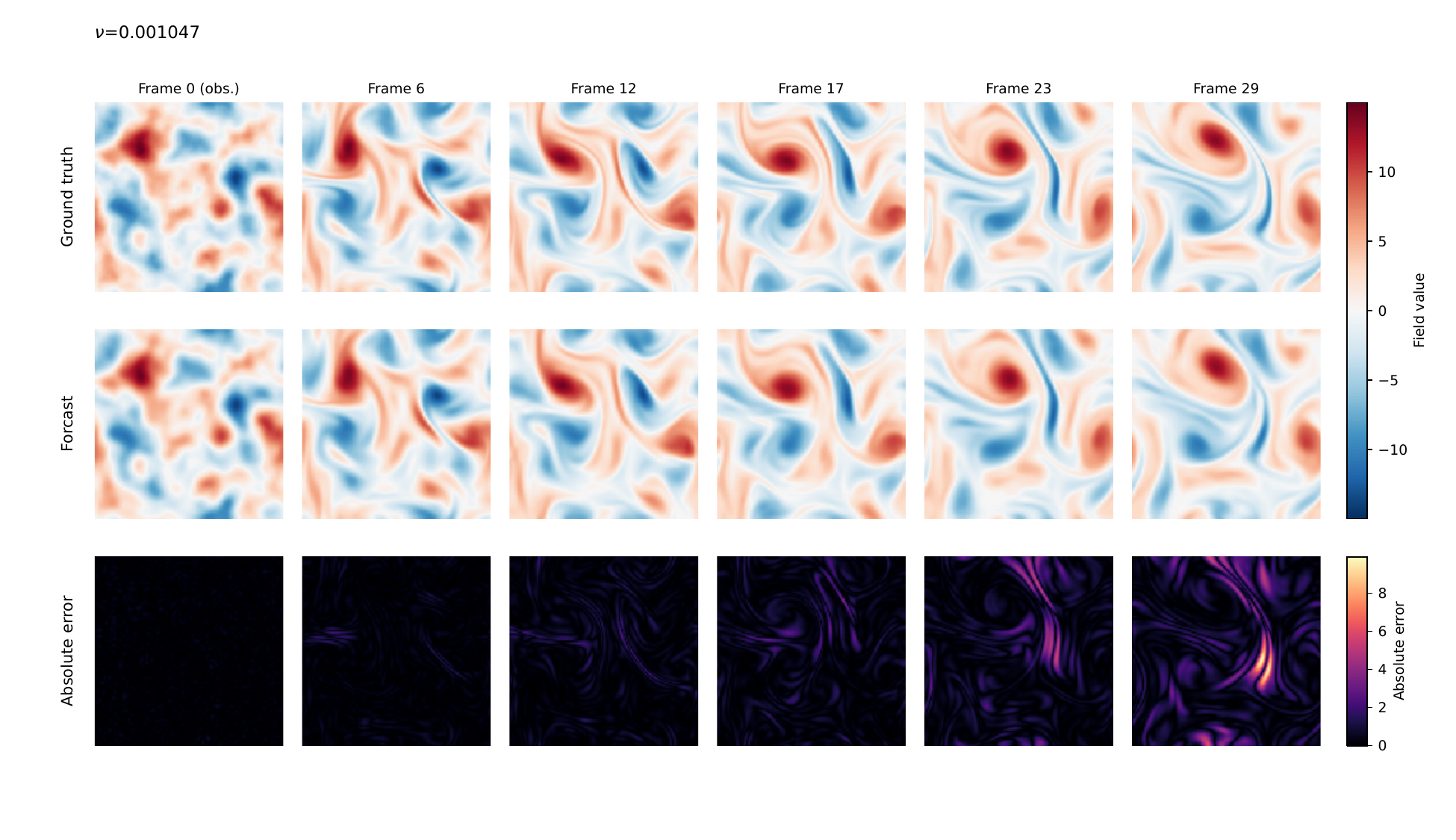}
        \caption{ID: $\nu=0.001047$}
        \label{fig:vorticity_id_vis_00200}
    \end{subfigure}

    \begin{subfigure}{\textwidth}
        \centering
        \includegraphics[width=\linewidth]{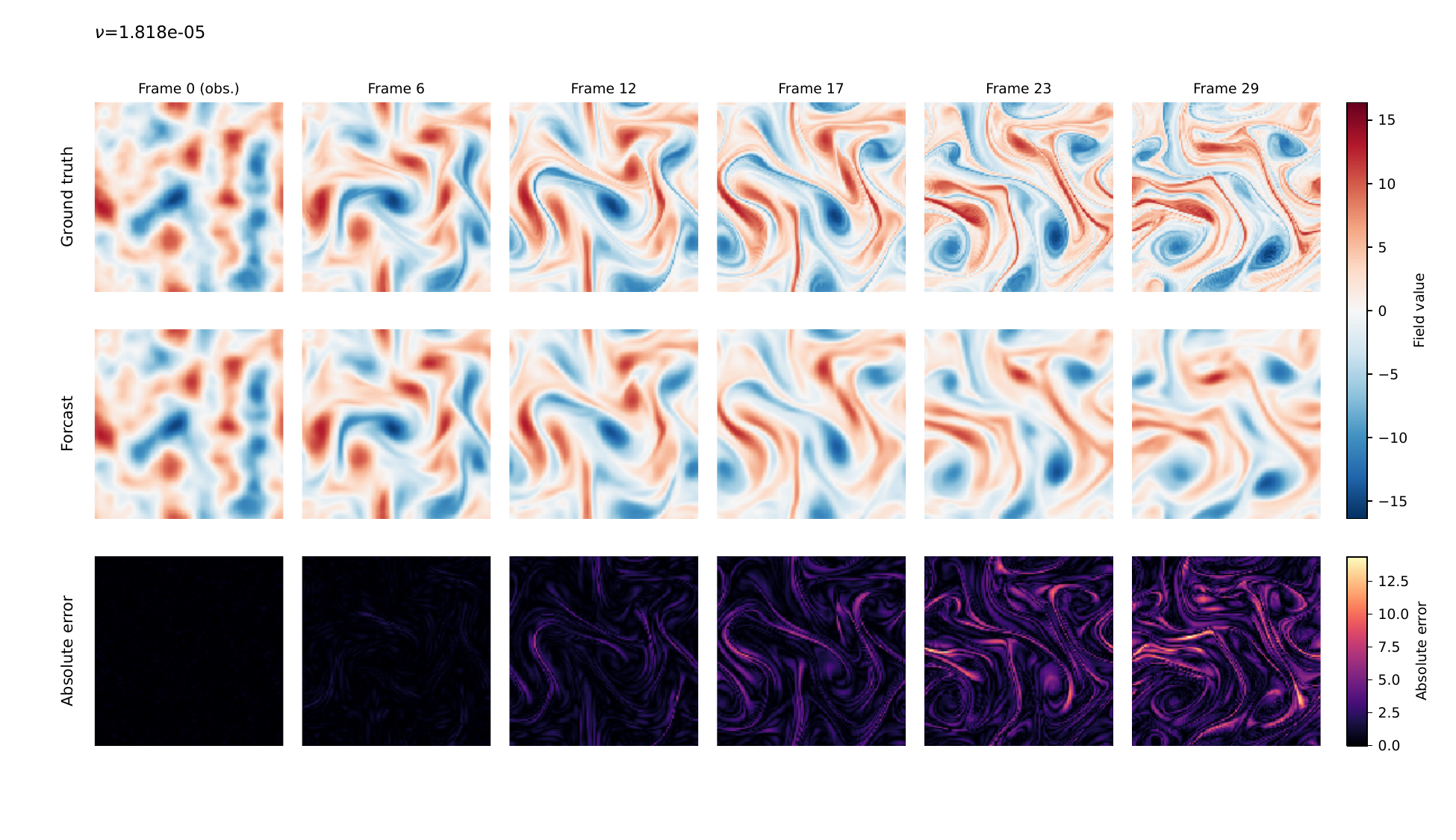}
        \caption{OOD: $\nu=1.818\times10^{-5}$}
        \label{fig:vorticity_ood_vis_00000}
    \end{subfigure}

    \caption{\textbf{Qualitative Vorticity forecasts under ID and OOD viscosity regimes.}}
    \label{fig:vorticity_visualization}
\end{figure}

Figure~\ref{fig:vorticity_visualization} visualizes Vorticity trajectories under representative ID and OOD viscosity values $\nu$, showing the ground truth, forecast, and absolute error across representative rollout frames.

\subsubsection{Gray--Scott}

\begin{figure}[h]
    \centering

    \begin{subfigure}{\textwidth}
        \centering
        \includegraphics[width=\linewidth]{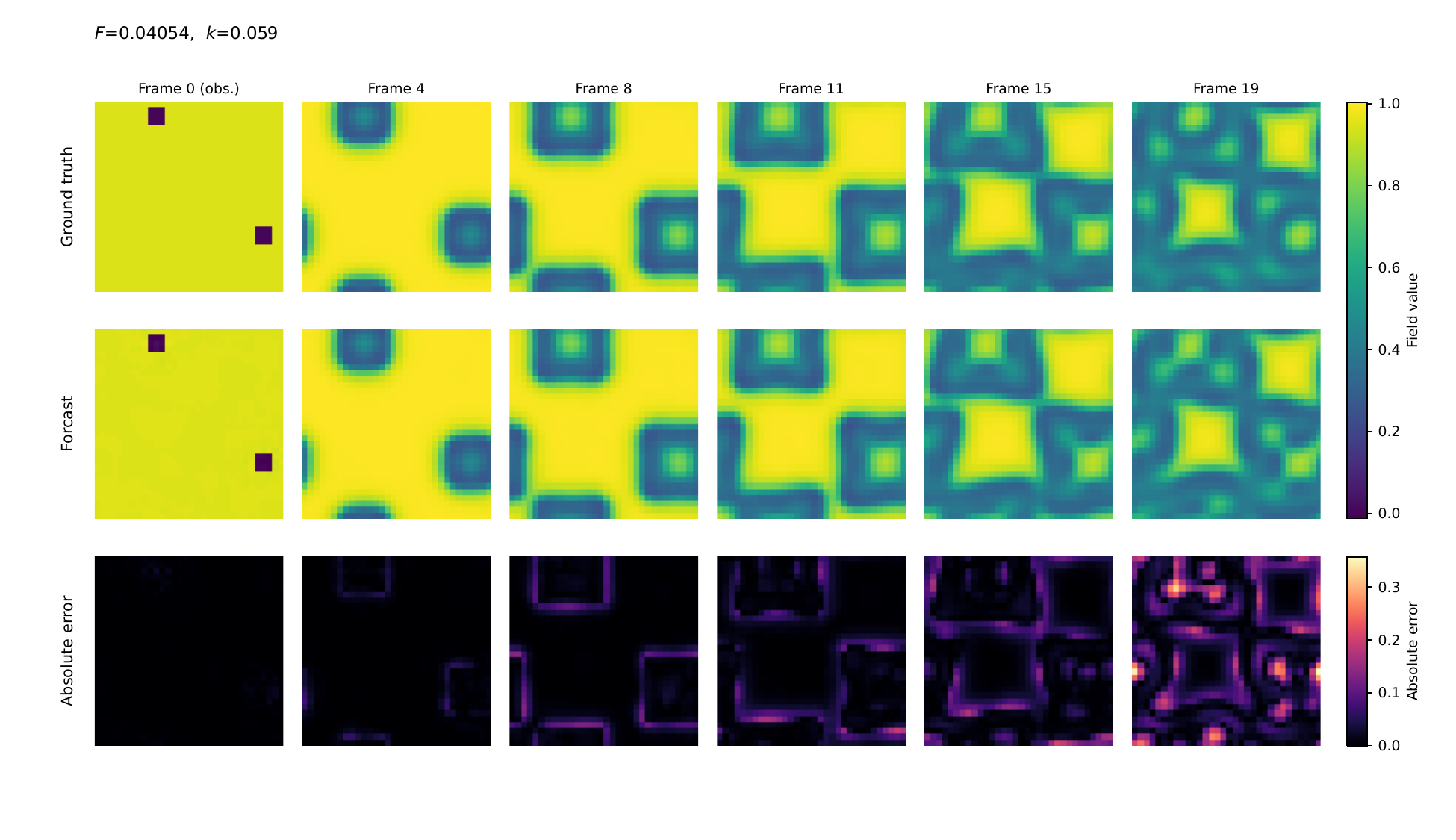}
        \caption{ID: $F=0.04054,\ k=0.059$}
        \label{fig:gs_id_vis_00050}
    \end{subfigure}

    \begin{subfigure}{\textwidth}
        \centering
        \includegraphics[width=\linewidth]{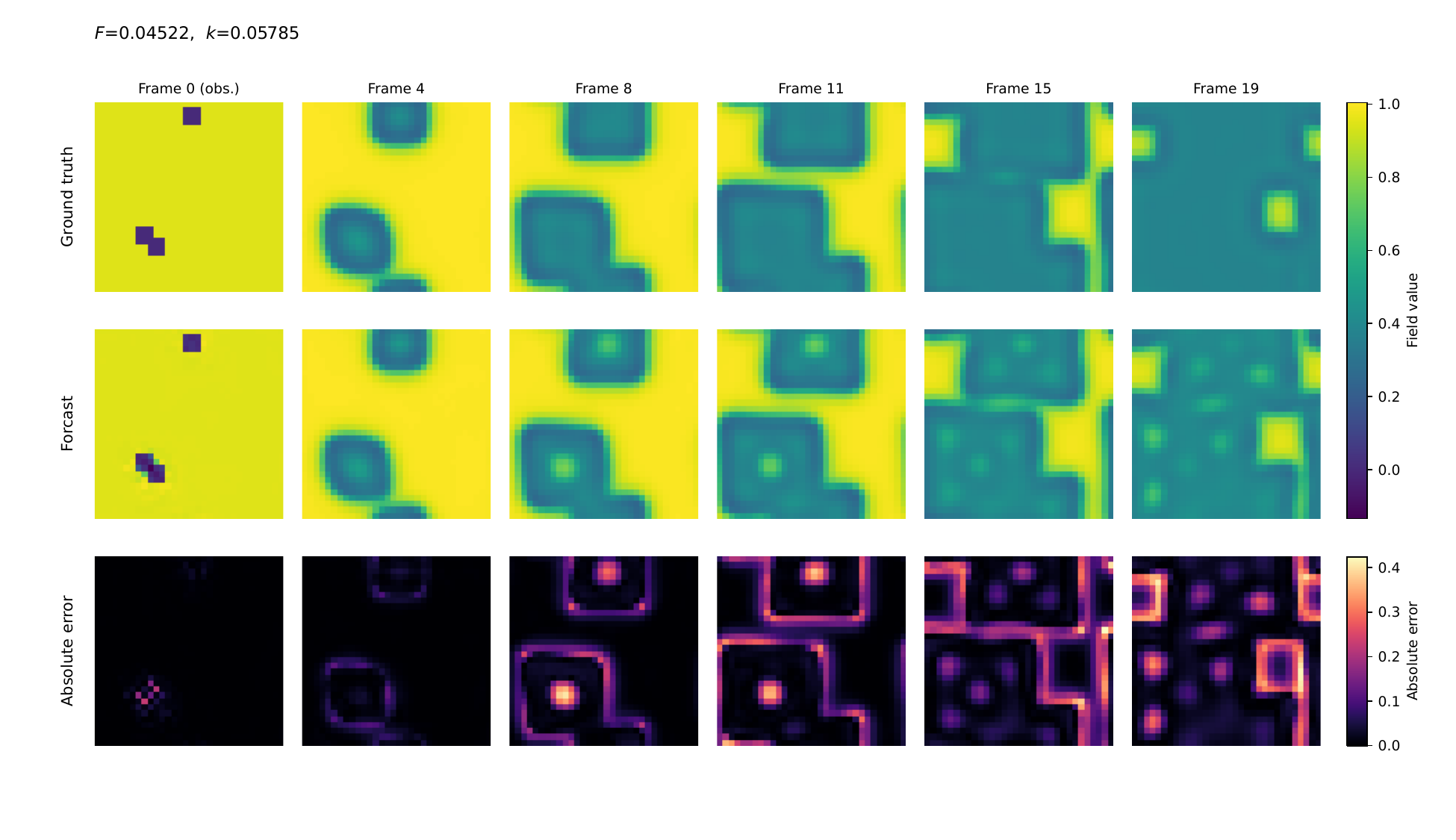}
        \caption{OOD: $F=0.04522,\ k=0.05785$}
        \label{fig:gs_ood_vis_00010}
    \end{subfigure}

    \caption{\textbf{Qualitative Gray--Scott forecasts under ID and OOD reaction parameters.}}
    \label{fig:gs_visualization}
\end{figure}

Figure~\ref{fig:gs_visualization} visualizes Gray--Scott trajectories under representative ID and OOD reaction parameters $(F,k)$, showing the ground truth, forecast, and absolute error across representative rollout frames.

\subsubsection{HeterNS}

\begin{figure}[p]
    \centering
    \captionsetup[subfigure]{font=small,skip=2pt}

    \begin{subfigure}{0.72\textwidth}
        \centering
        \includegraphics[width=\linewidth]{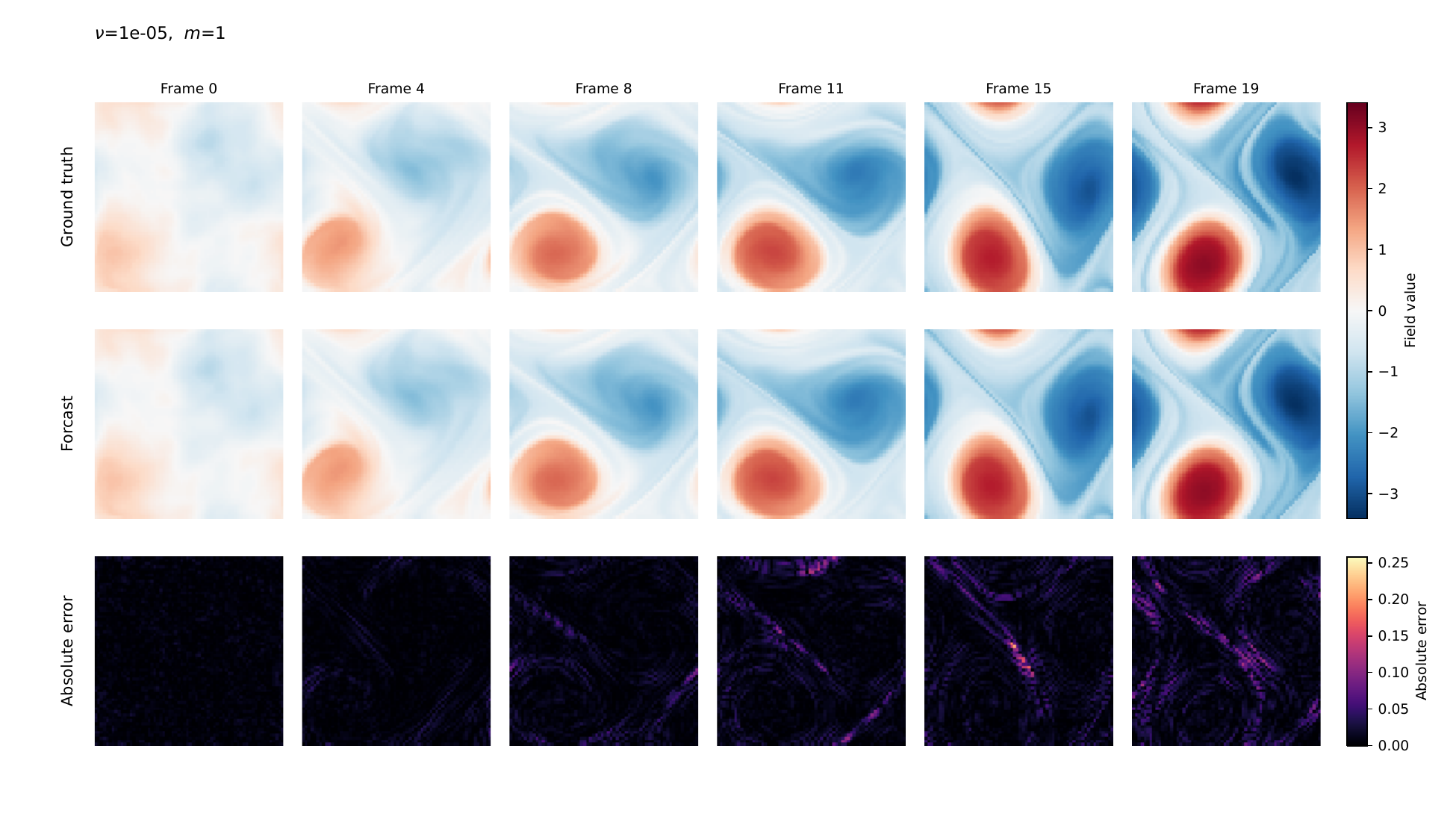}
        \caption{ID: $\nu=10^{-5},\ m=1$}
        \label{fig:heterns_id_vis}
    \end{subfigure}

    \vspace{1mm}

    \begin{subfigure}{0.72\textwidth}
        \centering
        \includegraphics[width=\linewidth]{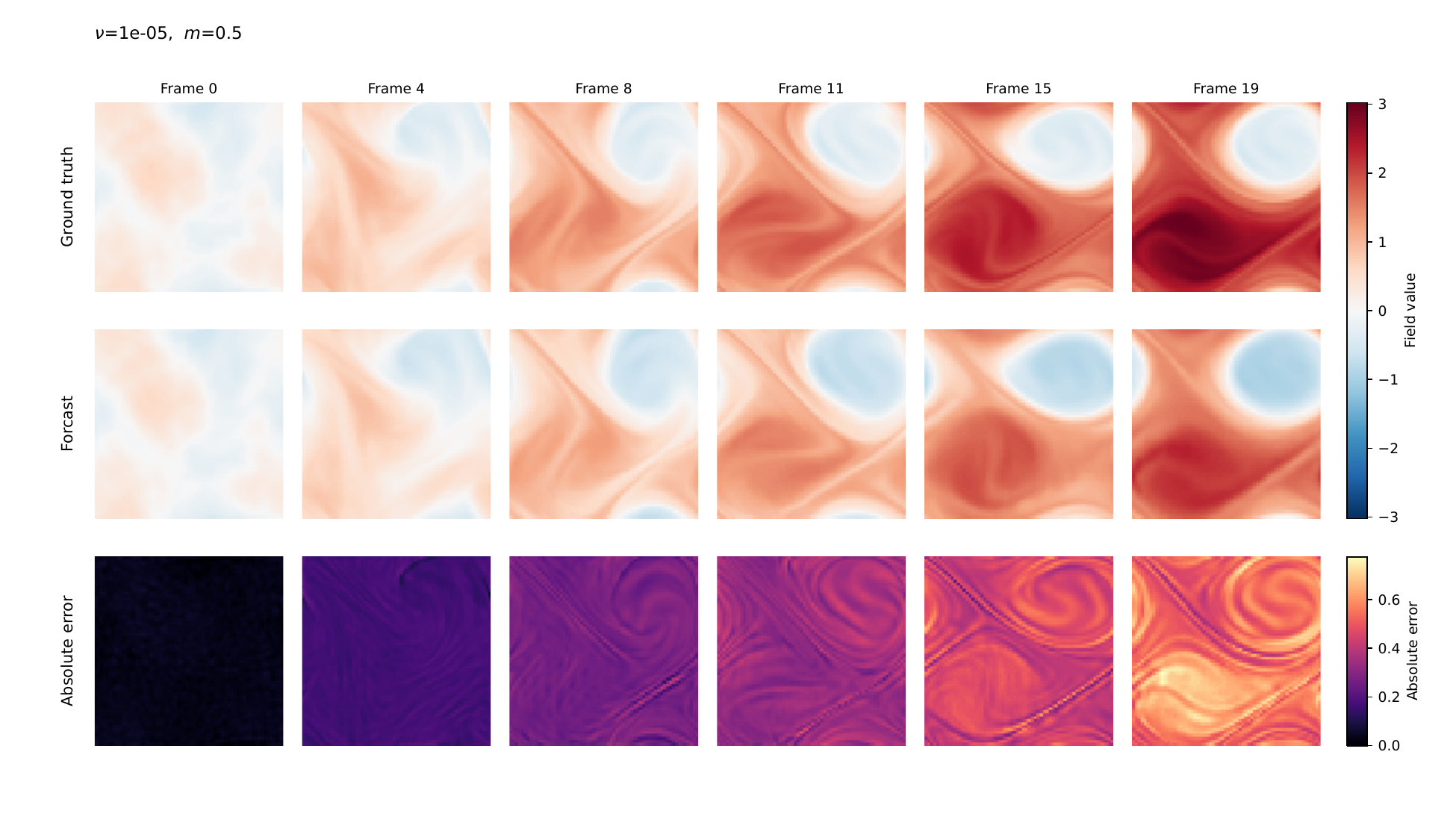}
        \caption{OOD forcing: $\nu=10^{-5},\ m=0.5$}
        \label{fig:heterns_ood_force_vis}
    \end{subfigure}

    \vspace{1mm}

    \begin{subfigure}{0.72\textwidth}
        \centering
        \includegraphics[width=\linewidth]{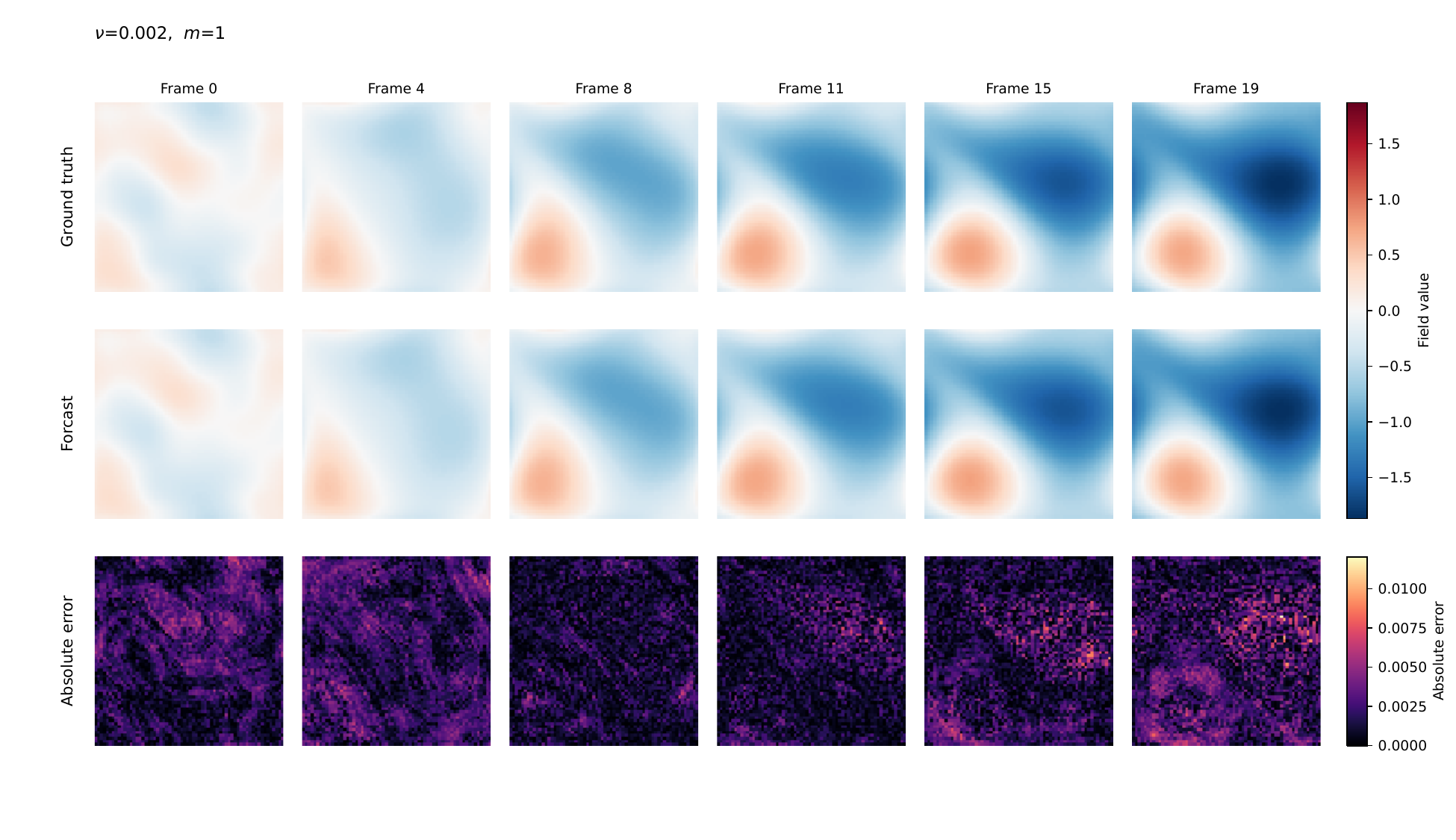}
        \caption{OOD viscosity: $\nu=0.002,\ m=1$}
        \label{fig:heterns_ood_nu_vis}
    \end{subfigure}

    \caption{\textbf{Qualitative HeterNS forecasts under ID and OOD governing conditions.}}
    \label{fig:heterns_visualization}
\end{figure}

Figure~\ref{fig:heterns_visualization} visualizes HeterNS trajectories under representative ID, OOD forcing, and OOD viscosity conditions, showing the ground truth, forecast, and absolute error across representative rollout frames.

\end{document}